\documentclass{article}

\usepackage[final]{neurips}

\usepackage[utf8]{inputenc} 
\usepackage[T1]{fontenc}    
\usepackage{hyperref}       
\usepackage{url}            
\usepackage{booktabs}       
\usepackage{amsfonts}       
\usepackage{nicefrac}       
\usepackage{microtype}      
\usepackage{xcolor}         

\usepackage{graphicx}

\graphicspath{{figs/}}

\title{On the Cross-lingual Transferability of Pre-trained wav2vec2-based Models}

\author{
    Jonatas Grosman \\
    Department of Informatics \\
    Pontifical Catholic University of Rio de Janeiro \\
    Rio de Janeiro - Brazil \\
    \texttt{jgrosman@inf.puc-rio.br} \\
    \And
    Cassio Almeida \\
    National School of Statistical Sciences \\
    Brazilian Institute of Geography and Statistics \\
    Rio de Janeiro - Brazil \\
    \texttt{cassio.almeida@ibge.gov.br} \\
    \And
    Guilherme Schardong \\
    Institute of Systems and Robotics \\
    University of Coimbra \\
    Coimbra - Portugal \\
    \texttt{guilherme.schardong@isr.uc.pt} \\
    \And
    Hélio Lopes \\
    Department of Informatics \\
    Pontifical Catholic University of Rio de Janeiro \\
    Rio de Janeiro - Brazil \\
    \texttt{lopes@inf.puc-rio.br} \\
}

\begin{document}

\maketitle

\begin{abstract}
Using representations provided by a large pre-trained model has become the primary strategy for achieving state-of-the-art results in a wide range of tasks. A recently proposed large pre-trained model, wav2vec 2.0, was seminal for several other works on pre-training large models on speech data. Many models are being pre-trained using the same architecture as wav2vec 2.0 and are getting state-of-the-art in various speech-related tasks. Previous work has demonstrated that the data used during the pre-training of these wav2vec2-based models can impact the model's performance in downstream tasks, and this should be taken into consideration before utilizing these models. However, few works have proposed investigating further how the transfer knowledge of these pre-trained models behaves in different languages, even when the target language differs from the one used during the model's pre-training. Our work aims to investigate the cross-lingual transferability of these wav2vec2-based models. We performed several fine-tuning experiments on the speech recognition task in 18 languages using 15 large pre-trained models. The results of our experiments showed us that the size of data used during the pre-training of these models is not as important to the final performance as the diversity. We noticed that the performance of Indo-European languages is superior to non-Indo-European languages in the evaluated models. We have observed a positive cross-lingual transfer of knowledge using monolingual models, which was evident in all the languages we used, but more pronounced when the language used during pre-training was more similar to the downstream task language. With these findings, we aim to assist the scientific community in utilizing existing wav2vec2-based pre-trained models, as well as facilitate the pre-training of new ones.
\end{abstract}

\section{Introduction}

The pre-training of large models to produce speech representations, which are then used for posterior fine-tuning on speech-related downstream tasks, has recently received considerable attention. Currently, most of these models are trained on a massive amount of unlabeled data using a self-supervised learning approach. A recently proposed large pre-trained model, wav2vec 2.0 \citep{baevski2020wav2vec}, was trained using a self-supervised approach and has obtained the state-of-the-art in several speech-related tasks. Several large pre-trained models have been built using the architecture proposed in the wav2vec 2.0 work, modifying only the training technique or the data used in pre-training \citep{conneau2020unsupervised, hsu2021hubert,chen2021wavlm, babu2021xls, wang2021unispeech}. These models are usually pre-trained in thousands of hours of data from different languages (multilingual models) or just one (monolingual models), achieving good results on downstream tasks even when only a small amount of data is available during fine-tuning. In the speech recognition downstream task, it is possible to obtain good results by fine-tuning these models on only 10 minutes of audio \citep{baevski2020wav2vec}. This is a critical capability when dealing with languages with very few resources available.

Most of the available wav2vec2-based models are pre-trained using only English speech data. However, there are some pre-trained multilingual models, and in some cases, they present even better results on speech recognition tasks than monolinguals that have been trained in the same language \citep{conneau2020unsupervised,babu2021xls}, so the use of multilingual models seems to be a good alternative even though a monolingual model is available. The languages used during pre-training of these multilingual models generally range from a few \citep{wang2021unispeech} to hundreds \citep{babu2021xls}. So, most of the languages spoken in the world, about 7000 \citep{ethnologue}, are missing from these models' training data.

As we currently do not have pre-trained models available in all languages spoken worldwide, we must investigate the cross-lingual transferability of existing wav2vec2-based models. We believe that the variability of languages makes a difference in the final result of the multilingual models. However, in a study where the authors used 128 languages in a wav2vec2-based model pre-training \citep{babu2021xls}, they found no significant difference in the model's performance when comparing the results with another multilingual model that used only 53 languages in its pre-training \citep{conneau2020unsupervised}. The pre-trained model with more languages was significantly better only when it increased the network's size. Perhaps the variability caused by the interference between languages during the pre-training of multilingual models is enough for a model to perform well regardless of the number of languages used. \textbf{Do multilingual wav2vec2-based models of the same size perform similarly during fine-tuning?} Another point that we must take into account regarding the transferability of these multilingual models is that the imbalance in the data of different languages used during pre-training can harm the final performance of the models. For example, in the case of the XLSR-53 model \citep{conneau2020unsupervised}, out of the 56 thousand hours of audio used during pre-training, 44 thousand were English. Some pre-trained multilingual models attempt to mitigate this problem by introducing a factor to balance the contribution of languages during pre-training \citep{babu2021xls}. Still, even with this balancing factor, languages with more diverse and larger datasets are generally Indo-European. We think these pre-trained models can perform better in this language family on downstream tasks. \textbf{Do Indo-European languages perform better compared to non-Indo-European ones on fine-tuning?} In the absence of monolingual models for a given language, some studies have achieved good results by continuing the pre-training of English monolingual models \citep{gupta2021clsril, kim2021k}. \textbf{But even when not continuing the pre-training of a monolingual model in cross-lingual scenarios, is it possible to obtain good results?} In this work, we provide evidence to help solve these doubts by evaluating the results of several fine-tuning experiments in the speech recognition task using 15 models and 18 languages. Our results showed that the performance of Indo-European languages is generally superior to that of non-Indo-European languages in the evaluated models. We have seen different performances using different multilingual models, and we realized that the size of data used during the pre-training of these models is not as crucial to the final performance as the diversity. We observed a positive cross-lingual transfer of knowledge in all the evaluated monolingual models and languages, but this effect was more pronounced when the languages shared a similar language family.

The rest of the paper presents works related to ours (Section \ref{related_work}), details the models, data, and methods used in our experiments (Section \ref{methodology}), and then provides the results and discussions of these experiments (Section \ref{results}). Then we present the conclusions of our work in Section \ref{conclusion}.

\section{Related Work}
\label{related_work}

The work that proposed the wav2vec 2.0 architecture \citep{baevski2020wav2vec} was seminal for several other works on pre-training large models on speech data using self-supervised techniques. Most of these works investigate the \textbf{transferability on large models pre-trained from scratch}. The XLSR-53 \citep{conneau2020unsupervised} was the first version of a multilingual wav2vec2-based model. During its pre-training, this model utilized data from 53 languages, in 56k hours of audio from various domains. One of the most significant findings of this work is that the performance of fine-tuned multilingual models consistently outperforms that of monolingual ones in most cases. To evaluate how these multilingual models behave at even larger scales, considering the size of the network and the amount and variety of training data, a new model called XLS-R was proposed \citep{babu2021xls}. Its pre-training data contained about 436k hours of audio and supported 128 languages, with network sizes ranging from 300M to 2B parameters. In another work \citep{wang2021unispeech}, the authors modified the training approach of the original wav2vec 2.0. They utilized labeled data in their pre-training within a hybrid training regime that combined self-supervised learning and supervised learning techniques, based on the phonetic markers of the transcripts. Several configurations were tested in the experiments conducted in this work, resulting in a monolingual model in English and a multilingual model containing four languages in the pre-training data (en, fr, es, it). In general, the multilingual version performed better on the speech recognition task for languages not seen during its pre-training when compared to the pre-trained English version. Further studies investigated alternative methods for pre-training a model using the wav2vec2 architecture. For example, HuBERT \citep{hsu2021hubert}, which employs a BERT-like prediction loss approach using labels defined for the sound units of each sample during pre-training, utilizing offline clustering techniques. Other works investigated using a speaker-aware approach \citep{chen2021unispeech, chen2021wavlm}. Their approach modified the pre-training data by adding speaker overlays, achieving state-of-the-art performance in several tasks of SUPERB \citep{yang2021superb}, particularly those focused on speaker identification. In another work, the authors investigated how wav2vec2-based models behave in domain shift situations \citep{hsu2021robust}. The authors reported that the similarity between domains improves the final fine-tuning performance. Still, pre-training data from multiple domains favors the model's generalization to domains not seen during pre-training.

Some works investigate the \textbf{cross-lingual transferability on large models already pre-trained}. Most of these works focus on assessing the transferability of English monolingual pre-trained models. A model called CLSRIL-23 \citep{gupta2021clsril} was pre-trained using data from 23 Indic languages, utilizing the weights of the wav2vec 2.0 model at initialization. This model obtained excellent results in speech recognition tasks in several Indic languages. The authors also pre-trained some monolingual models of some Indic languages, but these models could not outperform the results of the multilingual version. Another work obtained excellent results in several benchmarks for the speech recognition task in Korean using a similar approach \citep{kim2021k}. With a focus on the Flemish-Dutch language \citep{ponceletvan2021}, another work conducted several fine-tuning experiments on pre-trained models. The authors used five pre-trained models in this work. Two of them were wav2vec2-based, one was monolingual (wav2vec 2.0), and another was multilingual (XLSR-53). Having these two obtained the best performance among the other pre-trained models. In another study, the authors investigated the ability of an English pre-trained wav2vec-based model to transfer knowledge to other languages using limited data, comparing its results with those of a fully supervised approach \citep{yiwang2021}. Some works propose adjustments in the fine-tuning strategy to improve transfer knowledge, such as a study that investigated the use of English audio transliterations using characters from the target language in the fine-tuning of low-resource languages \citep{kharemittal2021}. Similarly, with the aim of modifying the training data, another work \citep{xu2021simple} constructed a training dataset comprising only the phonetic transcription of multilingual audio files. Then, they built a phoneme recognizer by fine-tuning the XLSR-53 model, surpassing the results of several previous works in zero-shot learning situations in several languages. Other authors have proposed more sophisticated fine-tuning techniques for adapting monolingual models. For example, a work proposed a technique that enables monolingual models to achieve similar performance to multilingual models in certain languages \citep{khuranalaurent2022}.

Most of these works only investigated the cross-lingual transferability of monolingual models in English to a small number of languages. In our work, we conducted experiments with a greater quantity and diversity than those found in previous works, as well as investigating a phenomenon that has not been observed in previous works: cross-lingual transferability between language families.

\section{Methodology}
\label{methodology}

We performed several fine-tuning experiments on wav2vec2-based pre-trained large models. The wav2vec2 architecture can be summarized in a convolutional encoder feature that maps raw audio $X$ to the set of latent speech representations $Z$, which serve as input to a Transformer network \citep{vaswani2017attention} that maps $Z$ to contextual representations $C$.

The difference between the pre-trained models we selected mainly resides in the training strategy and data used during training. We can divide these models into monolingual and multilingual. The first approach uses only data from a single language during pre-training, while the second approach utilizes data from multiple languages. We selected 15 pre-trained models for our experiments. There are different versions of the selected pre-trained models, each with varying parameter sizes. In our experiments, we used the LARGE model from each of them, which has approximately 300 million parameters. During our experiments, we selected data from 18 languages from the Common Voice dataset \citep{ardila2019common} (release 7.0). The Common Voice is a crowdsourced dataset covering dozens of languages and is one of the most extensive multilingual datasets for the speech recognition task. As previous works demonstrated that with a few hours of data, it is possible to obtain results similar to state-of-the-art in several benchmarks in the speech recognition task \citep{baevski2020wav2vec,conneau2020unsupervised,babu2021xls}, we selected the 18 languages for our experiments among those that contained at least five hours of training data in the Common Voice. We used the Common Voice train split during the fine-tuning. The validation and test splits were only used during the evaluation stage. In Appendix~\ref{sec:appendix-models_data} we provide more details about the models and languages used in our experiments. 

We followed an approach similar to that adopted in \citep{baevski2020wav2vec} for the fine-tuning of the pre-trained models. Which consists of using the Connectionist Temporal Classification (CTC) \citep{graves2006connectionist} loss during the fine-tuning of the models for the speech recognition task, adding a linear projection of size $|V|$ randomly initialized on top of the pre-trained network, where $V$ is a set containing all the different characters in the target language (including special characters and whitespace) plus one special character called a blank token, used to control the repetition of characters during decoding the network output. We adopted an automated strategy to generate this $V$ set during our experiments. We selected only non-numeric characters with a frequency greater than or equal to 0.001\% in the training dataset for each language.

We fine-tuned each of the 18 languages on all 15 pre-trained models we selected. To reduce the effects of randomness in our analyses, we repeated this fine-tuning process three times with different seeds. As the data available in Common Voice can vary significantly from one language to another, we randomly selected only five hours of data for each language in each fine-tuning round. To establish a baseline for our analyses, in addition to the 15 selected models, we fine-tuned a wav2vec2 network with randomly initialized weights.

In all fine-tuning and evaluation experiments, we used the HuggingSound tool \citep{grosman2022huggingsound}. We fine-tuned the network using Adam \citep{kingma2014adam} for 4k steps with a learning rate equal to 6e-5. We employ a tri-stage learning rate strategy, where the training learning rate increases linearly during the first 10\% of optimization steps, remains constant for the next 40\%, and decays linearly thereafter. The training batch size consisted of 32 samples, where samples with a duration longer than 12 seconds were discarded during training to prevent memory overflow issues, resulting in a maximum batch size of 384 seconds.

As done in \citep{baevski2020wav2vec}, to leverage the features already learned during the network's pre-training process, we froze the encoder feature block during our fine-tuning experiments. However, we updated the entire network during the training process for baseline models, which were not previously pre-trained.

We used the Character Error Rate (CER) as the evaluation metric for all fine-tuned models. This metric is similar to the Word Error Rate (WER) metric~\citep{morris2004and}, but operates on the character level, computing the minimum number of operations required to transform the predicted transcription into the expected one and extracts the ratio between this value and the total number of characters in the reference transcription, as you can see in the equation~\ref{eq:cer}.

\begin{equation}
  CER = \frac{S + D + I}{S + D + C}
  \label{eq:cer}
\end{equation}

where $S$, $D$, and $I$ are respectively the number of substitutions, deletions, and insertions needed to transform the predicted transcription into the reference one. And $C$ is the number of correct characters in the predicted transcription. We adopted the CER metric as it is generic enough to handle transcripts from any of the experiment's selected languages. For example, the WER metric is incompatible for evaluating Chinese transcriptions, as a group of characters with a white space between them forming a word is not a feature present in Chinese language transcriptions.

We performed statistical analyses for comparisons between models and languages in two stages in our evaluation experiments. First, a statistical test was applied to detect a significant difference between any two groups. Then, if any significant effect was detected, we applied a multiple comparison test to determine the significance between the groups. The statistical test we used for the first stage was the Kruskal–Wallis test \citep{kruskal1952}. For the second stage, we used the Conover-Iman test \citep{conover1979}. We used a significance level of $0.05$ to reject or not the null hypothesis in our tests.

\section{Results and Discussion}
\label{results}

In the following sections, we summarize the results found during our experiments and discuss them. You can find more details about the results of our experiments in appendices \ref{sec:appendix-overall_results}, \ref{sec:appendix-rq1-results}, \ref{sec:appendix-rq2-results}, and \ref{sec:appendix-rq3-results}.

\subsection{Multilingual models performance in different languages}

To test whether the multilingual models perform similarly, we trained each multilingual model three times for each language in the Common Voice dataset. We calculated the CER for each model/language pair for the validation and test splits. Then we performed Kruskal-Wallis tests to assess whether one model performs quite differently from the others in each split. The results show that at least one model has a significant performance difference from the others ($p < 0.05$) in most languages. The only languages and splits where the trained models exhibit statistically similar performance are Dutch in the validation dataset ($p = 0.086$) and Japanese in the test dataset ($p = 0.057$). Details regarding the performance for all the languages can be seen in Appendix~\ref{sec:appendix-rq1-results}, Table~\ref{tab:multilingual_overall_performance}.

Table~\ref{tab:multilingual_performance_ranking_frequency} shows the frequency with which a multilingual model appears at each ranking position per language based on their average CER. From this, we notice that models XLS-R and XLS-53 never rank last for both splits, with XLS-R obtaining the overall best results ($8$ times as the best-ranked model). Additionally, Unispeech-ML appears as the second-best-performing model, ranking as a top performer for $6$ languages considered. However, upon inspecting the languages for which it ranked best, we see that they are either used during pre-training (en, es, fr, it) or have the worst overall performance for all models (ja, zh). Unispeech-ML ranks as the second-to-last model, with 4th place in 8 languages, behind only VoxPopuli-100k, which ranks 4th in 10 languages. A detailed ranking view is presented in Appendix~\ref{sec:appendix-rq1-results}, Table~\ref{tab:multilingual_performance_ranking}.

\begin{table}
    \caption{The frequency with which a multilingual fine-tuned model appears in a particular place in each of the 18 languages' ranking.}
    \footnotesize
    \centering
    \begin{tabular}{p{21mm}p{2mm}p{2mm}p{2mm}p{2mm}}
    \toprule
    & \textbf{1st} & \textbf{2nd} & \textbf{3rd} & \textbf{4th}\\
    \midrule
    \multicolumn{5}{c}{\textbf{Validation}}\\
    \midrule
    XLS-R & 8 & 5 & 5 & 0\\
    UniSpeech-ML & 6 & 0 & 4 & 8\\
    XLSR-53 & 2 & 10 & 6 & 0\\
    VP-100k & 2 & 3 & 3 & 10\\
    \midrule
    \multicolumn{5}{c}{\textbf{Test}}\\
    \midrule
    XLS-R & 8 & 7 & 3 & 0\\
    UniSpeech-ML & 6 & 0 & 4 & 8\\
    XLSR-53 & 3 & 8 & 7 & 0\\
    VP-100k & 1 & 3 & 4 & 10\\
    \bottomrule
    \end{tabular}
    \label{tab:multilingual_performance_ranking_frequency}
\end{table}

By performing a Conover-Iman test for each pair of models on a language basis, we observe that models XLS-53 and XLS-R present no significant performance difference in the same six languages for both splits. These languages are Arabic (ar), Persian (fa), French (fr), Dutch (nl), Japanese (ja), and Chinese (zh). Other model pairs perform similarly on at most $3$ of the languages considered. As for the languages themselves, Dutch, Japanese, and Arabic exhibit the highest statistical similarity between models for both splits. The Table~\ref{tab:multilingual_performance_posthoc_test_frequency} shows the frequency with which the models had no statistical difference ($p > 0.05$) for each split. A detailed view for all model pairs and languages is shown in Appendix~\ref{sec:appendix-rq1-results}, Table~\ref{tab:multilingual_performance_posthoc_test}.

\begin{table}
    \caption{The frequency with which a multilingual fine-tuned model pair showed no statistical difference (p-value above 0.05) in one of the 18 languages.}
    \footnotesize
    \centering
    \begin{tabular}{p{21mm}p{21mm}p{2mm}}
    \toprule
    \multicolumn{3}{c}{\textbf{Validation}}\\
    \midrule
    XLS-R & XLSR-53 & 6\\
    UniSpeech-ML & VP-100k & 3\\
    UniSpeech-ML & XLSR-53 & 3\\
    VP-100k & XLS-R & 2\\
    VP-100k & XLSR-53 & 2\\
    UniSpeech-ML & XLS-R & 1\\
    \midrule
    \multicolumn{3}{c}{\textbf{Test}}\\
    \midrule
    XLS-R & XLSR-53 & 6\\
    UniSpeech-ML & VP-100k & 3\\
    UniSpeech-ML & XLSR-53 & 3\\
    VP-100k & XLS-R & 3\\
    UniSpeech-ML & XLS-R & 2\\
    VP-100k & XLSR-53 & 2\\
    \bottomrule
    \end{tabular}
    \label{tab:multilingual_performance_posthoc_test_frequency}
\end{table}

Given the results presented above, we can infer that some multilingual models performed better than others. Regarding the data used during the pre-training of these multilingual models, it seems that the diversity of the data is more important than quantity. The XLSR-53 presented much better results than the VoxPopuli-100k. The XLSR-53~was pre-trained on $56$k hours of audio from diverse sources, while VoxPopuli-100k~used $100$k hours of speeches from the European Parliament. These findings align with the results reported in the literature~\citep{hsu2021robust}. From our results, we also notice that if we want to use a pre-trained model for a specific domain, we can achieve good results with less data for pre-training than most available pre-trained models. The Unispeech-ML has been pre-trained using only the Common Voice (shown in Table~\ref{tab:pre-trained-models}), the same dataset of our experiments, and presents exceptional results for the languages used during pre-training (en, es, fr, it) with considerably less training data being used ($2$k hours, compared to $436$k hours for XLS-R).

\subsection{Multilingual models performance in different language families}

\begin{figure}
  \centering
  \includegraphics[width=\columnwidth]{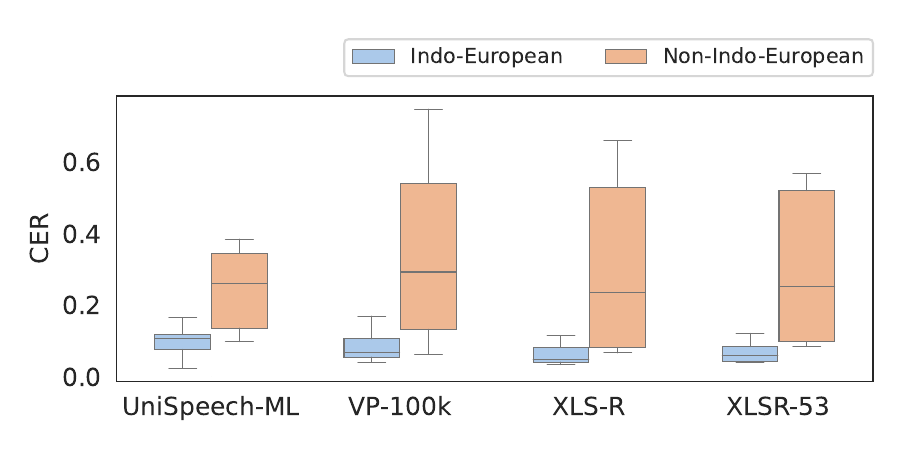}
  \caption{Multilingual pre-trained models performance over \textbf{grouped} language families on the \textbf{validation} set}
  \label{fig:f_multilingual_family_grouped_validation}
\end{figure}

\begin{figure}
  \centering
  \includegraphics[width=\columnwidth]{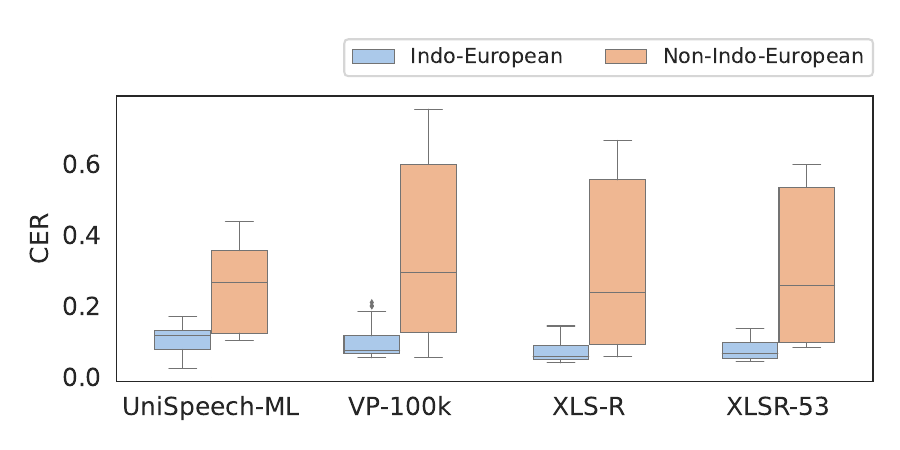}
  \caption{Multilingual pre-trained models performance over \textbf{grouped} language families on the \textbf{test} set}
  \label{fig:f_multilingual_family_grouped_test}
\end{figure}

When we grouped the results between Indo-European and non-Indo-European families, we noticed an apparent difference in performance between these two groups over all the multilingual models, as you can observe in Figures~\ref{fig:f_multilingual_family_grouped_validation} and ~\ref{fig:f_multilingual_family_grouped_test}. The non-Indo-European group presents a large dispersion of the CER. That overdispersion suggests a heterogeneous performance in each model depending on the language contained in the non-Indo-European group. We investigated the performances by language families to analyze this heterogeneity. The Uralic and Austronesian families present the best performances with an average CER similar to that obtained in the Indo-European family. The Japonic and Sino-Tibetan families have the highest average errors. The best performance for these families was an average CER of $0.382$ for the Japonic and $0.357$ for the Sino-Tibetan. The other families, Afro-Asian and Kra-Dai, present average values of intermediate CER. These results suggest that the great dispersion of the average values of CER in this group is due to the characteristics of the language families. Details can be seen in Appendix~\ref{sec:appendix-rq2-results}.

We performed Kruskal-Wallis tests to assess whether one group of languages performed differently from the other. The results indicated strong evidence ($p \approx 0$) of a significant difference in all the evaluated models in both validation and test splits. However, Japanese and Chinese have a much larger number of possible characters in their writing systems than the other languages we evaluated. This difference can make the speech recognition task more complex in these two languages. We then ran statistical tests comparing the performance of Indo-European and non-Indo-European languages, excluding Chinese and Japanese languages, to assess whether the differences between the family groups persisted even after removing these languages from our analysis. This test revealed that, even when excluding Chinese and Japanese, there is a statistically significant difference between Indo-European and non-Indo-European families in all evaluated models. Details of these tests can be found in Appendix~\ref{sec:appendix-rq2-results}, in Tables~\ref{tab:multilingual_overall_performance_grouped_family_T} and \ref{tab:multilingual_overall_performance_grouped_family_T_wt_ja_zh}. Considering the evidence from the statistical test, and given that the CER averages for all models in the Indo-European group were consistently lower than those in the non-Indo-European group, we can conclude that all models performed better for the Indo-European family. To reinforce these results, we proceeded with another test using the models as factors instead of the group of languages. The test results showed that only the Indo-European group has at least one model with a different performance from the others. The non-Indo-European group has no significant differences in the performance between models. For detailed results see in Appendix~\ref{sec:appendix-rq2-results} in Table~\ref{tab:multilingual_overall_performance_grouped_family}.

Our results showed that the Indo-European language group performed consistently better than the non-Indo-European language group. Except for Estonian and Indonesian, all other non-Indo-European languages performed much worse than the Indo-European ones. This divergent performance between the groups may be related to the fact that the data used during the pre-training of all the multilingual models we used in our experiments were more diverse and abundant for the Indo-European languages. This imbalance may have made languages in this family have a more significant influence on the weights of the pre-trained network. And what we observe in our experiments is just a reflection of that. This pre-training data effect aligns with the findings in \citep{conneau2020unsupervised}, which argue that the similarity of languages used in pre-training plays a crucial role in the model's final performance.

\subsection{Monolingual models' performance in different languages}

For our investigation regarding cross-lingual transferability in monolingual models, we selected 11 models. The vast majority of publicly available monolingual models are in English. Our monolingual model selection reflects the dominance of the English model, as 6 out of 11 selected models were English. Our analysis was then segmented between English and non-English models. The non-English models were pre-trained using the VoxPopuli dataset in Spanish, French, Italian, Dutch, and Swedish. In addition to the selected monolingual models, we have included a model without pre-training in our tests to establish a performance baseline.

The statistical tests we conducted using the performance of all pre-trained English models showed that, for all languages, at least one model had significantly different performance in both the validation and test splits. We performed a Conover-Iman test to see which model(s) had different behavior. Comparing the number of times each model pair presented statistically similar results, surprisingly, \texttt{UniSpeech-SAT} produced similar results with models without pre-training in $14$ cases for the validation dataset and $15$ for the test dataset. A more detailed view is shown in Table~\ref{tab:en_monolingual_performance_posthoc_test_frequency}, Appendix~\ref{sec:appendix-rq3-results}. A similar result for the VoxPopuli pre-trained models is shown for the model pre-trained in Dutch, where this model and models without pre-training presented $p > 0.05$ for $10$ languages in both splits (Appendix~\ref{sec:appendix-rq3-results}, Table~\ref{tab:vp_monolingual_performance_posthoc_test_frequency}).

Tables~\ref{tab:en_monolingual_performance_ranking_frequency} and~\ref{tab:vp_monolingual_performance_ranking_frequency} show the ranking frequency of the models tested. We have included the best multilingual models for each language for comparison. Unsurprisingly, the multilingual models rank better for most languages, except for English, where \texttt{UniSpeech} presents superior results.

\begin{table}
    \caption{The frequency with which an English monolingual fine-tuned model appears in a particular place in each of the 18 languages' ranking. Multilingual models are presented in bold-face.}
    \footnotesize
    \centering
    \begin{tabular}{p{21mm}p{2mm}p{2mm}p{2mm}p{2mm}p{2mm}p{2mm}p{2mm}p{2mm}}
    \toprule
    & \textbf{1st} & \textbf{2nd} & \textbf{3rd} & \textbf{4th} & \textbf{5th} & \textbf{6th} & \textbf{7th} & \textbf{8th}\\
    \midrule
    \multicolumn{9}{c}{\textbf{Validation}}\\
    \midrule
    \textbf{XLS-R} & 8 & 0 & 0 & 0 & 0 & 0 & 0 & 0\\
    \textbf{UniSpeech-ML} & 5 & 1 & 0 & 0 & 0 & 0 & 0 & 0\\
    \textbf{XLSR-53} & 2 & 0 & 0 & 0 & 0 & 0 & 0 & 0\\
    \textbf{VP-100k} & 2 & 0 & 0 & 0 & 0 & 0 & 0 & 0\\
    UniSpeech & 1 & 2 & 1 & 3 & 4 & 6 & 1 & 0\\
    WavLM & 0 & 9 & 4 & 2 & 1 & 1 & 1 & 0\\
    R-wav2vec2 & 0 & 5 & 7 & 3 & 1 & 1 & 1 & 0\\
    HuBERT & 0 & 1 & 1 & 3 & 7 & 5 & 1 & 0\\
    wav2vec2 & 0 & 0 & 4 & 6 & 4 & 4 & 0 & 0\\
    UniSpeech-SAT & 0 & 0 & 1 & 1 & 1 & 1 & 14 & 0\\
    no-pretraining & 0 & 0 & 0 & 0 & 0 & 0 & 0 & 18\\
    \midrule
    \multicolumn{9}{c}{\textbf{Test}}\\
    \midrule
    \textbf{XLS-R} & 8 & 0 & 0 & 0 & 0 & 0 & 0 & 0\\
    \textbf{UniSpeech-ML} & 5 & 1 & 0 & 0 & 0 & 0 & 0 & 0\\
    \textbf{XLSR-53} & 3 & 0 & 0 & 0 & 0 & 0 & 0 & 0\\
    \textbf{VP-100k} & 1 & 0 & 0 & 0 & 0 & 0 & 0 & 0\\
    UniSpeech & 1 & 3 & 1 & 2 & 3 & 7 & 1 & 0\\
    WavLM & 0 & 8 & 6 & 1 & 1 & 1 & 1 & 0\\
    R-wav2vec2 & 0 & 5 & 9 & 1 & 1 & 1 & 1 & 0\\
    HuBERT & 0 & 1 & 1 & 3 & 8 & 4 & 1 & 0\\
    UniSpeech-SAT & 0 & 0 & 1 & 1 & 0 & 2 & 14 & 0\\
    wav2vec2 & 0 & 0 & 0 & 10 & 5 & 3 & 0 & 0\\
    no-pretraining & 0 & 0 & 0 & 0 & 0 & 0 & 0 & 18\\
    \bottomrule
    \end{tabular}
    \label{tab:en_monolingual_performance_ranking_frequency}
\end{table}

\begin{table}
    \caption{The frequency with which a VoxPopuli monolingual fine-tuned model appears in a particular place in each of the 18 languages' ranking. Multilingual models are presented in bold-face.}
    \footnotesize
    \centering
    \begin{tabular}{p{25mm}p{3mm}p{3mm}p{3mm}p{3mm}p{3mm}p{3mm}p{3mm}}
    \toprule
    & \textbf{1st} & \textbf{2nd} & \textbf{3rd} & \textbf{4th} & \textbf{5th} & \textbf{6th} & \textbf{7th}\\
    \midrule
    \multicolumn{8}{c}{\textbf{Validation}}\\
    \midrule
    \textbf{XLS-R} & 8 & 0 & 0 & 0 & 0 & 0 & 0\\
    \textbf{UniSpeech-ML} & 6 & 0 & 0 & 0 & 0 & 0 & 0\\
    \textbf{XLSR-53} & 2 & 0 & 0 & 0 & 0 & 0 & 0\\
    \textbf{VP-100k} & 2 & 0 & 0 & 0 & 0 & 0 & 0\\
    VP-es & 0 & 5 & 2 & 3 & 4 & 4 & 0\\
    VP-sv & 0 & 5 & 3 & 2 & 2 & 6 & 0\\
    VP-fr & 0 & 4 & 5 & 7 & 2 & 0 & 0\\
    VP-nl & 0 & 3 & 2 & 2 & 7 & 4 & 0\\
    VP-it & 0 & 1 & 6 & 4 & 3 & 4 & 0\\
    no-pretraining & 0 & 0 & 0 & 0 & 0 & 0 & 18\\
    \midrule
    \multicolumn{8}{c}{\textbf{Test}}\\
    \midrule
    \textbf{XLS-R} & 8 & 0 & 0 & 0 & 0 & 0 & 0\\
    \textbf{UniSpeech-ML} & 6 & 0 & 0 & 0 & 0 & 0 & 0\\
    \textbf{XLSR-53} & 3 & 0 & 0 & 0 & 0 & 0 & 0\\
    \textbf{VP-100k} & 1 & 0 & 0 & 0 & 0 & 0 & 0\\
    VP-sv & 0 & 5 & 2 & 3 & 3 & 5 & 0\\
    VP-fr & 0 & 4 & 4 & 7 & 2 & 1 & 0\\
    VP-es & 0 & 4 & 3 & 3 & 4 & 4 & 0\\
    VP-nl & 0 & 3 & 4 & 0 & 7 & 4 & 0\\
    VP-it & 0 & 2 & 5 & 5 & 2 & 4 & 0\\
    no-pretraining & 0 & 0 & 0 & 0 & 0 & 0 & 18\\
    \bottomrule
    \end{tabular}
    \label{tab:vp_monolingual_performance_ranking_frequency}
\end{table}

The VoxPopuli results show an interesting fact: when pre-training and fine-tuning take the language family and subgroup into consideration, the resulting model yields better results, at least for Indo-European languages. Let us focus on two languages, Portuguese (pt) and German (de). Portuguese and German belong to the same family (Indo-European) but different subgroups (Italic and Germanic). For Portuguese, the monolingual models ranking excluding the multilingual models is VP-es, VP-it, VP-fr, VP-nl, and VP-sv. The first $3$ models are pre-trained in languages of the same subgroup as Portuguese. Making the same comparison for German, the model ranking is composed of VP-nl, VP-sv, VP-fr, VP-es, and VP-it. The first $2$ models are pre-trained in languages of the same subgroup as German. The same pattern also occurs in other Germanic and Italic languages.

The results of our experiments demonstrated a positive transfer of knowledge in nearly all the scenarios we evaluated, where all the monolingual models outperformed the models without pre-training. Furthermore, it appears that we can enhance knowledge transfer by using models pre-trained in a language similar to the fine-tuning language. A final interesting finding is about the performance of \texttt{UniSpeech} pre-trained in English and fine-tuned in both Japanese (Figure~\ref{fig:f_overall_ja} in Appendix~\ref{sec:appendix-rq3-results}) and Chinese (Figure~\ref{fig:f_overall_zh}, Appendix~\ref{sec:appendix-rq3-results}). In both cases, the performance of \texttt{UniSpeech} surpassed that of some multilingual models pre-trained using Japanese and Chinese speech data. Perhaps this behavior is associated with the low representation of these languages in the pre-training of these multilingual models. The more detailed results can be seen in Appendix~\ref{sec:appendix-rq3-results}.

\section{Conclusion}
\label{conclusion}

To evaluate the cross-lingual transferability of pre-trained wav2vec2-based models, we conducted fine-tuning experiments for the speech recognition task in large pre-trained models. To the best of our knowledge, we employ a combination of models and languages on a scale that has not been seen in other works. More precisely, we used 15 models and 18 languages during our experiments.

The results obtained from our experiments showed that Indo-European languages generally performed superiorly compared to other language families. It may be related to the fact that the amount and diversity of data used during the pre-training of the models were greater for Indo-European languages than for other language families. Another point we noticed during our experiments was that the amount of data used during models' pre-training does not seem to have as much influence on the final performance of the model as the diversity. Multilingual models with more varied data generally performed better than other models with more data but less diversity. When investigating the effect of cross-lingual transferability in monolingual models, we found a positive transfer of knowledge in all evaluated scenarios. We observed that knowledge transfer appears stronger when the language used during fine-tuning belongs to the same subgroup as the language used during the model's pre-training. We observed that multilingual models perform better than monolingual models, even when the data used for pre-training and fine-tuning were from the same language. With our findings, we aim to contribute to the pre-training of new models, as well as the effective use of existing ones.

In future work, we would like to investigate more in-depth why Indo-European languages generally performed better than other language families in our experiments. For this, we can adopt approaches similar to those in previous works, which have continued the pre-training of already pre-trained models~\citep{gupta2021clsril, kim2021k}. We could continue pre-training using different combinations of language families in the pre-training data and evaluating the effects this has on the final results of the model on different languages. We would also like to investigate other potential effects resulting from the imbalanced data used during pre-training. For instance, most of the data used during the pre-training of these models typically comes from male voices; this gender imbalance may affect the final performance of the pre-trained model.

\section*{Limitations}
All our findings are limited to the behavior we observed in publicly available pre-trained models. We could obtain more robust results by pre-training large wav2vec2-based models from scratch using different language combinations in the training data. However, the high cost of pre-training those models makes this kind of analysis extremely expensive, and we did not have enough resources. Even the fine-tuning of these large models has a high cost, which meant that we had to run a limited number of fine-tuning rounds using the same hyperparameter configuration defined in the initial stage of our work, varying only the seed during the training. So we could not test whether our findings still held up using different combinations of hyperparameters during the fine-tuning experiments.

Since we only use the Common Voice dataset in our experiments, we cannot determine whether our findings would remain valid in a different domain. The decision not to run our experiments on multiple datasets is due to the high cost of fine-tuning, as mentioned above. Therefore, due to our limited resources, we had to choose between a more significant variability of datasets or languages. We preferred to opt for the second option. Therefore, we chose to use the Common Voice dataset, the only publicly available dataset for the Speech Recognition task with a sufficiently diverse set of languages to perform the experiments we had imagined. But even though the distribution of speech hours by language is quite unbalanced. There are many more hours in Indo-European languages than in other language families. Few non-Indo-European languages were able to provide the five hours of training data required for our experiments, as reflected in the final selection of languages we used. Those limitations make our findings for Indo-European families more robust than those related to other language families.

\section*{Acknowledgements}

The first author wish to acknowledge the financial support from the Brazilian National Research Council (CNPq; grant 141763/2018-3).

\bibliography{references}
\bibliographystyle{acl_natbib}

\clearpage

\appendix

\section{Models and data details}
\label{sec:appendix-models_data}

This section presents the details about the models and data used in our experiments. You can see the details about the pre-trained models used in Table~\ref{tab:pre-trained-models} and a summary of the languages and data in Table~\ref{tab:languages}

\begin{table*}
    \caption{Pre-trained models used during our experiments}
    \label{tab:pre-trained-models}
    \footnotesize
    \centering
    \begin{tabular}{ p{35mm} p{18mm} p{75mm} }
    \toprule
    \textbf{Pre-trained model} & \textbf{Languages} & \textbf{Pre-training data} \\
    \midrule
    wav2vec 2.0~\citeyearpar{baevski2020wav2vec} & en & 60k hours (Libri-Light~\citeyearpar{librilight}) \\
    XLSR-53~\citeyearpar{conneau2020unsupervised} & 53 languages & 56k hours (MLS~\citeyearpar{Pratap2020MLSAL}, Common Voice~\citeyearpar{ardila2019common}, BABEL~\citeyearpar{gales2014speech}) \\
    HuBERT~\citeyearpar{hsu2021hubert} & en & 60k hours (Libri-Light~\citeyearpar{librilight}) \\
    UniSpeech~\citeyearpar{wang2021unispeech} & en & 1350 hours (Common Voice~\citeyearpar{ardila2019common})  \\
    UniSpeech-ML~\citeyearpar{wang2021unispeech} & en, fr, es, it & 1961 hours (Common Voice~\citeyearpar{ardila2019common})  \\
    VP-100k~\citeyearpar{wang2021voxpopuli} & en, de, fr, es, pl, it, ro, hu, cs, nl, fi, hr, sk, sl, et, lt, pt, bg, el, lv, mt, sv, da & 100k hours (VoxPopuli~\citeyearpar{wang2021voxpopuli})  \\
    VP-sv~\citeyearpar{wang2021voxpopuli} & sv & 16k hours (VoxPopuli~\citeyearpar{wang2021voxpopuli})  \\
    VP-nl~\citeyearpar{wang2021voxpopuli} & nl & 19k hours (VoxPopuli~\citeyearpar{wang2021voxpopuli})  \\
    VP-it~\citeyearpar{wang2021voxpopuli} & it & 21k hours (VoxPopuli~\citeyearpar{wang2021voxpopuli})  \\
    VP-es~\citeyearpar{wang2021voxpopuli} & es & 21k hours (VoxPopuli~\citeyearpar{wang2021voxpopuli})  \\
    VP-fr~\citeyearpar{wang2021voxpopuli} & fr & 22k hours (VoxPopuli~\citeyearpar{wang2021voxpopuli})  \\
    Robust wav2vec 2.0~\citeyearpar{hsu2021robust} & en & 63k hours (Libri-Light~\citeyearpar{librilight}, Common Voice~\citeyearpar{ardila2019common}, Switchboard~\citeyearpar{godfrey1993switchboard}, Fisher~\citeyearpar{cieri2004fisher})  \\
    UniSpeech-SAT~\citeyearpar{chen2021unispeech} & en & 94k hours (Libri-Light~\citeyearpar{librilight}, GigaSpeech~\citeyearpar{chen2021gigaspeech}, VoxPopuli~\citeyearpar{wang2021voxpopuli})  \\
    WavLM~\citeyearpar{chen2021wavlm} & en & 94k hours (Libri-Light~\citeyearpar{librilight}, GigaSpeech~\citeyearpar{chen2021gigaspeech}, VoxPopuli~\citeyearpar{wang2021voxpopuli})  \\
    XLS-R~\citeyearpar{babu2021xls} & 128 languages & 436k hours (VoxPopuli~\citeyearpar{wang2021voxpopuli}, MLS~\citeyearpar{Pratap2020MLSAL}, Common Voice~\citeyearpar{ardila2019common}, VoxLingua107~\citeyearpar{valk2021voxlingua107}, BABEL~\citeyearpar{gales2014speech}) \\
    \bottomrule
    \end{tabular}
\end{table*}

\begin{table*}
    \caption{Summary of the languages used in our experiments and the Common Voice \citep{ardila2019common} dataset size in hours for each of these languages. The language families are described according to the Ethnologue \citep{ethnologue} catalog.}
    \label{tab:languages}
    \footnotesize
    \centering
    \begin{tabular}{llllrrr}
    \toprule
    \textbf{Language} & \textbf{ISO code} & \textbf{Family} & \textbf{Sub-grouping} & \textbf{Train} & \textbf{Val.} & \textbf{Test}\\
    \midrule
    Arabic & ar & Afro-Asiatic & Semitic & 30h & 12h & 12h\\
    Chinese & zh & Sino-Tibetan & Chinese & 31h & 14h & 15h\\
    Dutch & nl & Indo-European & Germanic & 31h & 13h & 13h\\
    English & en & Indo-European & Germanic & 1209h & 27h & 26h\\
    Estonian & et & Uralic & Coastal Finnic & 5h & 4h & 4h\\
    French & fr & Indo-European & Italic & 545h & 25h & 25h\\
    German & de & Indo-European & Germanic & 572h & 26h & 26h\\
    Indonesian & id & Austronesian & Malayo-Polynesian & 5h & 1h & 0.5h\\
    Italian & it & Indo-European & Italic & 190h & 24h & 25h\\
    Japanese & ja & Japonic &  & 8h & 2h & 5h\\
    Persian & fa & Indo-European & Indo-Iranian & 13h & 8h & 10h\\
    Polish & pl & Indo-European & Balto-Slavic & 16h & 8h & 9h\\
    Portuguese & pt & Indo-European & Italic & 14h & 8h & 9h\\
    Russian & ru & Indo-European & Balto-Slavic & 25h & 13h & 13h\\
    Spanish & es & Indo-European & Italic & 291h & 25h & 26h\\
    Swedish & sv & Indo-European & Germanic & 7h & 4h & 5h\\
    Thai & th & Kra-Dai & Kam-Tai & 27h & 12h & 0.3h\\
    Ukrainian & uk & Indo-European & Balto-Slavic & 7h & 6h & 6h\\
    \bottomrule
    \end{tabular}
\end{table*}

\section{Overall results}
\label{sec:appendix-overall_results}

This section presents the overall results of our tests in all languages considered.


\begin{figure*}
  \centering
  \includegraphics[width=\textwidth]{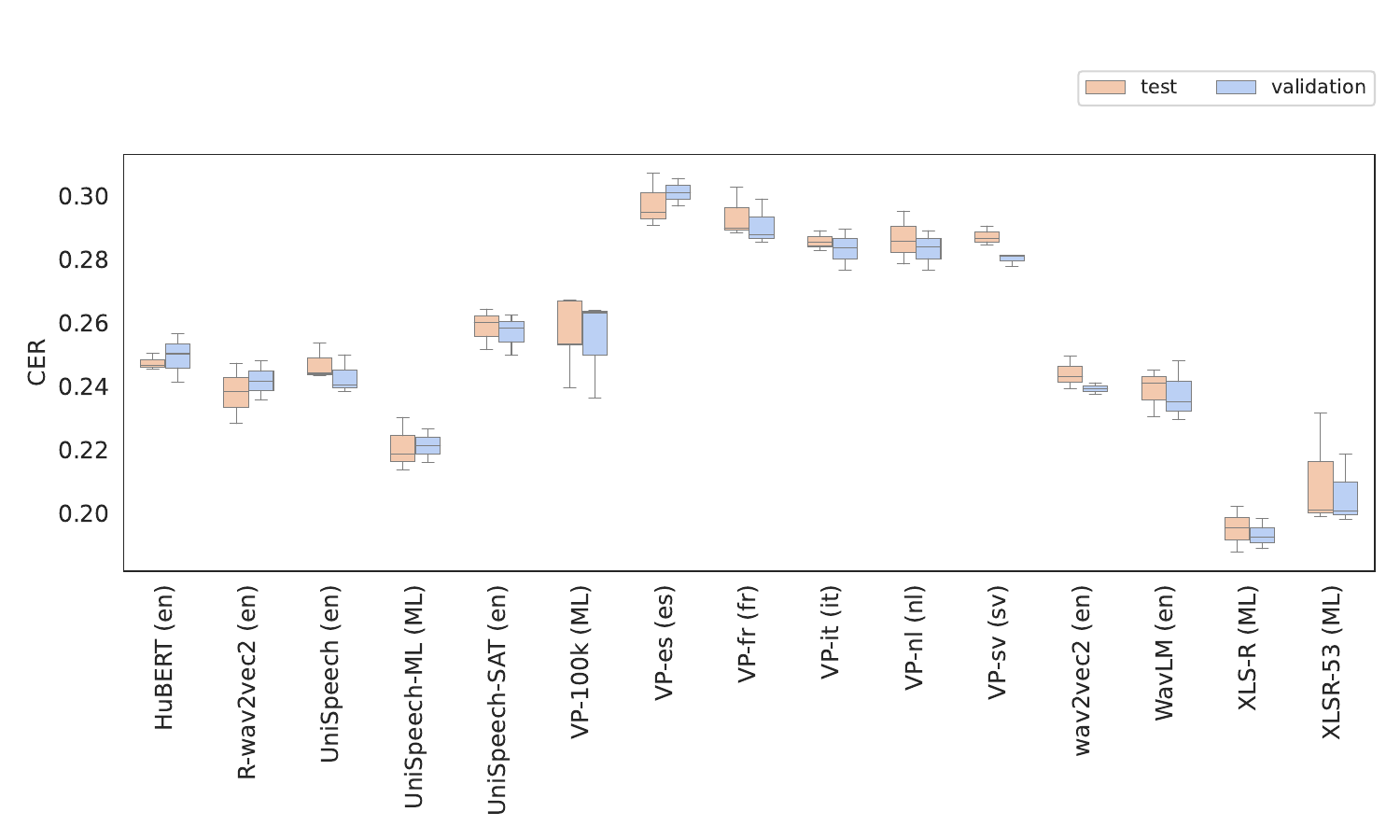}
  \caption{Overall performance over the pre-trained models for \textbf{Arabic} language}
  \label{fig:f_overall_ar}
\end{figure*}

\begin{figure*}
  \centering
  \includegraphics[width=\textwidth]{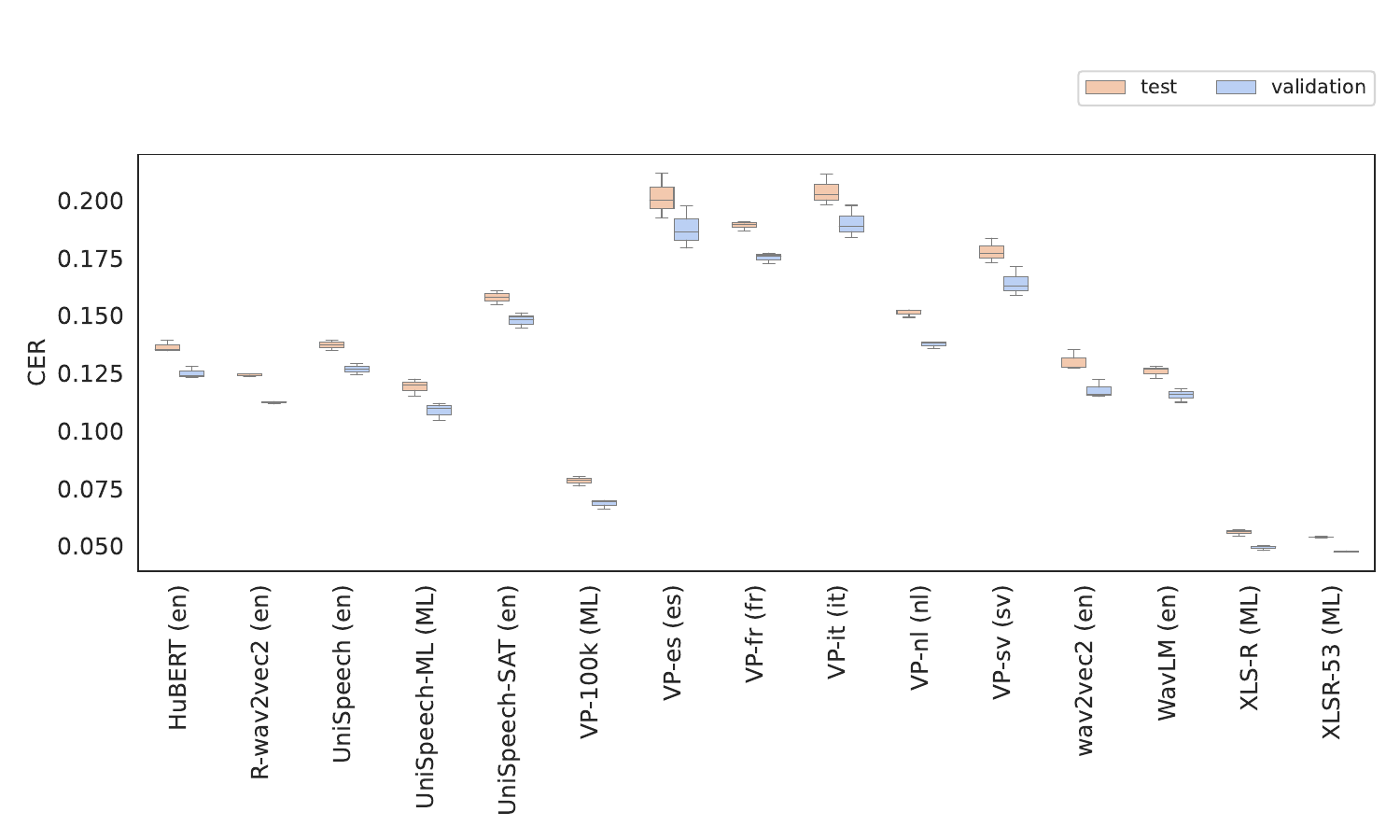}
  \caption{Overall performance over the pre-trained models for \textbf{German} language}
  \label{fig:f_overall_de}
\end{figure*}

\begin{figure*}
  \centering
  \includegraphics[width=\textwidth]{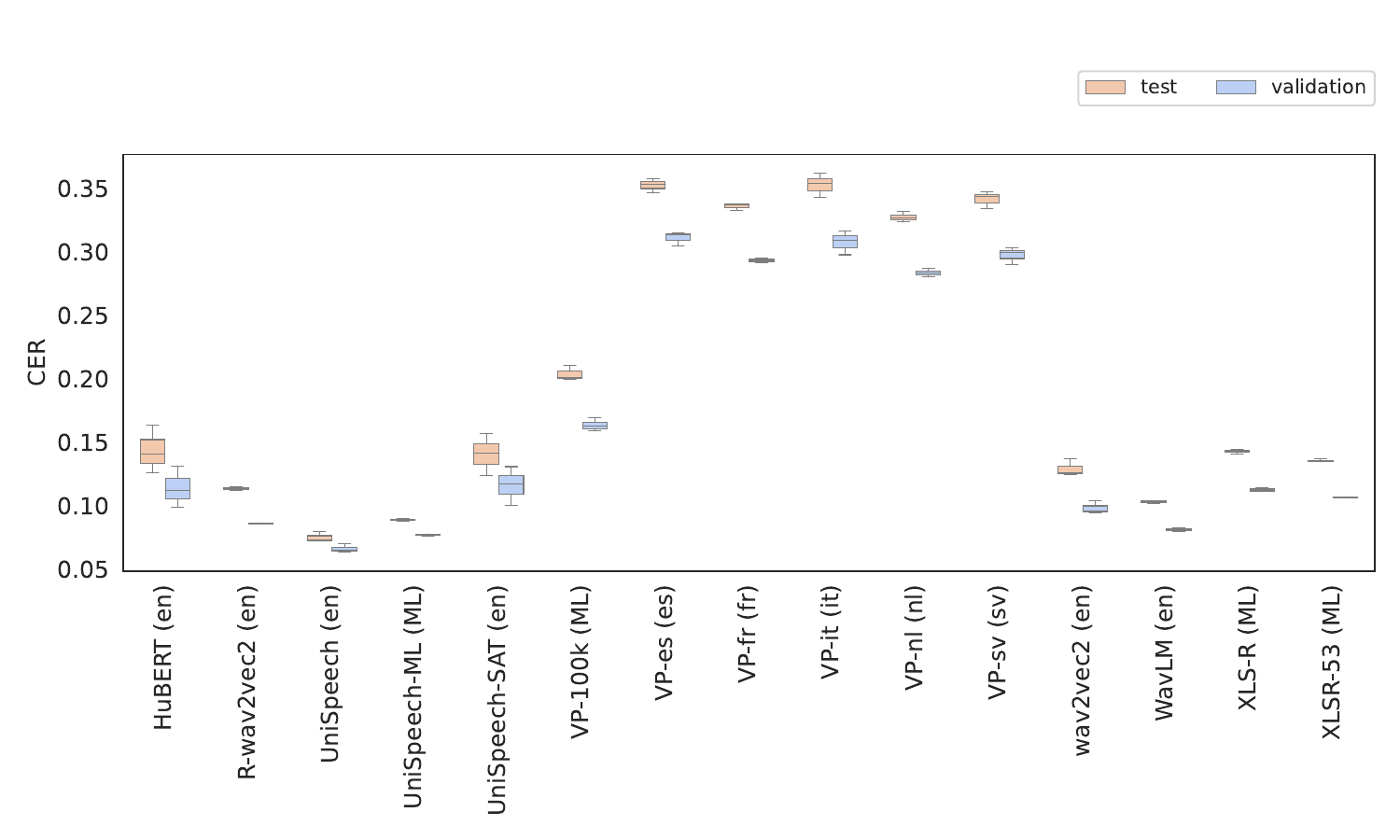}
  \caption{Overall performance over the pre-trained models for \textbf{English} language}
  \label{fig:f_overall_en}
\end{figure*}

\begin{figure*}
  \centering
  \includegraphics[width=\textwidth]{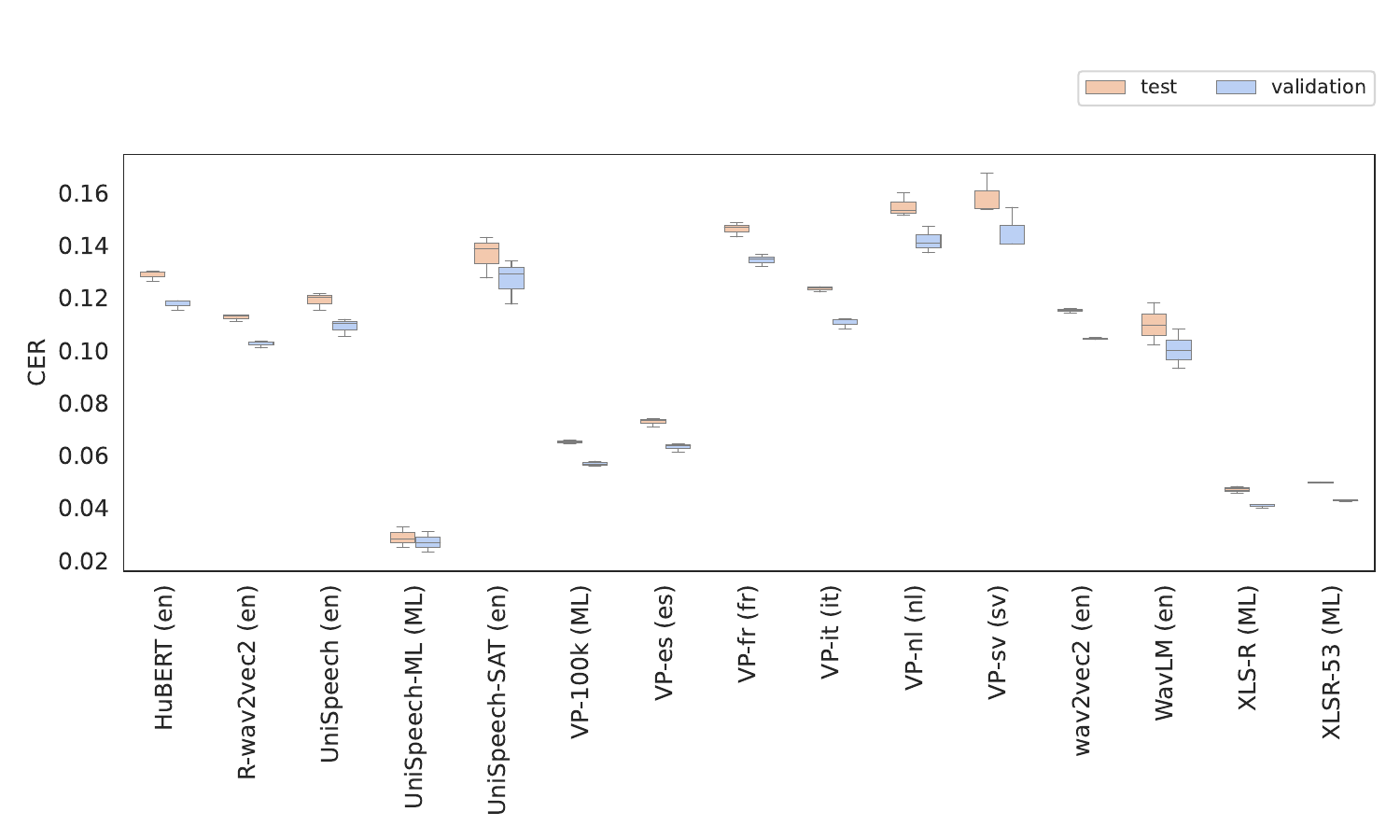}
  \caption{Overall performance over the pre-trained models for \textbf{Spanish} language}
  \label{fig:f_overall_es}
\end{figure*}

\begin{figure*}
  \centering
  \includegraphics[width=\textwidth]{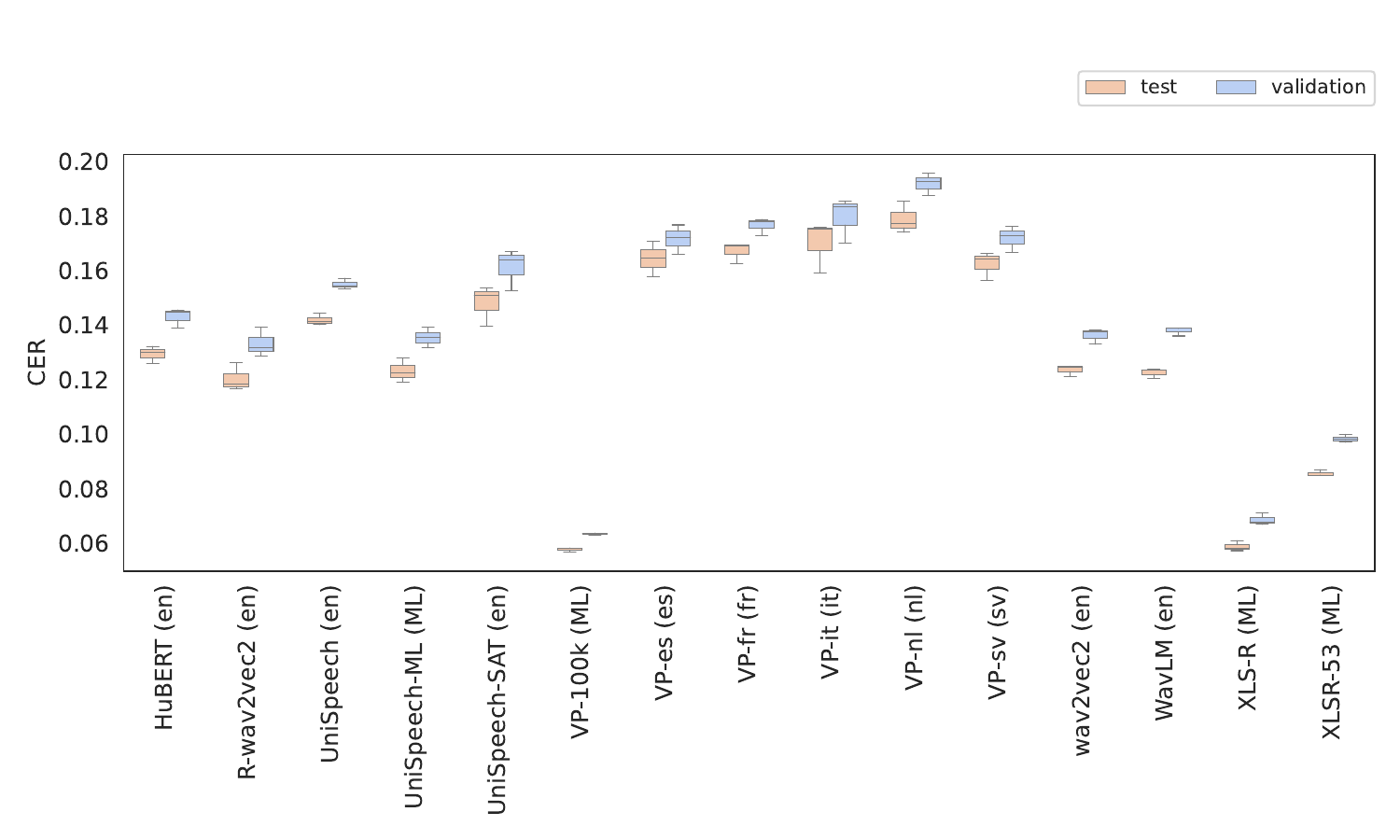}
  \caption{Overall performance over the pre-trained models for \textbf{Estonian} language}
  \label{fig:f_overall_et}
\end{figure*}

\begin{figure*}
  \centering
  \includegraphics[width=\textwidth]{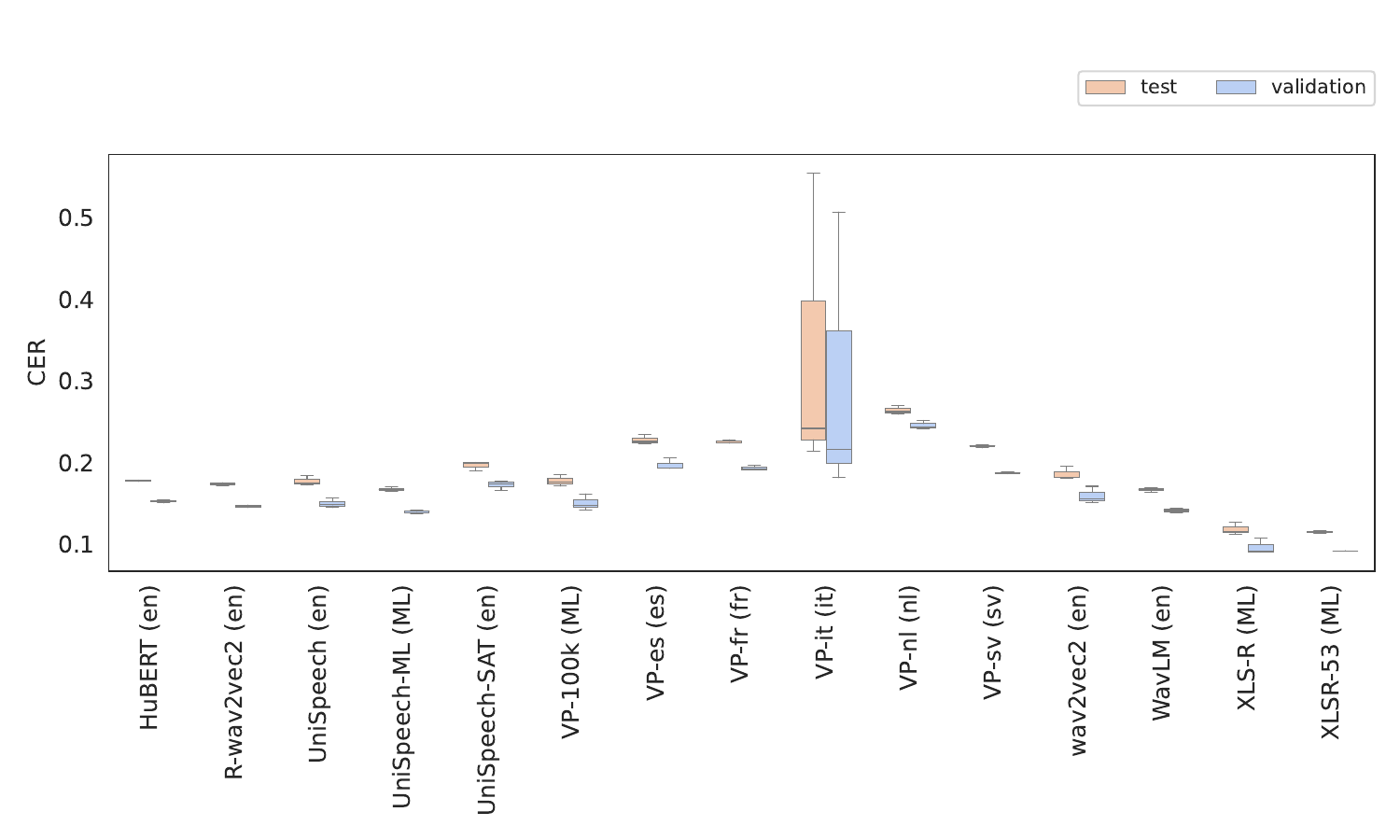}
  \caption{Overall performance over the pre-trained models for \textbf{Persian} language}
  \label{fig:f_overall_fa}
\end{figure*}

\begin{figure*}
  \centering
  \includegraphics[width=\textwidth]{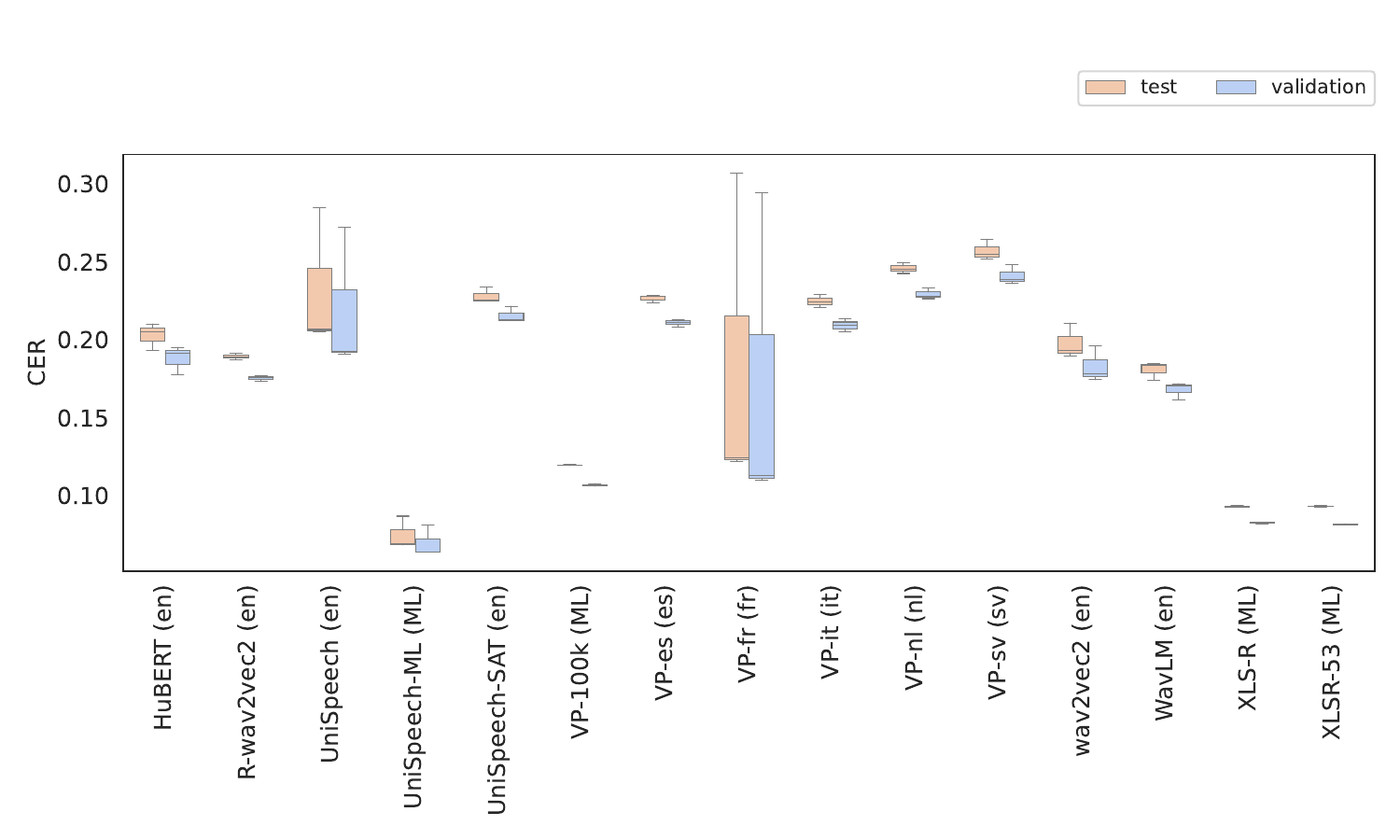}
  \caption{Overall performance over the pre-trained models for \textbf{French} language}
  \label{fig:f_overall_fr}
\end{figure*}

\begin{figure*}
  \centering
  \includegraphics[width=\textwidth]{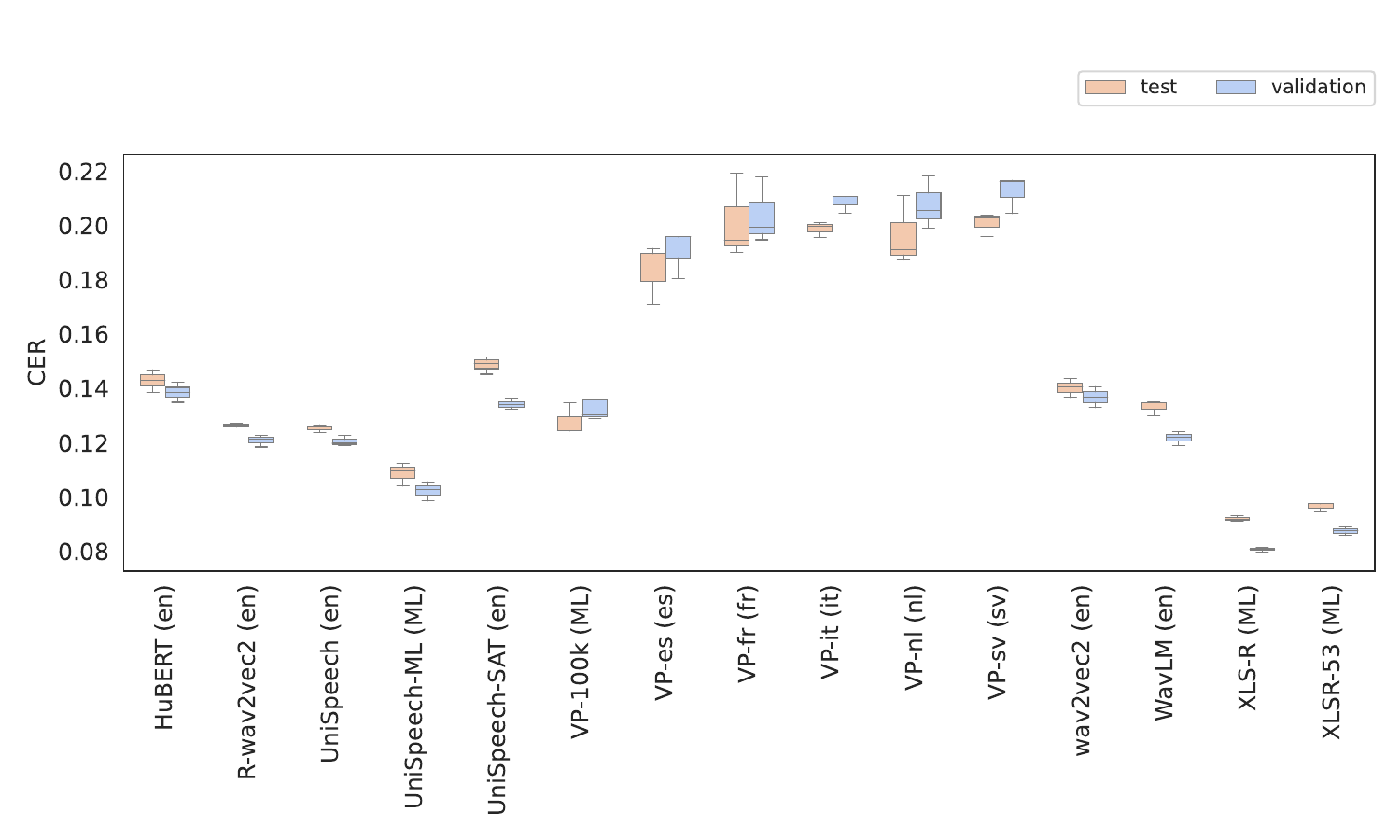}
  \caption{Overall performance over the pre-trained models for \textbf{Indonesian} language}
  \label{fig:f_overall_id}
\end{figure*}

\begin{figure*}
  \centering
  \includegraphics[width=\textwidth]{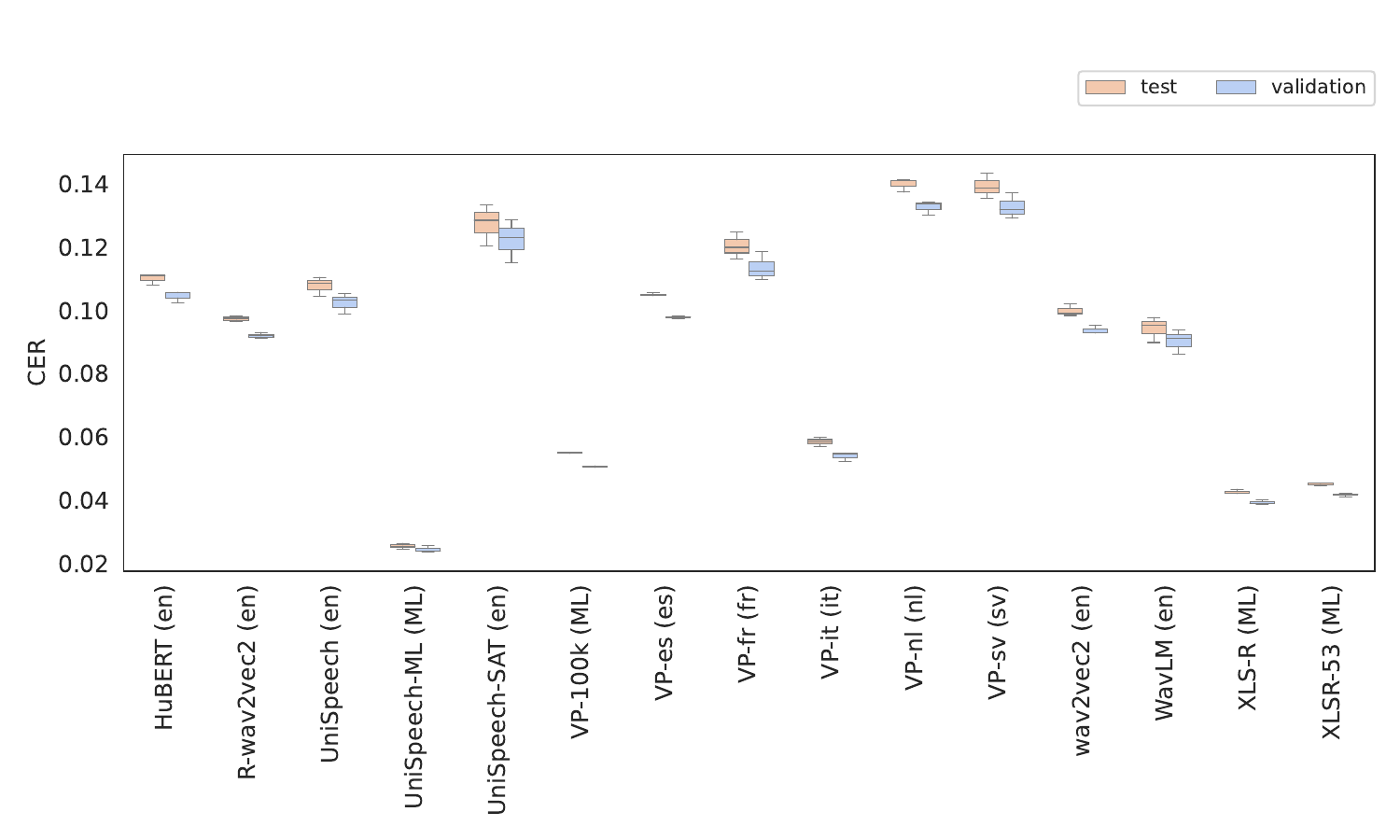}
  \caption{Overall performance over the pre-trained models for \textbf{Italian} language}
  \label{fig:f_overall_it}
\end{figure*}

\begin{figure*}
  \centering
  \includegraphics[width=\textwidth]{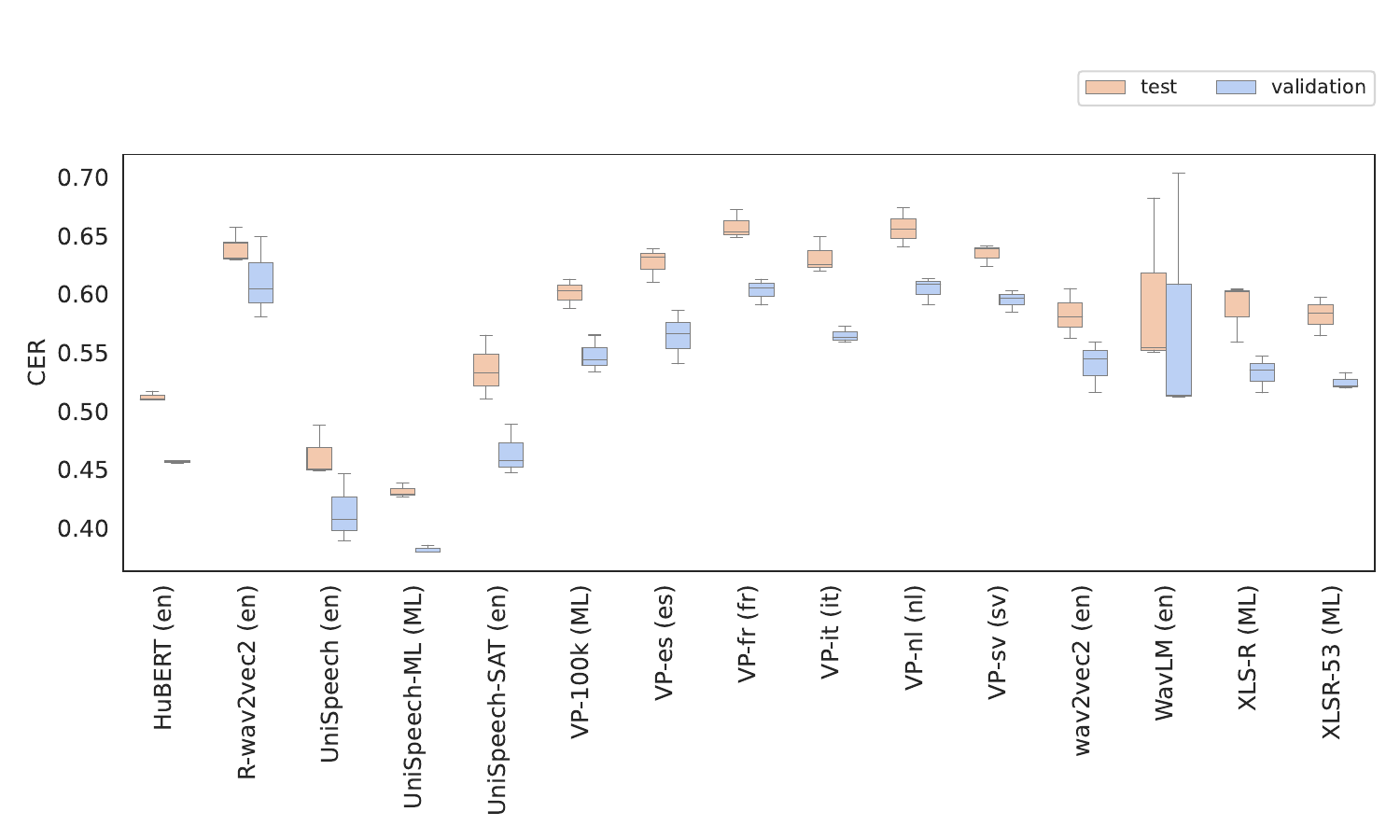}
  \caption{Overall performance over the pre-trained models for \textbf{Japanese} language}
  \label{fig:f_overall_ja}
\end{figure*}

\begin{figure*}
  \centering
  \includegraphics[width=\textwidth]{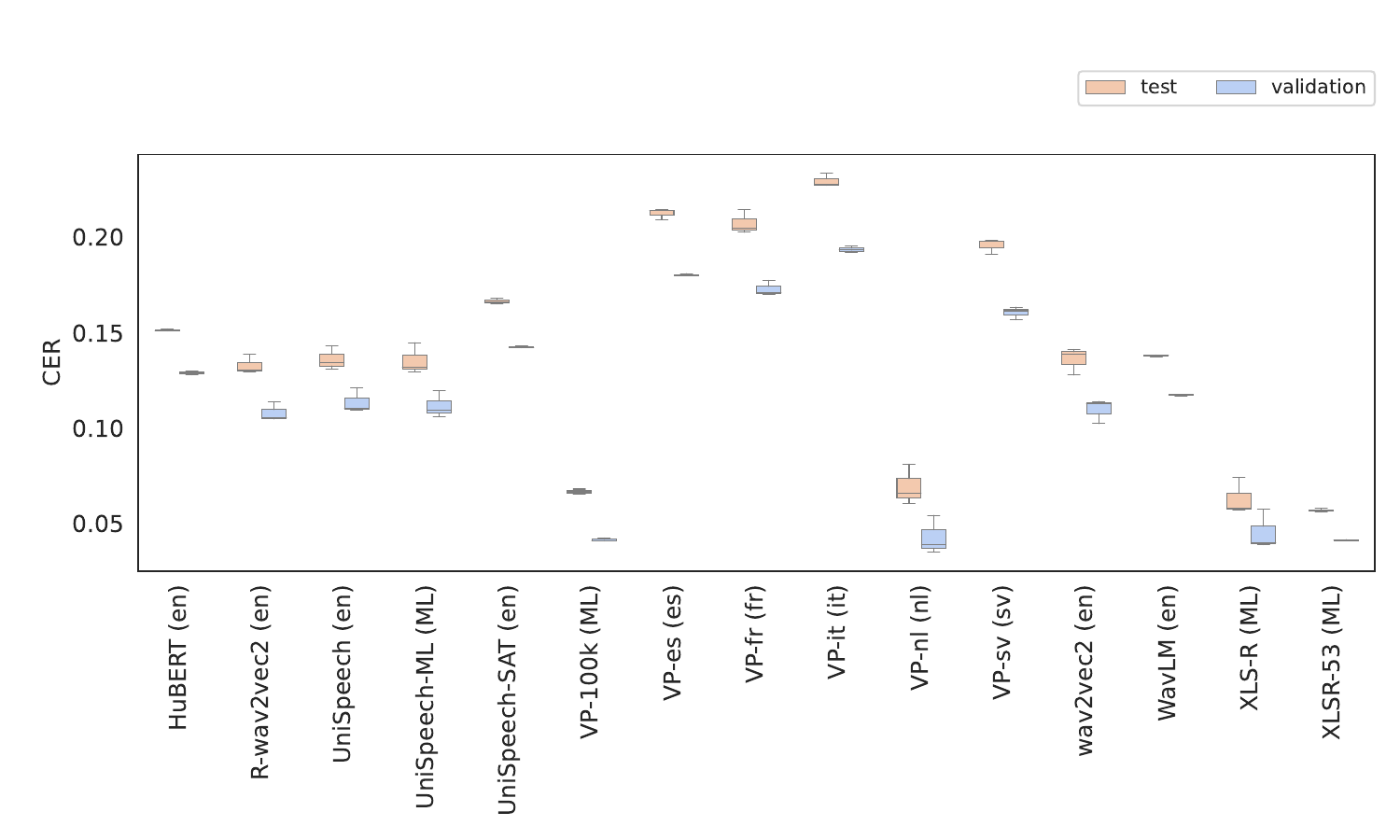}
  \caption{Overall performance over the pre-trained models for \textbf{Dutch} language}
  \label{fig:f_overall_nl}
\end{figure*}

\begin{figure*}
  \centering
  \includegraphics[width=\textwidth]{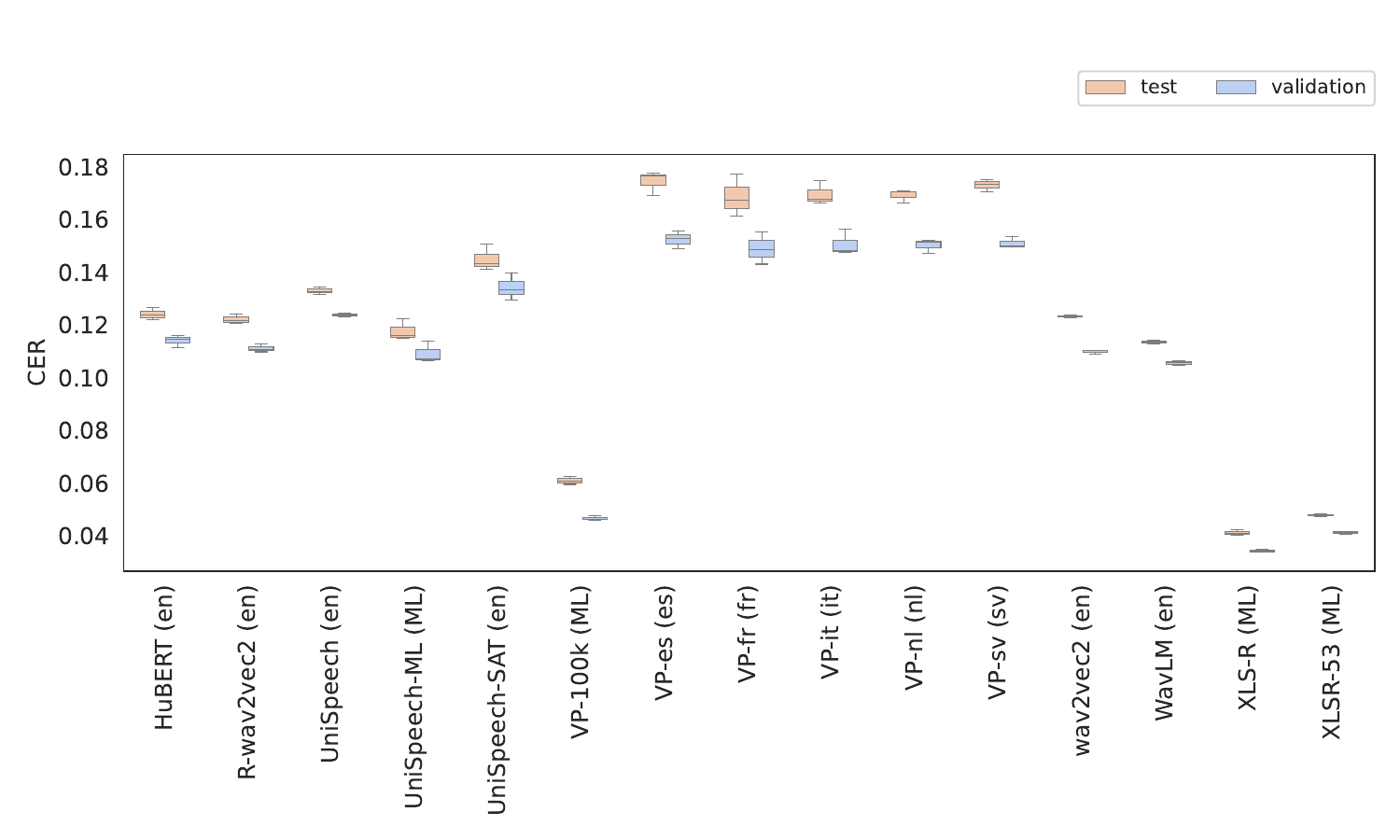}
  \caption{Overall performance over the pre-trained models for \textbf{Polish} language}
  \label{fig:f_overall_pl}
\end{figure*}

\begin{figure*}
  \centering
  \includegraphics[width=\textwidth]{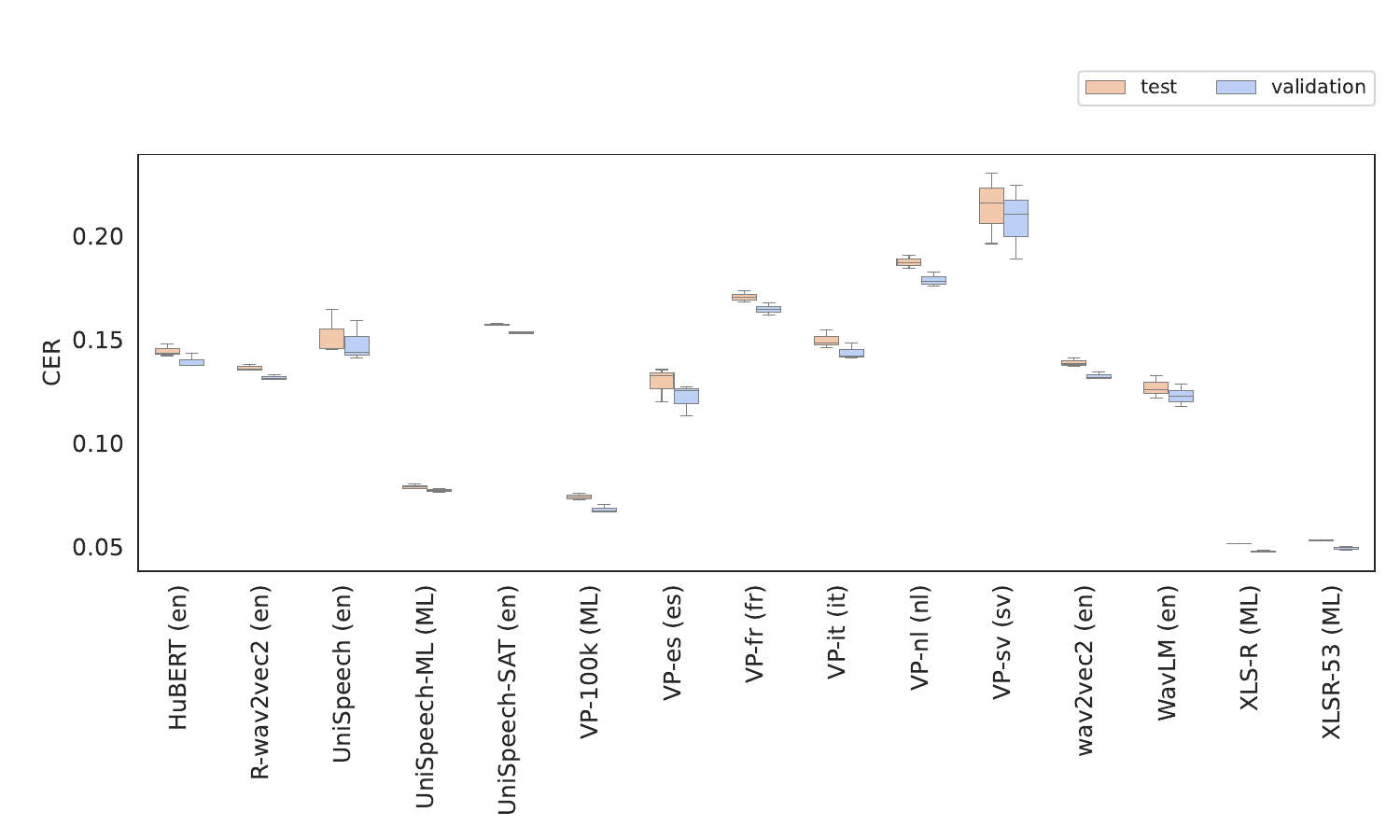}
  \caption{Overall performance over the pre-trained models for \textbf{Portuguese} language}
  \label{fig:f_overall_pt}
\end{figure*}

\begin{figure*}
  \centering
  \includegraphics[width=\textwidth]{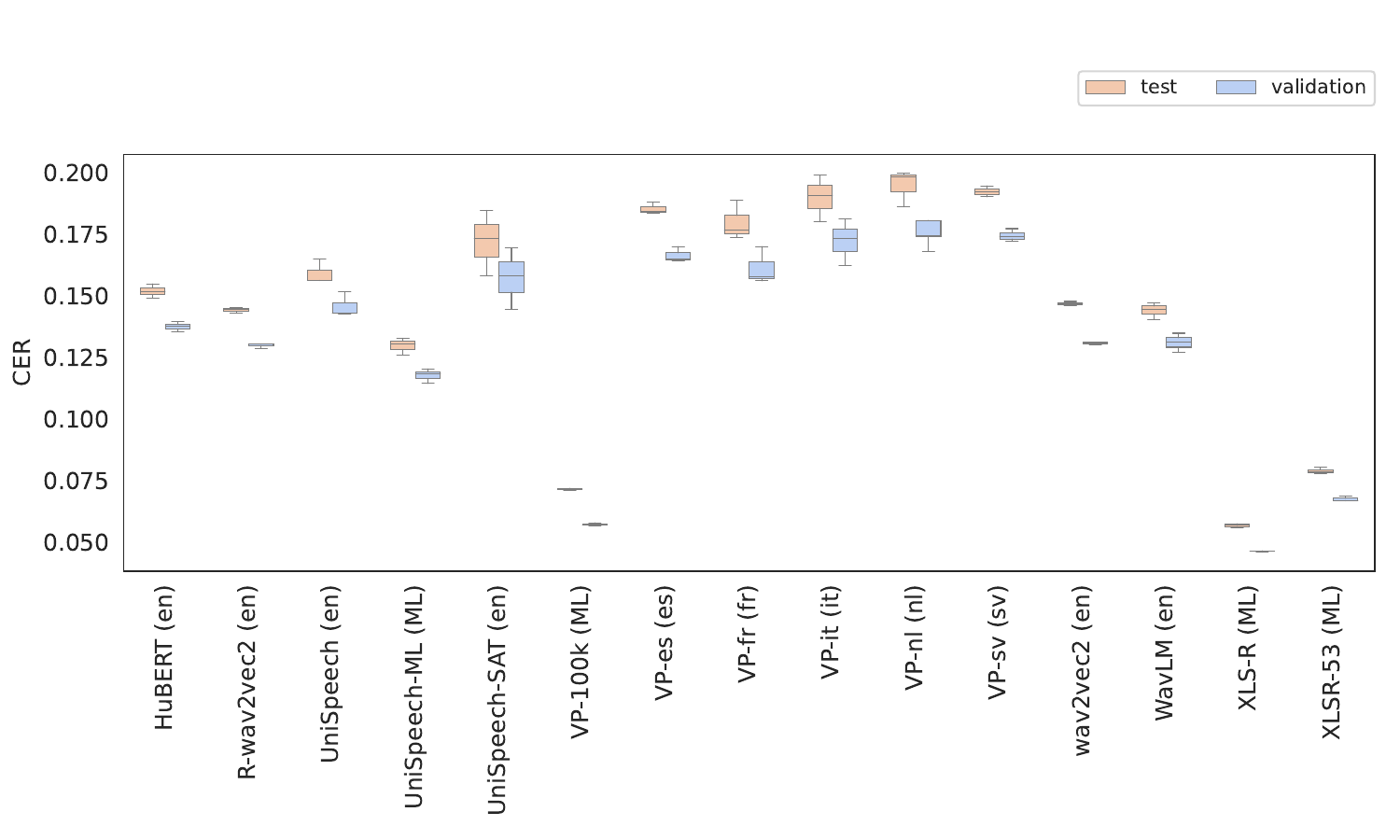}
  \caption{Overall performance over the pre-trained models for \textbf{Russian} language}
  \label{fig:f_overall_ru}
\end{figure*}

\begin{figure*}
  \centering
  \includegraphics[width=\textwidth]{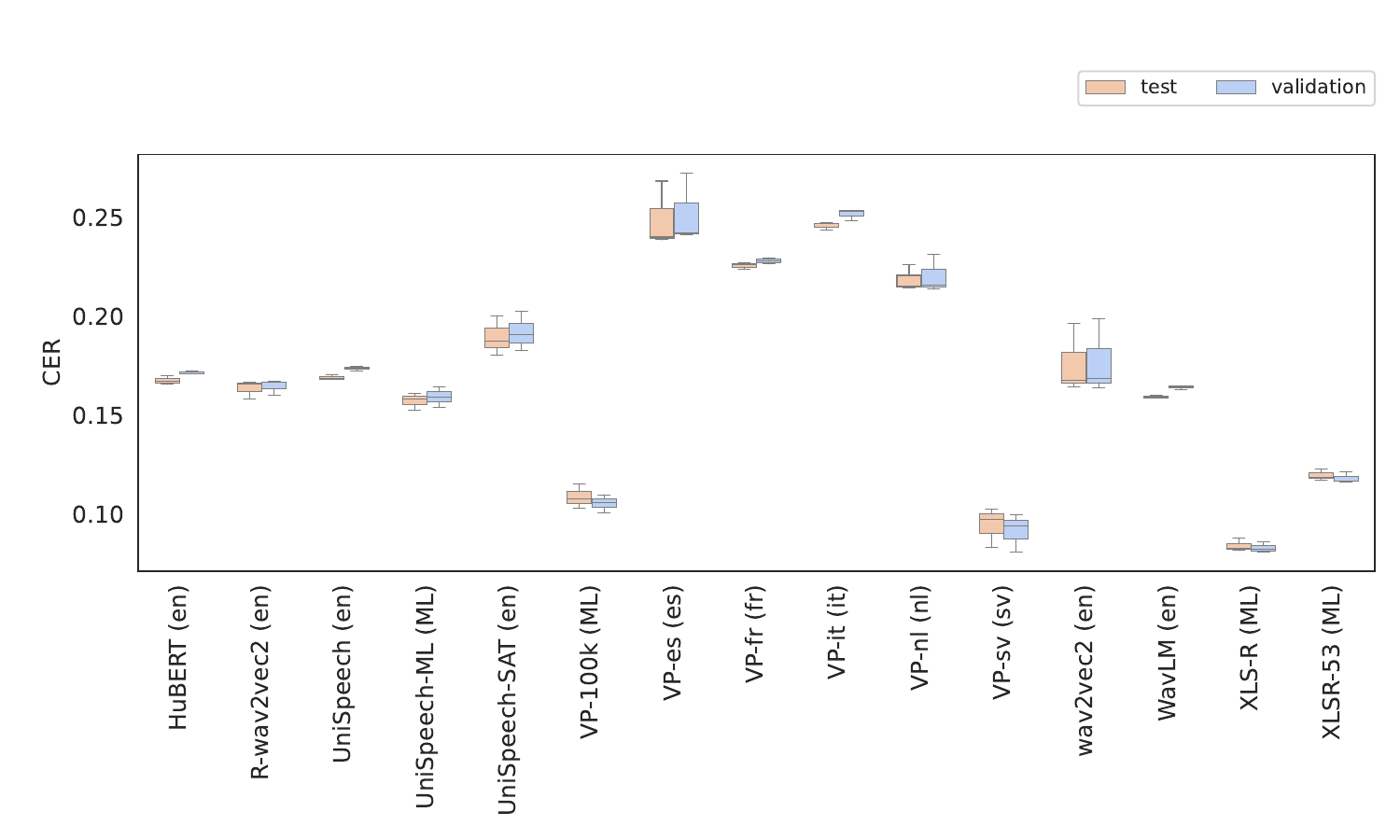}
  \caption{Overall performance over the pre-trained models for \textbf{Swedish} language}
  \label{fig:f_overall_sv}
\end{figure*}

\begin{figure*}
  \centering
  \includegraphics[width=\textwidth]{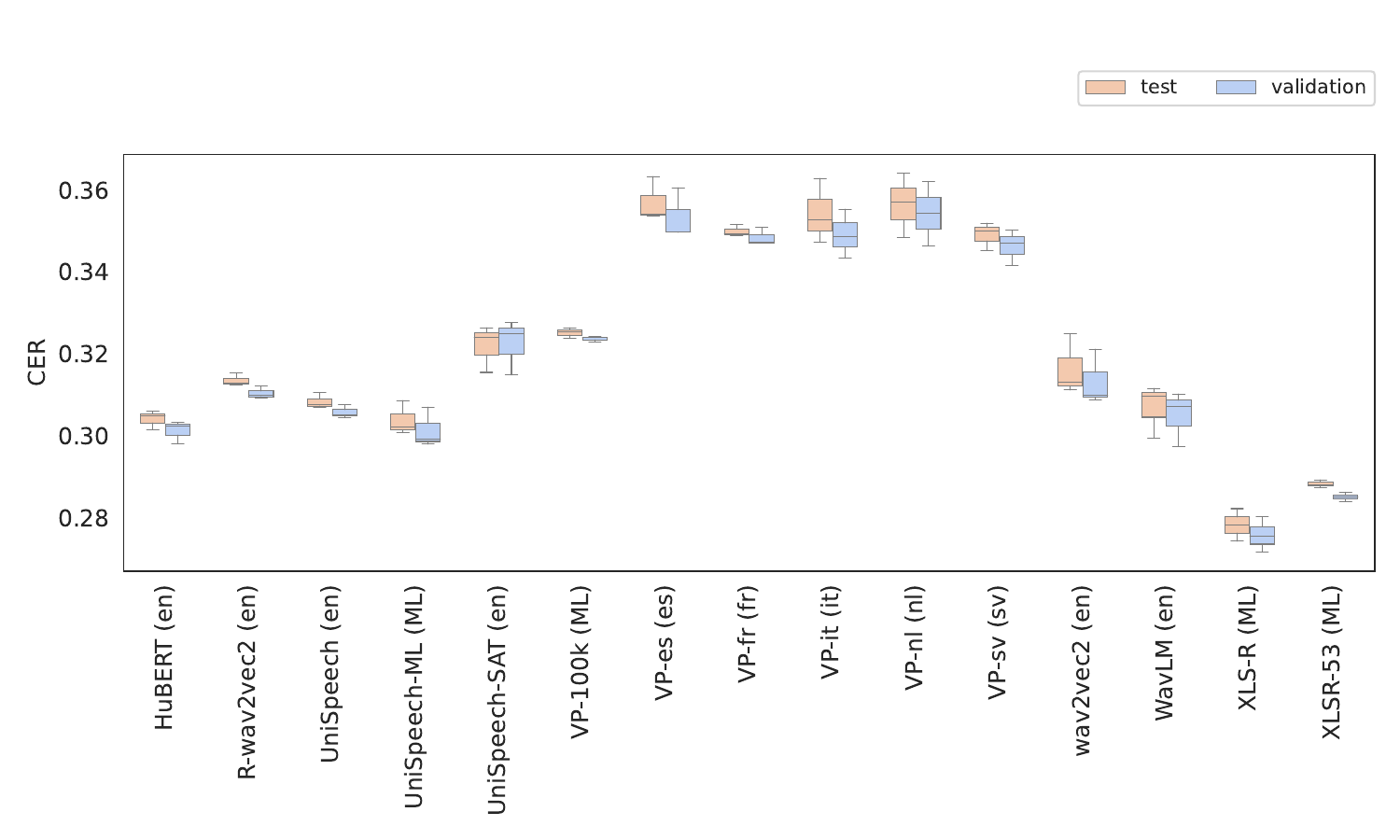}
  \caption{Overall performance over the pre-trained models for \textbf{Thai} language}
  \label{fig:f_overall_th}
\end{figure*}

\begin{figure*}
  \centering
  \includegraphics[width=\textwidth]{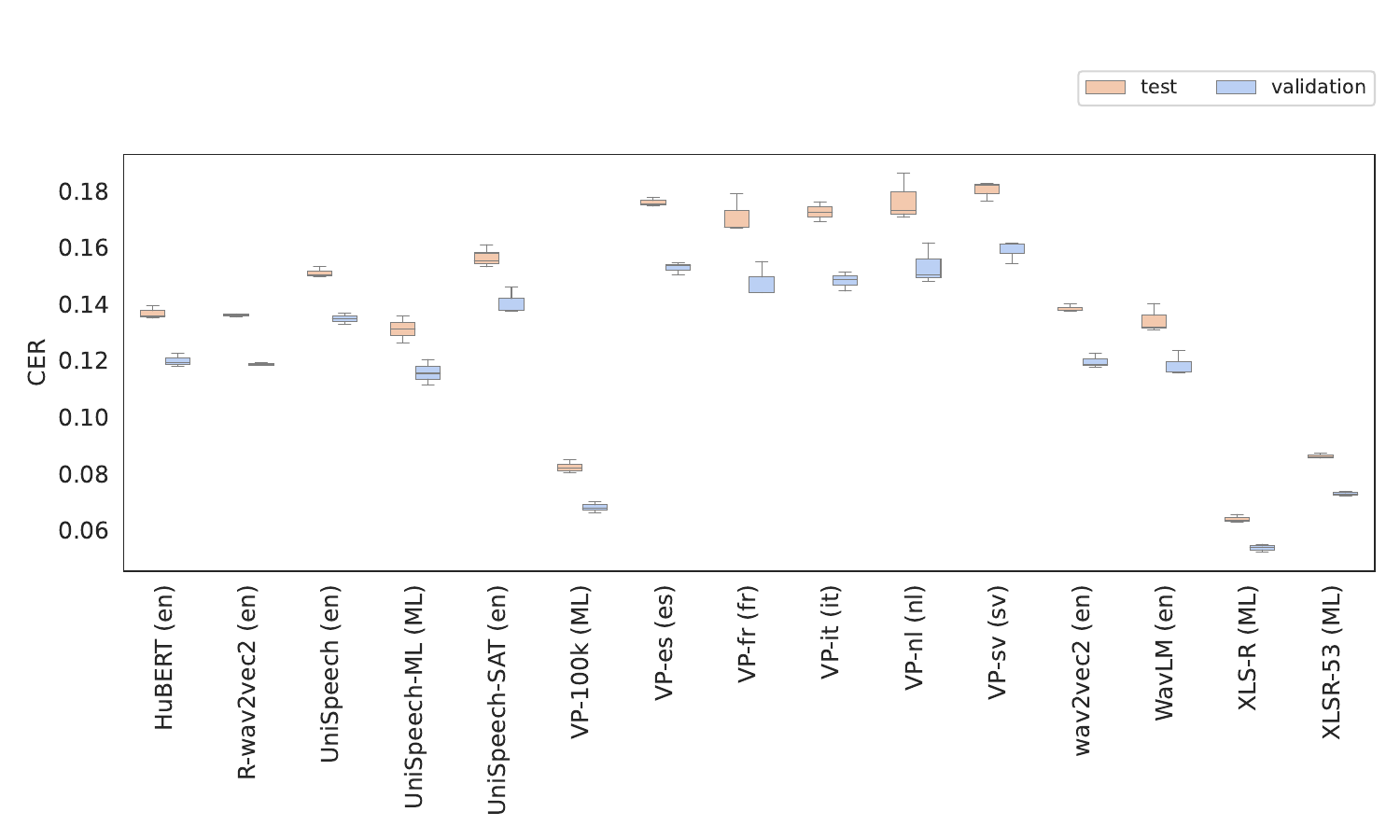}
  \caption{Overall performance over the pre-trained models for \textbf{Ukrainian} language}
  \label{fig:f_overall_uk}
\end{figure*}

\begin{figure*}
  \centering
  \includegraphics[width=\textwidth]{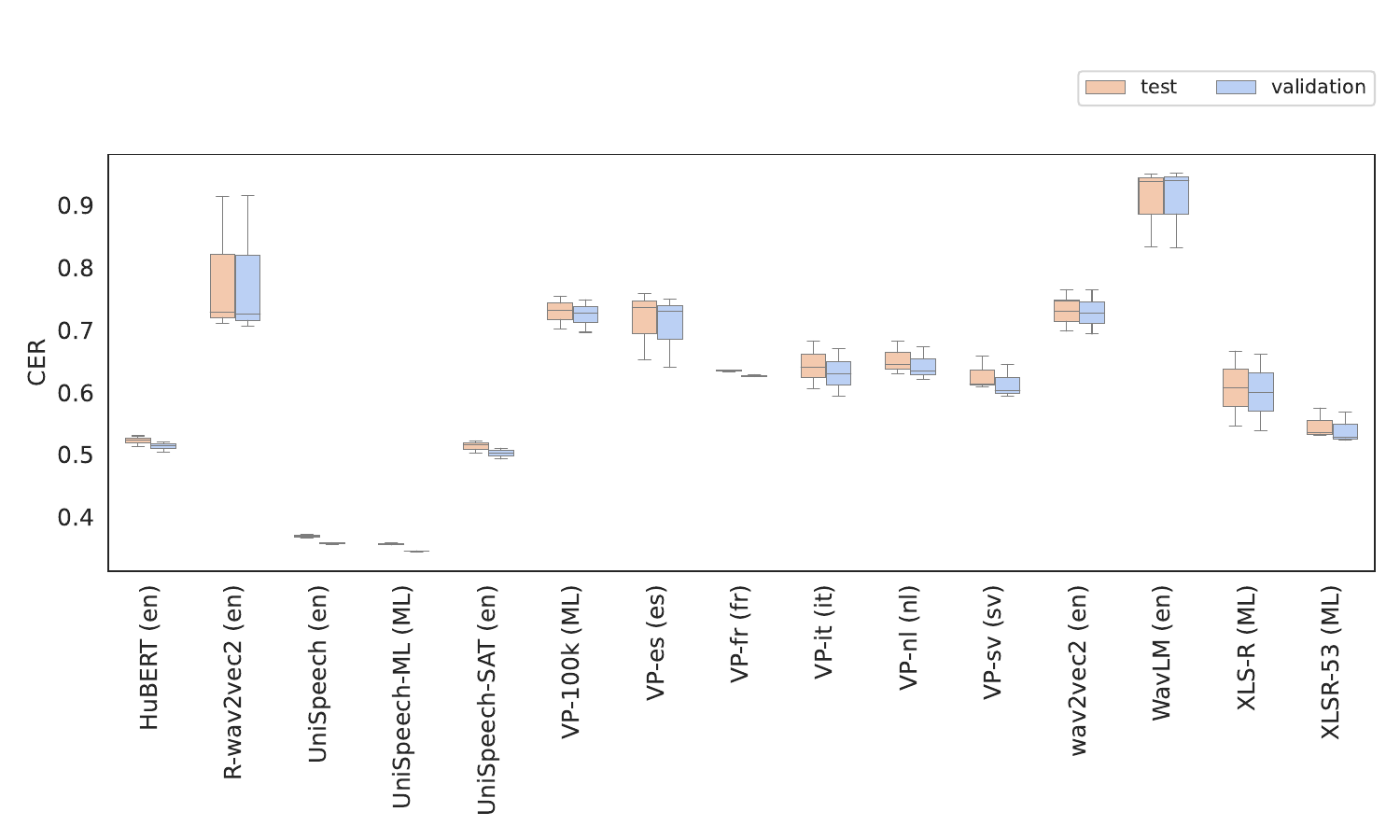}
  \caption{Overall performance over the pre-trained models for \textbf{Chinese} language}
  \label{fig:f_overall_zh}
\end{figure*}

\section{Multilingual models performance (languages)}
\label{sec:appendix-rq1-results}

This section contains tables and charts of the multilingual models' performance over the languages. The average CER for the multilingual models and the p-values of the statistical tests may be found in Table~\ref{tab:multilingual_overall_performance}. The ranking of the models for each language is presented in Table~\ref{tab:multilingual_performance_ranking}. Table~\ref{tab:multilingual_performance_posthoc_test} shows the p-values obtained by the post-hoc tests of each pair of models.


\begin{table*}
    \caption{Average CER on multilingual fine-tuned models and the p-value of the hypothesis test on models' performance. Values of $p > 0.05$ are marked in bold-face.}
    \label{tab:multilingual_overall_performance}
    \small
    \centering
    \begin{tabular}{lrrrrr}
    \toprule
    & \textbf{US-ML} & \textbf{VP-100k} & \textbf{XLS-R} & \textbf{XLSR-53} & \textbf{p-value}\\
    \midrule
    \multicolumn{6}{c}{\textbf{Validation}}\\
    \midrule
    ar & 0.221 ± 0.005 & 0.254 ± 0.016 & 0.193 ± 0.005 & 0.206 ± 0.011 & 0.023\\
    de & 0.109 ± 0.004 & 0.069 ± 0.002 & 0.049 ± 0.001 & 0.048 ± 0.000 & 0.016\\
    en & 0.077 ± 0.001 & 0.164 ± 0.005 & 0.113 ± 0.002 & 0.107 ± 0.000 & 0.016\\
    es & 0.027 ± 0.004 & 0.057 ± 0.001 & 0.041 ± 0.001 & 0.043 ± 0.000 & 0.016\\
    et & 0.135 ± 0.004 & 0.063 ± 0.000 & 0.069 ± 0.002 & 0.098 ± 0.001 & 0.016\\
    fa & 0.140 ± 0.002 & 0.151 ± 0.010 & 0.097 ± 0.010 & 0.092 ± 0.000 & 0.024\\
    fr & 0.070 ± 0.010 & 0.107 ± 0.001 & 0.083 ± 0.001 & 0.082 ± 0.000 & 0.019\\
    id & 0.102 ± 0.004 & 0.133 ± 0.007 & 0.081 ± 0.001 & 0.088 ± 0.002 & 0.016\\
    it & 0.024 ± 0.001 & 0.051 ± 0.000 & 0.039 ± 0.001 & 0.042 ± 0.000 & 0.016\\
    ja & 0.382 ± 0.003 & 0.548 ± 0.016 & 0.533 ± 0.016 & 0.525 ± 0.007 & 0.044\\
    nl & 0.112 ± 0.007 & 0.041 ± 0.001 & 0.045 ± 0.010 & 0.041 ± 0.000 & \textbf{0.086}\\
    pl & 0.109 ± 0.004 & 0.047 ± 0.001 & 0.034 ± 0.001 & 0.041 ± 0.001 & 0.016\\
    pt & 0.077 ± 0.001 & 0.068 ± 0.002 & 0.048 ± 0.000 & 0.049 ± 0.001 & 0.016\\
    ru & 0.118 ± 0.003 & 0.057 ± 0.001 & 0.046 ± 0.000 & 0.067 ± 0.001 & 0.016\\
    sv & 0.159 ± 0.005 & 0.106 ± 0.005 & 0.083 ± 0.003 & 0.118 ± 0.003 & 0.016\\
    th & 0.301 ± 0.005 & 0.324 ± 0.001 & 0.276 ± 0.004 & 0.285 ± 0.001 & 0.016\\
    uk & 0.116 ± 0.004 & 0.068 ± 0.002 & 0.054 ± 0.001 & 0.073 ± 0.001 & 0.016\\
    zh & 0.346 ± 0.001 & 0.725 ± 0.026 & 0.601 ± 0.061 & 0.540 ± 0.025 & 0.019\\
    \midrule
    \multicolumn{6}{c}{\textbf{Test}}\\
    \midrule
    ar & 0.221 ± 0.008 & 0.258 ± 0.016 & 0.195 ± 0.007 & 0.211 ± 0.018 & 0.038\\
    de & 0.119 ± 0.004 & 0.078 ± 0.002 & 0.056 ± 0.002 & 0.054 ± 0.000 & 0.016\\
    en & 0.089 ± 0.001 & 0.204 ± 0.006 & 0.143 ± 0.002 & 0.136 ± 0.001 & 0.016\\
    es & 0.029 ± 0.004 & 0.065 ± 0.001 & 0.047 ± 0.001 & 0.050 ± 0.000 & 0.016\\
    et & 0.123 ± 0.004 & 0.058 ± 0.001 & 0.059 ± 0.002 & 0.086 ± 0.001 & 0.025\\
    fa & 0.167 ± 0.003 & 0.178 ± 0.007 & 0.119 ± 0.007 & 0.115 ± 0.001 & 0.025\\
    fr & 0.075 ± 0.010 & 0.120 ± 0.000 & 0.093 ± 0.001 & 0.093 ± 0.000 & 0.024\\
    id & 0.109 ± 0.004 & 0.128 ± 0.006 & 0.092 ± 0.001 & 0.097 ± 0.002 & 0.016\\
    it & 0.025 ± 0.001 & 0.055 ± 0.000 & 0.043 ± 0.001 & 0.045 ± 0.000 & 0.015\\
    ja & 0.432 ± 0.006 & 0.601 ± 0.013 & 0.589 ± 0.026 & 0.582 ± 0.017 & \textbf{0.057}\\
    nl & 0.135 ± 0.008 & 0.067 ± 0.001 & 0.063 ± 0.010 & 0.057 ± 0.001 & 0.031\\
    pl & 0.118 ± 0.004 & 0.061 ± 0.001 & 0.041 ± 0.001 & 0.048 ± 0.001 & 0.016\\
    pt & 0.079 ± 0.001 & 0.074 ± 0.002 & 0.051 ± 0.000 & 0.053 ± 0.000 & 0.016\\
    ru & 0.130 ± 0.003 & 0.071 ± 0.000 & 0.057 ± 0.001 & 0.079 ± 0.001 & 0.016\\
    sv & 0.157 ± 0.004 & 0.109 ± 0.006 & 0.084 ± 0.003 & 0.120 ± 0.003 & 0.016\\
    th & 0.304 ± 0.004 & 0.325 ± 0.001 & 0.278 ± 0.004 & 0.288 ± 0.001 & 0.016\\
    uk & 0.131 ± 0.005 & 0.082 ± 0.002 & 0.064 ± 0.001 & 0.086 ± 0.001 & 0.016\\
    zh & 0.357 ± 0.002 & 0.730 ± 0.026 & 0.607 ± 0.060 & 0.547 ± 0.024 & 0.019\\
    \bottomrule
    \end{tabular}
\end{table*}

\begin{table*}
    \caption{Multilingual fine-tuned models ranking based on the average CER}
    \label{tab:multilingual_performance_ranking}
    \small
    \centering
    \begin{tabular}{lllll}
    \toprule
    & \textbf{1st} & \textbf{2nd} & \textbf{3rd} & \textbf{4th}\\
    \midrule
    \multicolumn{5}{c}{\textbf{Validation}}\\
    \midrule
    ar & XLS-R & XLSR-53 & UniSpeech-ML & VP-100k\\
    de & XLSR-53 & XLS-R & VP-100k & UniSpeech-ML\\
    en & UniSpeech-ML & XLSR-53 & XLS-R & VP-100k\\
    es & UniSpeech-ML & XLS-R & XLSR-53 & VP-100k\\
    et & VP-100k & XLS-R & XLSR-53 & UniSpeech-ML\\
    fa & XLSR-53 & XLS-R & UniSpeech-ML & VP-100k\\
    fr & UniSpeech-ML & XLSR-53 & XLS-R & VP-100k\\
    id & XLS-R & XLSR-53 & UniSpeech-ML & VP-100k\\
    it & UniSpeech-ML & XLS-R & XLSR-53 & VP-100k\\
    ja & UniSpeech-ML & XLSR-53 & XLS-R & VP-100k\\
    nl & VP-100k & XLSR-53 & XLS-R & UniSpeech-ML\\
    pl & XLS-R & XLSR-53 & VP-100k & UniSpeech-ML\\
    pt & XLS-R & XLSR-53 & VP-100k & UniSpeech-ML\\
    ru & XLS-R & VP-100k & XLSR-53 & UniSpeech-ML\\
    sv & XLS-R & VP-100k & XLSR-53 & UniSpeech-ML\\
    th & XLS-R & XLSR-53 & UniSpeech-ML & VP-100k\\
    uk & XLS-R & VP-100k & XLSR-53 & UniSpeech-ML\\
    zh & UniSpeech-ML & XLSR-53 & XLS-R & VP-100k\\
    \midrule
    \multicolumn{5}{c}{\textbf{Test}}\\
    \midrule
    ar & XLS-R & XLSR-53 & UniSpeech-ML & VP-100k\\
    de & XLSR-53 & XLS-R & VP-100k & UniSpeech-ML\\
    en & UniSpeech-ML & XLSR-53 & XLS-R & VP-100k\\
    es & UniSpeech-ML & XLS-R & XLSR-53 & VP-100k\\
    et & VP-100k & XLS-R & XLSR-53 & UniSpeech-ML\\
    fa & XLSR-53 & XLS-R & UniSpeech-ML & VP-100k\\
    fr & UniSpeech-ML & XLS-R & XLSR-53 & VP-100k\\
    id & XLS-R & XLSR-53 & UniSpeech-ML & VP-100k\\
    it & UniSpeech-ML & XLS-R & XLSR-53 & VP-100k\\
    ja & UniSpeech-ML & XLSR-53 & XLS-R & VP-100k\\
    nl & XLSR-53 & XLS-R & VP-100k & UniSpeech-ML\\
    pl & XLS-R & XLSR-53 & VP-100k & UniSpeech-ML\\
    pt & XLS-R & XLSR-53 & VP-100k & UniSpeech-ML\\
    ru & XLS-R & VP-100k & XLSR-53 & UniSpeech-ML\\
    sv & XLS-R & VP-100k & XLSR-53 & UniSpeech-ML\\
    th & XLS-R & XLSR-53 & UniSpeech-ML & VP-100k\\
    uk & XLS-R & VP-100k & XLSR-53 & UniSpeech-ML\\
    zh & UniSpeech-ML & XLSR-53 & XLS-R & VP-100k\\
    \bottomrule
    \end{tabular}
\end{table*}

\begin{table*}
    \caption{The p-value of post-hoc hypothesis test on multilingual models' performance. Values of $p > 0.05$ are marked in bold-face.}
    \label{tab:multilingual_performance_posthoc_test}
    \centering
    \small
    \begin{tabular}{p{8mm}p{15mm}p{15mm}p{15mm}p{15mm}p{15mm}p{15mm}}
    \toprule
    & \textbf{US-ML} & \textbf{US-ML} & \textbf{US-ML} & \textbf{VP-100k} & \textbf{VP-100k} & \textbf{XLS-R}\\
    & \textbf{VP-100k} & \textbf{XLS-R} & \textbf{XLSR-53} & \textbf{XLS-R} & \textbf{XLSR-53} & \textbf{XLSR-53}\\
    \midrule
    \multicolumn{7}{c}{\textbf{Validation}}\\
    \midrule
    ar & \textbf{0.091} & 0.012 & \textbf{0.138} & 0.001 & 0.007 & \textbf{0.138}\\
    de & 0.019 & 0.000 & 0.000 & 0.019 & 0.000 & 0.019\\
    en & 0.000 & 0.000 & 0.019 & 0.019 & 0.000 & 0.019\\
    es & 0.000 & 0.019 & 0.000 & 0.000 & 0.019 & 0.019\\
    et & 0.000 & 0.000 & 0.019 & 0.019 & 0.000 & 0.019\\
    fa & \textbf{0.097} & 0.019 & 0.044 & 0.002 & 0.003 & \textbf{0.461}\\
    fr & 0.000 & 0.003 & 0.040 & 0.040 & 0.003 & \textbf{0.058}\\
    id & 0.019 & 0.000 & 0.019 & 0.000 & 0.000 & 0.019\\
    it & 0.000 & 0.019 & 0.000 & 0.000 & 0.019 & 0.019\\
    ja & 0.012 & 0.049 & \textbf{0.219} & \textbf{0.587} & \textbf{0.219} & \textbf{0.587}\\
    nl & \textbf{0.160} & \textbf{0.075} & \textbf{0.162} & \textbf{1.000} & \textbf{1.000} & \textbf{1.000}\\
    pl & 0.019 & 0.000 & 0.000 & 0.000 & 0.019 & 0.019\\
    pt & 0.019 & 0.000 & 0.000 & 0.000 & 0.019 & 0.019\\
    ru & 0.000 & 0.000 & 0.019 & 0.019 & 0.019 & 0.000\\
    sv & 0.000 & 0.000 & 0.019 & 0.019 & 0.019 & 0.000\\
    th & 0.019 & 0.000 & 0.019 & 0.000 & 0.000 & 0.019\\
    uk & 0.000 & 0.000 & 0.019 & 0.019 & 0.019 & 0.000\\
    zh & 0.000 & 0.003 & 0.040 & 0.040 & 0.003 & \textbf{0.058}\\
    \midrule
    \multicolumn{7}{c}{\textbf{Test}}\\
    \midrule
    ar & \textbf{0.130} & \textbf{0.126} & \textbf{0.347} & 0.006 & 0.047 & \textbf{0.297}\\
    de & 0.019 & 0.000 & 0.000 & 0.019 & 0.000 & 0.019\\
    en & 0.000 & 0.000 & 0.019 & 0.019 & 0.000 & 0.019\\
    es & 0.000 & 0.019 & 0.000 & 0.000 & 0.019 & 0.019\\
    et & 0.003 & 0.003 & \textbf{0.109} & \textbf{0.809} & 0.032 & 0.035\\
    fa & \textbf{0.109} & 0.035 & 0.032 & 0.003 & 0.003 & \textbf{0.809}\\
    fr & 0.001 & 0.044 & 0.024 & 0.024 & 0.044 & \textbf{0.461}\\
    id & 0.019 & 0.000 & 0.019 & 0.000 & 0.000 & 0.019\\
    it & 0.000 & 0.016 & 0.000 & 0.000 & 0.016 & 0.016\\
    ja & 0.026 & \textbf{0.075} & \textbf{0.225} & \textbf{0.832} & \textbf{0.374} & \textbf{0.832}\\
    nl & \textbf{0.088} & 0.038 & 0.003 & \textbf{0.403} & \textbf{0.059} & \textbf{0.116}\\
    pl & 0.019 & 0.000 & 0.000 & 0.000 & 0.019 & 0.019\\
    pt & 0.019 & 0.000 & 0.000 & 0.000 & 0.019 & 0.019\\
    ru & 0.000 & 0.000 & 0.019 & 0.019 & 0.019 & 0.000\\
    sv & 0.000 & 0.000 & 0.019 & 0.019 & 0.019 & 0.000\\
    th & 0.019 & 0.000 & 0.019 & 0.000 & 0.000 & 0.019\\
    uk & 0.000 & 0.000 & 0.019 & 0.019 & 0.019 & 0.000\\
    zh & 0.000 & 0.003 & 0.040 & 0.040 & 0.003 & \textbf{0.058}\\
    \bottomrule
    \end{tabular}
\end{table*}

\section{Multilingual models performance (language families)}
\label{sec:appendix-rq2-results}

\begin{figure*}
  \centering
  \includegraphics[width=160mm]{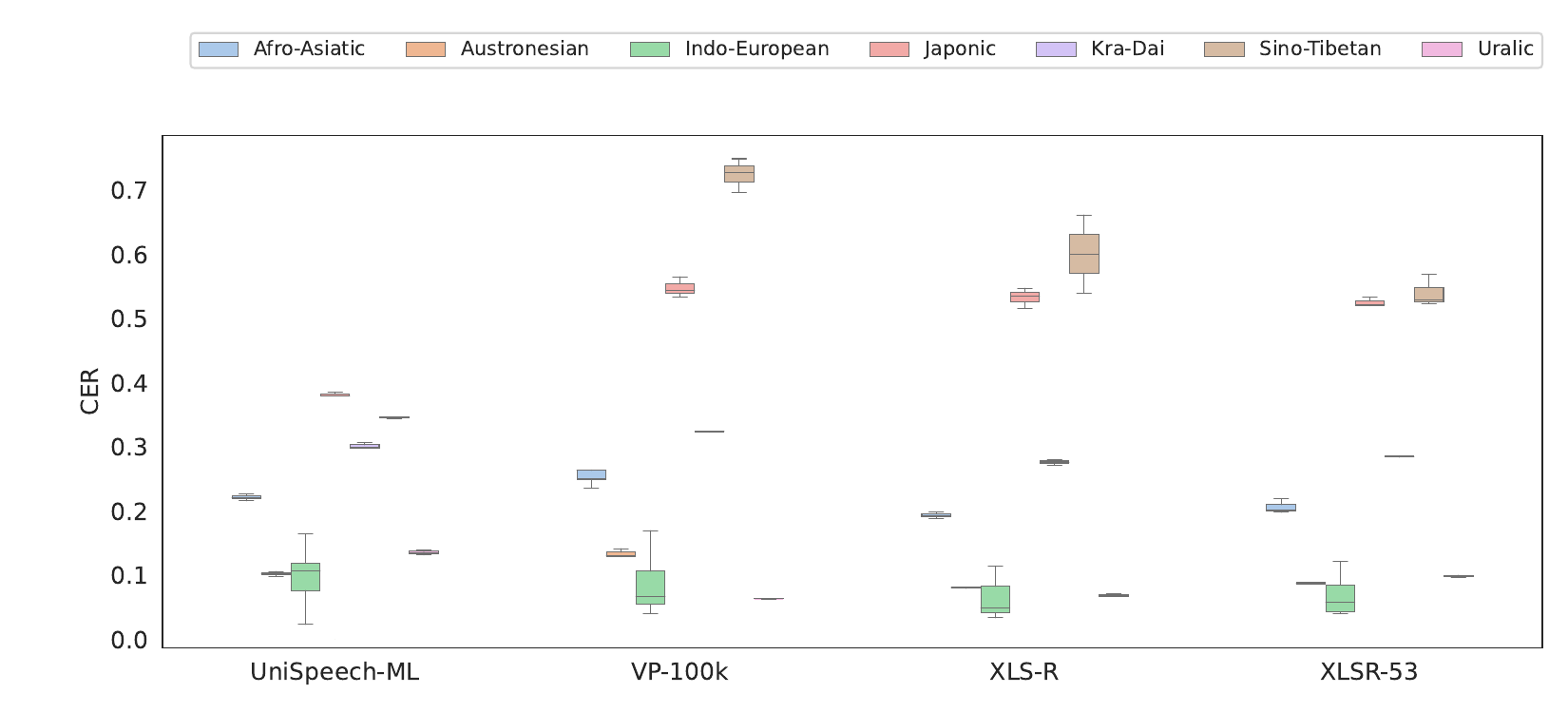}
  \caption{Multilingual pre-trained models performance(CER) over language families on the \textbf{validation} set}
  \label{fig:f_multilingual_family_validation}
\end{figure*}

\begin{figure*}
  \centering
  \includegraphics[width=160mm]{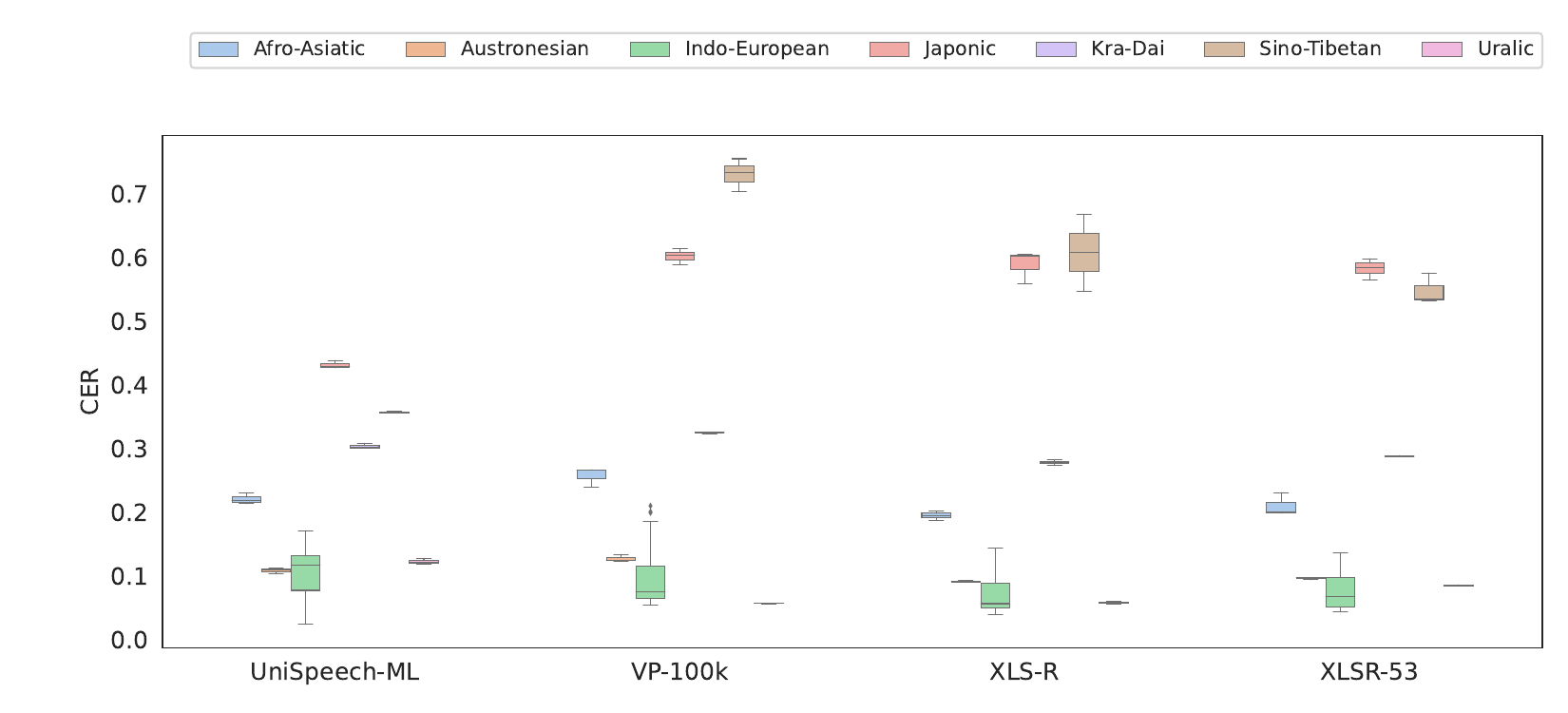}
  \caption{Multilingual pre-trained models performance(CER) over language families on the \textbf{test} set}
  \label{fig:f_multilingual_family_test}
\end{figure*}

This section contains tables and charts of the multilingual models performance over the language families. The average CER for the multilingual models and the p-values of the statistical tests may be found in Tables~\ref{tab:multilingual_overall_performance_grouped_family_T} and~\ref{tab:multilingual_overall_performance_grouped_family}. The ranking of the models for each language family group is presented in Table~\ref{tab:multilingual_performance_ranking_grouped_family}. Table~\ref{tab:multilingual_performance_posthoc_test_indo} shows the p-values obtained by the post-hoc tests of each pair of models. To explore the diversity of language families in the non-Indo-European group, we describe the basic statistics of language families in Table~\ref{tab:multilingual_overall_performance_family}, Figure~\ref{fig:f_multilingual_family_validation} and Figure~\ref{fig:f_multilingual_family_test}


\begin{table*}
    \caption{Average CER plus or minus the standard deviation on multilingual fine-tuned models and the p-value of hypothesis test on models' performance on Indo-European and non-Indo-European language family}
    \label{tab:multilingual_overall_performance_grouped_family_T}
    \small
    \centering
    \begin{tabular}{p{25mm}p{35mm}p{35mm}p{12mm}}
    \toprule
    & \textbf{Indo-European} & \textbf{non-Indo-European} & \textbf{p-value}\\
    \midrule
    \multicolumn{4}{c}{\textbf{Validation}}\\
    \midrule
    UniSpeech-ML & 0.095 ± 0.040 & 0.248 ± 0.107 & 0.000\\
    VP-100k & 0.082 ± 0.040 & 0.341 ± 0.237 & 0.000\\
    XLS-R & 0.061 ± 0.025 & 0.292 ± 0.214 & 0.000\\
    XLSR-53 & 0.067 ± 0.027 & 0.290 ± 0.189 & 0.000\\
    \midrule
    \multicolumn{4}{c}{\textbf{Test}}\\
    \midrule
    UniSpeech-ML & 0.105 ± 0.045 & 0.258 ± 0.122 & 0.000\\
    VP-100k & 0.097 ± 0.047 & 0.350 ± 0.249 & 0.000\\
    XLS-R & 0.072 ± 0.031 & 0.303 ± 0.228 & 0.000\\
    XLSR-53 & 0.078 ± 0.031 & 0.302 ± 0.205 & 0.000\\
    \bottomrule
    \end{tabular}
\end{table*}

\begin{table*}
    \caption{Average CER plus or minus the standard deviation on multilingual fine-tuned models and the p-value of hypothesis test on models' performance on Indo-European and non-Indo-European language family (\textbf{without Chinese and Japanese languages})}
    \label{tab:multilingual_overall_performance_grouped_family_T_wt_ja_zh}
    \small
    \centering
    \begin{tabular}{p{25mm}p{35mm}p{35mm}p{12mm}}
    \toprule
    & \textbf{Indo-European} & \textbf{non-Indo-European} & \textbf{p-value}\\
    \midrule
    \multicolumn{4}{c}{\textbf{Validation}}\\
    \midrule
    UniSpeech-ML & 0.095 ± 0.040 & 0.190 ± 0.081 & 0.001\\
    VP-100k & 0.082 ± 0.040 & 0.194 ± 0.106 & 0.001\\
    XLS-R & 0.061 ± 0.025 & 0.155 ± 0.089 & 0.001\\
    XLSR-53 & 0.067 ± 0.027 & 0.169 ± 0.085 & 0.000\\
    \midrule
    \multicolumn{4}{c}{\textbf{Test}}\\
    \midrule
    UniSpeech-ML & 0.105 ± 0.045 & 0.189 ± 0.083 & 0.012\\
    VP-100k & 0.097 ± 0.047 & 0.192 ± 0.110 & 0.018\\
    XLS-R & 0.072 ± 0.031 & 0.156 ± 0.091 & 0.001\\
    XLSR-53 & 0.078 ± 0.031 & 0.170 ± 0.088 & 0.001\\
    \bottomrule
    \end{tabular}
\end{table*}

\begin{table*}
    \caption{Average CER plus or minus the standard deviation on multilingual fine-tuned models and the p-value of hypothesis test on performance between models on Indo-European and non-Indo-European language family groups. Values of $p > 0.05$ are marked in bold-face.}
    \label{tab:multilingual_overall_performance_grouped_family}
    \small
    \centering
    \begin{tabular}{p{30mm}p{20mm}p{20mm}p{20mm}p{20mm}p{12mm}}
    \toprule
    & \textbf{US-ML} & \textbf{VP-100k} & \textbf{XLS-R} & \textbf{XLSR-53} & \textbf{p-value}\\
    \midrule
    \multicolumn{6}{c}{\textbf{Validation}}\\
    \midrule
    Indo-European & 0.095 ± 0.040 & 0.082 ± 0.040 & 0.061 ± 0.025 & 0.067 ± 0.027 & 0.001\\
    non-Indo-European & 0.248 ± 0.107 & 0.341 ± 0.237 & 0.292 ± 0.214 & 0.290 ± 0.189 & \textbf{0.861}\\
    \midrule
    \multicolumn{6}{c}{\textbf{Test}}\\
    \midrule
    Indo-European & 0.105 ± 0.045 & 0.097 ± 0.047 & 0.072 ± 0.031 & 0.078 ± 0.031 & 0.001\\
    non-Indo-European & 0.258 ± 0.122 & 0.350 ± 0.249 & 0.303 ± 0.228 & 0.302 ± 0.205 & \textbf{0.807}\\
    \bottomrule
    \end{tabular}
\end{table*}

\begin{table*}
    \caption{Multilingual fine-tuned models ranking based on the average CER on Indo-European and non-Indo-European language family groups}
    \label{tab:multilingual_performance_ranking_grouped_family}
    \centering
    \small
    \begin{tabular}{p{30mm}p{20mm}p{20mm}p{20mm}p{20mm}}
    \toprule
    & \textbf{1st} & \textbf{2nd} & \textbf{3rd} & \textbf{4th}\\
    \midrule
    \multicolumn{5}{c}{\textbf{Validation}}\\
    \midrule
    Indo-European & XLS-R & XLSR-53 & VP-100k & US-ML\\
    non-Indo-European & US-ML & XLSR-53 & XLS-R & VP-100k\\
    \midrule
    \multicolumn{5}{c}{\textbf{Test}}\\
    \midrule
    Indo-European & XLS-R & XLSR-53 & VP-100k & US-ML\\
    non-Indo-European & US-ML & XLSR-53 & XLS-R & VP-100k\\
    \bottomrule
    \end{tabular}
\end{table*}

\begin{table*}
    \caption{The p-value of post-hoc Conover-Iman hypothesis test on models' performance on Indo-European languages. Values of $p > 0.05$ are marked in bold-face.}
    \label{tab:multilingual_performance_posthoc_test_indo}
    \centering
    \small
    \begin{tabular}{p{15mm}p{15mm}p{15mm}p{15mm}p{15mm}p{15mm}p{15mm}}
    \toprule
    \textbf{Split} & \textbf{US-ML} & \textbf{US-ML} & \textbf{US-ML} & \textbf{VP-100k} & \textbf{VP-100k} & \textbf{XLS-R}\\
    & \textbf{VP-100k} & \textbf{XLS-R} & \textbf{XLSR-53} & \textbf{XLS-R} & \textbf{XLSR-53} & \textbf{XLSR-53}\\
    \midrule
    Validation & \textbf{0.335} & 0.000 & 0.010 & 0.044 & \textbf{0.335} & \textbf{0.335}\\
    Test & \textbf{0.569} & 0.001 & 0.013 & 0.025 & \textbf{0.145} & \textbf{0.569}\\
    \bottomrule
    \end{tabular}
\end{table*}


\begin{table*}
    \caption{Average CER plus or minus the standard deviation on multilingual fine-tuned models performance grouped by language family}
    \label{tab:multilingual_overall_performance_family}
    \centering
    \small
    \begin{tabular}{p{30mm}p{20mm}p{20mm}p{20mm}p{20mm}}
    \toprule
    & \textbf{US-ML} & \textbf{VP-100k} & \textbf{XLS-R} & \textbf{XLSR-53}\\
    \midrule
    \multicolumn{5}{c}{\textbf{Validation}}\\
    \midrule
    Afro-Asiatic & 0.221 ± 0.005 & 0.254 ± 0.016 & 0.193 ± 0.005 & 0.206 ± 0.011\\
    Austronesian & 0.102 ± 0.004 & 0.133 ± 0.007 & 0.081 ± 0.001 & 0.088 ± 0.002\\
    Indo-European & 0.095 ± 0.040 & 0.082 ± 0.040 & 0.061 ± 0.025 & 0.067 ± 0.027\\
    Japonic & 0.382 ± 0.003 & 0.548 ± 0.016 & 0.533 ± 0.016 & 0.525 ± 0.007\\
    Kra-Dai & 0.301 ± 0.005 & 0.324 ± 0.001 & 0.276 ± 0.004 & 0.285 ± 0.001\\
    Sino-Tibetan & 0.346 ± 0.001 & 0.725 ± 0.026 & 0.601 ± 0.061 & 0.540 ± 0.025\\
    Uralic & 0.135 ± 0.004 & 0.063 ± 0.000 & 0.069 ± 0.002 & 0.098 ± 0.001\\
    \midrule
    \multicolumn{5}{c}{\textbf{Test}}\\
    \midrule
    Afro-Asiatic & 0.221 ± 0.008 & 0.258 ± 0.016 & 0.195 ± 0.007 & 0.211 ± 0.018\\
    Austronesian & 0.109 ± 0.004 & 0.128 ± 0.006 & 0.092 ± 0.001 & 0.097 ± 0.002\\
    Indo-European & 0.105 ± 0.045 & 0.097 ± 0.047 & 0.072 ± 0.031 & 0.078 ± 0.031\\
    Japonic & 0.432 ± 0.006 & 0.601 ± 0.013 & 0.589 ± 0.026 & 0.582 ± 0.017\\
    Kra-Dai & 0.304 ± 0.004 & 0.325 ± 0.001 & 0.278 ± 0.004 & 0.288 ± 0.001\\
    Sino-Tibetan & 0.357 ± 0.002 & 0.730 ± 0.026 & 0.607 ± 0.060 & 0.547 ± 0.024\\
    Uralic & 0.123 ± 0.004 & 0.058 ± 0.001 & 0.059 ± 0.002 & 0.086 ± 0.001\\
    \bottomrule
    \end{tabular}
\end{table*}

\section{Monolingual models performance}
\label{sec:appendix-rq3-results}

This section contains tables and charts to support our findings for RQ3. We divided these results into two parts: English pre-trained monolingual models, and VoxPopuli pre-trained monolingual models. All tables also present the non-pre-trained models (NP) as a performance baseline. Table~\ref{tab:en_monolingual_overall_performance} presents the average CER of English pre-trained monolingual models as well as Kruskal-Wallis p-values. A model ranking per language is presented in Table~\ref{tab:en_monolingual_performance_ranking}, with the addition of multilingual models for comparison. The results of Conover-Imam tests between pairs of models are shown in Table~\ref{tab:en_monolingual_performance_posthoc_test}, and a count of $p > 0.05$ per model-pair and language is presented in Table~\ref{tab:en_monolingual_performance_posthoc_test_frequency}.

As for the VoxPopuli data, Table~\ref{tab:vp_monolingual_overall_performance} presents the average CER for each model/language, with Kruskal-Wallis p-values as well. A frequency of model rankings is shown in Table~\ref{tab:vp_monolingual_performance_ranking_frequency}. The Conover-Imam p-value test results for each model pair and language is shown in Table~\ref{tab:vp_monolingual_performance_posthoc_test}, while the count of $p > 0.05$ for each model pair/language is shown in Table~\ref{tab:vp_monolingual_performance_posthoc_test_frequency}.



\begin{table*}
    \caption{Average CER on \textbf{English monolingual} and no pre-trained fine-tuned models and the p-value of the hypothesis test on models' performance.}
    \label{tab:en_monolingual_overall_performance}
    \centering
    \small
    \begin{tabular}{lllllllll}
    \toprule
    & \textbf{HuBERT} & \textbf{R-wav2vec2} & \textbf{UniSpeech} & \textbf{US-SAT} & \textbf{wav2vec2} & \textbf{WavLM} & \textbf{NP} & \textbf{p-value}\\
    \midrule
    \multicolumn{9}{c}{\textbf{Validation}}\\
    \midrule
    ar & 0.249 ± 0.008 & 0.242 ± 0.006 & 0.243 ± 0.006 & 0.257 ± 0.006 & 0.239 ± 0.002 & 0.238 ± 0.010 & 0.782 ± 0.058 & 0.024\\
    de & 0.125 ± 0.003 & 0.112 ± 0.001 & 0.127 ± 0.002 & 0.148 ± 0.003 & 0.118 ± 0.004 & 0.116 ± 0.003 & 0.749 ± 0.058 & 0.005\\
    en & 0.114 ± 0.016 & 0.086 ± 0.001 & 0.066 ± 0.003 & 0.116 ± 0.015 & 0.098 ± 0.005 & 0.081 ± 0.001 & 0.762 ± 0.009 & 0.004\\
    es & 0.118 ± 0.002 & 0.103 ± 0.001 & 0.109 ± 0.003 & 0.127 ± 0.008 & 0.104 ± 0.000 & 0.100 ± 0.007 & 0.762 ± 0.039 & 0.006\\
    et & 0.143 ± 0.004 & 0.133 ± 0.005 & 0.155 ± 0.002 & 0.161 ± 0.008 & 0.136 ± 0.003 & 0.138 ± 0.002 & 0.725 ± 0.015 & 0.007\\
    fa & 0.153 ± 0.002 & 0.147 ± 0.001 & 0.150 ± 0.006 & 0.173 ± 0.006 & 0.160 ± 0.010 & 0.142 ± 0.002 & 0.770 ± 0.046 & 0.006\\
    fr & 0.188 ± 0.009 & 0.175 ± 0.002 & 0.219 ± 0.046 & 0.216 ± 0.005 & 0.183 ± 0.012 & 0.168 ± 0.006 & 0.716 ± 0.010 & 0.008\\
    id & 0.139 ± 0.004 & 0.121 ± 0.002 & 0.121 ± 0.002 & 0.134 ± 0.002 & 0.137 ± 0.004 & 0.122 ± 0.002 & 0.807 ± 0.085 & 0.008\\
    it & 0.105 ± 0.002 & 0.092 ± 0.001 & 0.102 ± 0.003 & 0.122 ± 0.007 & 0.094 ± 0.001 & 0.090 ± 0.004 & 0.728 ± 0.034 & 0.005\\
    ja & 0.457 ± 0.001 & 0.612 ± 0.035 & 0.414 ± 0.029 & 0.465 ± 0.022 & 0.540 ± 0.022 & 0.576 ± 0.110 & 0.948 ± 0.006 & 0.005\\
    nl & 0.129 ± 0.001 & 0.108 ± 0.005 & 0.114 ± 0.007 & 0.142 ± 0.000 & 0.110 ± 0.006 & 0.117 ± 0.000 & 0.723 ± 0.020 & 0.007\\
    pl & 0.114 ± 0.002 & 0.111 ± 0.002 & 0.124 ± 0.001 & 0.134 ± 0.005 & 0.110 ± 0.001 & 0.106 ± 0.001 & 0.798 ± 0.012 & 0.004\\
    pt & 0.139 ± 0.003 & 0.132 ± 0.001 & 0.148 ± 0.010 & 0.153 ± 0.001 & 0.132 ± 0.002 & 0.123 ± 0.005 & 0.771 ± 0.009 & 0.004\\
    ru & 0.137 ± 0.002 & 0.130 ± 0.001 & 0.146 ± 0.005 & 0.157 ± 0.012 & 0.131 ± 0.001 & 0.131 ± 0.004 & 0.754 ± 0.025 & 0.006\\
    sv & 0.172 ± 0.001 & 0.165 ± 0.004 & 0.174 ± 0.001 & 0.192 ± 0.010 & 0.177 ± 0.019 & 0.164 ± 0.001 & 0.767 ± 0.015 & 0.012\\
    th & 0.301 ± 0.003 & 0.310 ± 0.002 & 0.306 ± 0.002 & 0.323 ± 0.007 & 0.313 ± 0.007 & 0.305 ± 0.007 & 0.876 ± 0.098 & 0.010\\
    uk & 0.120 ± 0.002 & 0.119 ± 0.000 & 0.135 ± 0.002 & 0.141 ± 0.005 & 0.120 ± 0.003 & 0.118 ± 0.004 & 0.770 ± 0.003 & 0.012\\
    zh & 0.514 ± 0.008 & 0.783 ± 0.116 & 0.358 ± 0.002 & 0.503 ± 0.009 & 0.728 ± 0.035 & 0.908 ± 0.066 & 0.989 ± 0.007 & 0.004\\
    \midrule
    \multicolumn{9}{c}{\textbf{Test}}\\
    \midrule
    ar & 0.247 ± 0.003 & 0.238 ± 0.010 & 0.247 ± 0.006 & 0.259 ± 0.006 & 0.244 ± 0.005 & 0.239 ± 0.008 & 0.792 ± 0.050 & 0.022\\
    de & 0.137 ± 0.002 & 0.124 ± 0.001 & 0.137 ± 0.002 & 0.158 ± 0.003 & 0.130 ± 0.005 & 0.126 ± 0.003 & 0.751 ± 0.056 & 0.006\\
    en & 0.143 ± 0.019 & 0.114 ± 0.001 & 0.075 ± 0.004 & 0.141 ± 0.016 & 0.129 ± 0.007 & 0.103 ± 0.001 & 0.773 ± 0.001 & 0.005\\
    es & 0.129 ± 0.002 & 0.113 ± 0.001 & 0.119 ± 0.003 & 0.136 ± 0.008 & 0.115 ± 0.001 & 0.110 ± 0.008 & 0.767 ± 0.034 & 0.007\\
    et & 0.129 ± 0.003 & 0.120 ± 0.005 & 0.142 ± 0.002 & 0.148 ± 0.007 & 0.124 ± 0.002 & 0.123 ± 0.002 & 0.741 ± 0.019 & 0.007\\
    fa & 0.178 ± 0.001 & 0.174 ± 0.002 & 0.177 ± 0.006 & 0.197 ± 0.006 & 0.186 ± 0.008 & 0.167 ± 0.003 & 0.769 ± 0.041 & 0.006\\
    fr & 0.203 ± 0.009 & 0.189 ± 0.002 & 0.232 ± 0.046 & 0.228 ± 0.005 & 0.198 ± 0.011 & 0.181 ± 0.006 & 0.718 ± 0.010 & 0.007\\
    id & 0.143 ± 0.004 & 0.126 ± 0.001 & 0.125 ± 0.001 & 0.149 ± 0.003 & 0.140 ± 0.003 & 0.133 ± 0.003 & 0.814 ± 0.093 & 0.004\\
    it & 0.110 ± 0.002 & 0.097 ± 0.001 & 0.108 ± 0.003 & 0.127 ± 0.006 & 0.100 ± 0.002 & 0.094 ± 0.004 & 0.730 ± 0.034 & 0.004\\
    ja & 0.512 ± 0.004 & 0.639 ± 0.015 & 0.463 ± 0.022 & 0.536 ± 0.027 & 0.583 ± 0.021 & 0.596 ± 0.075 & 0.956 ± 0.006 & 0.007\\
    nl & 0.151 ± 0.001 & 0.133 ± 0.005 & 0.136 ± 0.006 & 0.166 ± 0.002 & 0.136 ± 0.007 & 0.138 ± 0.000 & 0.745 ± 0.025 & 0.011\\
    pl & 0.124 ± 0.002 & 0.122 ± 0.002 & 0.133 ± 0.001 & 0.145 ± 0.005 & 0.123 ± 0.000 & 0.113 ± 0.001 & 0.806 ± 0.017 & 0.005\\
    pt & 0.145 ± 0.003 & 0.136 ± 0.002 & 0.152 ± 0.011 & 0.157 ± 0.000 & 0.139 ± 0.002 & 0.127 ± 0.006 & 0.767 ± 0.008 & 0.004\\
    ru & 0.152 ± 0.003 & 0.144 ± 0.001 & 0.159 ± 0.005 & 0.172 ± 0.013 & 0.147 ± 0.001 & 0.144 ± 0.003 & 0.761 ± 0.026 & 0.005\\
    sv & 0.168 ± 0.002 & 0.164 ± 0.005 & 0.169 ± 0.001 & 0.190 ± 0.010 & 0.176 ± 0.017 & 0.160 ± 0.001 & 0.754 ± 0.013 & 0.012\\
    th & 0.304 ± 0.002 & 0.314 ± 0.002 & 0.308 ± 0.002 & 0.322 ± 0.006 & 0.316 ± 0.007 & 0.307 ± 0.007 & 0.878 ± 0.097 & 0.007\\
    uk & 0.137 ± 0.002 & 0.136 ± 0.001 & 0.151 ± 0.002 & 0.156 ± 0.004 & 0.138 ± 0.001 & 0.134 ± 0.005 & 0.776 ± 0.009 & 0.010\\
    zh & 0.523 ± 0.008 & 0.785 ± 0.113 & 0.370 ± 0.002 & 0.514 ± 0.010 & 0.731 ± 0.033 & 0.908 ± 0.064 & 0.989 ± 0.007 & 0.004\\
    \bottomrule
    \end{tabular}
\end{table*}

\begin{table*}
    \caption{\textbf{English monolingual} fine-tuned models ranking based on the average CER. For comparison purposes, the best multilingual and the fine-tuned model without pre-training are included in the ranking, both in bold-face font. Where, HB = HuBERT, RW = R-wav2vec2, US = UniSpeech, USS = UniSpeech-SAT, W = wav2vec2, WL = WavLM, XR = XLS-R, X53 = XLSR-53, USM = UniSpeech-ML, NP = no-pretraining, and VP = VP-100k. Multilingual and no-pretrained models are marked in bold-face.}
    \label{tab:en_monolingual_performance_ranking}
    \footnotesize
    \centering
    \begin{tabular}{lllllllll}
    \toprule
    & \textbf{1st} & \textbf{2nd} & \textbf{3rd} & \textbf{4th} & \textbf{5th} & \textbf{6th} & \textbf{7th} & \textbf{8th}\\
    \midrule
    \multicolumn{9}{c}{\textbf{Validation}}\\
    \midrule
    ar & \textbf{XR} & WL & W & RW & US & HB & USS & \textbf{NP}\\
    de & \textbf{X53} & RW & WL & W & HB & US & USS & \textbf{NP}\\
    en & US & \textbf{USM} & WL & RW & W & HB & USS & \textbf{NP}\\
    es & \textbf{USM} & WL & RW & W & US & HB & USS & \textbf{NP}\\
    et & \textbf{VP} & RW & W & WL & HB & US & USS & \textbf{NP}\\
    fa & \textbf{X53} & WL & RW & US & HB & W & USS & \textbf{NP}\\
    fr & \textbf{USM} & WL & RW & W & HB & USS & US & \textbf{NP}\\
    id & \textbf{XR} & RW & US & WL & USS & W & HB & \textbf{NP}\\
    it & \textbf{USM} & WL & RW & W & US & HB & USS & \textbf{NP}\\
    ja & \textbf{USM} & US & HB & USS & W & WL & RW & \textbf{NP}\\
    nl & \textbf{VP} & RW & W & US & WL & HB & USS & \textbf{NP}\\
    pl & \textbf{XR} & WL & W & RW & HB & US & USS & \textbf{NP}\\
    pt & \textbf{XR} & WL & RW & W & HB & US & USS & \textbf{NP}\\
    ru & \textbf{XR} & RW & WL & W & HB & US & USS & \textbf{NP}\\
    sv & \textbf{XR} & WL & RW & HB & US & W & USS & \textbf{NP}\\
    th & \textbf{XR} & HB & WL & US & RW & W & USS & \textbf{NP}\\
    uk & \textbf{XR} & WL & RW & HB & W & US & USS & \textbf{NP}\\
    zh & \textbf{USM} & US & USS & HB & W & RW & WL & \textbf{NP}\\
    \midrule
    \multicolumn{9}{c}{\textbf{Test}}\\
    \midrule
    ar & \textbf{XR} & RW & WL & W & HB & US & USS & \textbf{NP}\\
    de & \textbf{X53} & RW & WL & W & HB & US & USS & \textbf{NP}\\
    en & US & \textbf{USM} & WL & RW & W & USS & HB & \textbf{NP}\\
    es & \textbf{USM} & WL & RW & W & US & HB & USS & \textbf{NP}\\
    et & \textbf{VP} & RW & WL & W & HB & US & USS & \textbf{NP}\\
    fa & \textbf{X53} & WL & RW & US & HB & W & USS & \textbf{NP}\\
    fr & \textbf{USM} & WL & RW & W & HB & USS & US & \textbf{NP}\\
    id & \textbf{XR} & US & RW & WL & W & HB & USS & \textbf{NP}\\
    it & \textbf{USM} & WL & RW & W & US & HB & USS & \textbf{NP}\\
    ja & \textbf{USM} & US & HB & USS & W & WL & RW & \textbf{NP}\\
    nl & \textbf{X53} & RW & US & W & WL & HB & USS & \textbf{NP}\\
    pl & \textbf{XR} & WL & RW & W & HB & US & USS & \textbf{NP}\\
    pt & \textbf{XR} & WL & RW & W & HB & US & USS & \textbf{NP}\\
    ru & \textbf{XR} & RW & WL & W & HB & US & USS & \textbf{NP}\\
    sv & \textbf{XR} & WL & RW & HB & US & W & USS & \textbf{NP}\\
    th & \textbf{XR} & HB & WL & US & RW & W & USS & \textbf{NP}\\
    uk & \textbf{XR} & WL & RW & HB & W & US & USS & \textbf{NP}\\
    zh & \textbf{USM} & US & USS & HB & W & RW & WL & \textbf{NP}\\
    \bottomrule
    \end{tabular}
\end{table*}

\begin{table*}
    \caption{The p-value of post-hoc hypothesis test on \textbf{English monolingual} models' performance. Where, HB=HuBERT, RW=R-wav2vec2, US=UniSpeech, USS=UniSpeech-SAT, W=wav2vec2, and WL=WavLM. Values with $p > 0.05$ are shown in bold-face font.}
    \label{tab:en_monolingual_performance_posthoc_test}
    \centering
    \small
    \begin{tabular}{p{2mm}p{3mm}p{3mm}p{3mm}p{3mm}p{3mm}p{3mm}p{3mm}p{3mm}p{3mm}p{3mm}p{3mm}p{3mm}p{3mm}p{3mm}p{3mm}p{3mm}p{3mm}p{3mm}p{3mm}p{3mm}p{3mm}}
    \toprule
    & \textbf{HB} & \textbf{HB} & \textbf{HB} & \textbf{HB} & \textbf{HB} & \textbf{HB} & \textbf{RW} & \textbf{RW} & \textbf{RW} & \textbf{RW} & \textbf{RW} & \textbf{US} & \textbf{US} & \textbf{US} & \textbf{US} & \textbf{USS} & \textbf{USS} & \textbf{USS} & \textbf{W} & \textbf{W} & \textbf{WL}\\
    & \textbf{RW} & \textbf{US} & \textbf{USS} & \textbf{W} & \textbf{WL} & \textbf{NP} & \textbf{US} & \textbf{USS} & \textbf{W} & \textbf{WL} & \textbf{NP} & \textbf{USS} & \textbf{W} & \textbf{WL} & \textbf{NP} & \textbf{W} & \textbf{WL} & \textbf{NP} & \textbf{WL} & \textbf{NP} & \textbf{NP}\\
    \midrule
    \multicolumn{22}{c}{\textbf{Validation}}\\
    \midrule
    ar & \textbf{1.00} & \textbf{1.00} & \textbf{1.00} & \textbf{0.47} & \textbf{0.29} & \textbf{0.59} & \textbf{1.00} & \textbf{0.33} & \textbf{1.00} & \textbf{1.00} & 0.04 & \textbf{0.47} & \textbf{1.00} & \textbf{1.00} & 0.05 & \textbf{0.11} & \textbf{0.06} & \textbf{1.00} & \textbf{1.00} & 0.01 & 0.01\\
    de & 0.00 & \textbf{0.61} & 0.03 & 0.05 & 0.02 & 0.00 & 0.00 & 0.00 & \textbf{0.14} & \textbf{0.30} & 0.00 & \textbf{0.17} & 0.01 & 0.00 & 0.01 & 0.00 & 0.00 & \textbf{0.30} & \textbf{0.61} & 0.00 & 0.00\\
    en & 0.00 & 0.00 & \textbf{0.82} & \textbf{0.14} & 0.00 & 0.03 & 0.01 & 0.00 & \textbf{0.14} & \textbf{0.17} & 0.00 & 0.00 & 0.00 & \textbf{0.17} & 0.00 & \textbf{0.14} & 0.00 & 0.04 & 0.00 & 0.00 & 0.00\\
    es & 0.00 & \textbf{0.35} & \textbf{0.77} & 0.01 & 0.00 & \textbf{0.10} & 0.03 & 0.00 & \textbf{0.52} & \textbf{0.86} & 0.00 & \textbf{0.08} & \textbf{0.41} & 0.04 & 0.00 & 0.00 & 0.00 & \textbf{0.41} & \textbf{0.52} & 0.00 & 0.00\\
    et & \textbf{0.08} & \textbf{0.34} & \textbf{0.17} & \textbf{0.10} & \textbf{0.34} & 0.00 & 0.00 & 0.00 & \textbf{1.00} & \textbf{1.00} & 0.00 & \textbf{1.00} & 0.00 & 0.01 & \textbf{0.21} & 0.00 & 0.00 & \textbf{0.34} & \textbf{1.00} & 0.00 & 0.00\\
    fa & \textbf{0.17} & \textbf{0.49} & \textbf{0.07} & \textbf{0.49} & 0.01 & 0.00 & \textbf{0.47} & 0.00 & 0.02 & \textbf{0.47} & 0.00 & 0.02 & \textbf{0.40} & 0.03 & 0.00 & \textbf{0.47} & 0.00 & \textbf{0.47} & 0.00 & 0.03 & 0.00\\
    fr & \textbf{0.30} & \textbf{0.72} & \textbf{0.25} & \textbf{0.72} & 0.02 & 0.01 & 0.04 & 0.00 & \textbf{0.60} & \textbf{0.60} & 0.00 & \textbf{0.72} & \textbf{0.60} & 0.00 & \textbf{0.09} & \textbf{0.09} & 0.00 & \textbf{0.60} & 0.05 & 0.00 & 0.00\\
    id & 0.00 & 0.00 & \textbf{0.66} & \textbf{1.00} & 0.01 & \textbf{0.44} & \textbf{1.00} & 0.04 & 0.01 & \textbf{1.00} & 0.00 & 0.03 & 0.01 & \textbf{1.00} & 0.00 & \textbf{1.00} & \textbf{0.13} & 0.03 & 0.02 & \textbf{0.15} & 0.00\\
    it & 0.00 & \textbf{0.81} & \textbf{0.32} & 0.02 & 0.00 & 0.02 & 0.01 & 0.00 & \textbf{0.81} & \textbf{0.81} & 0.00 & \textbf{0.07} & \textbf{0.08} & 0.00 & 0.00 & 0.00 & 0.00 & \textbf{0.53} & \textbf{0.53} & 0.00 & 0.00\\
    ja & 0.00 & \textbf{0.21} & \textbf{1.00} & 0.01 & 0.01 & 0.00 & 0.00 & 0.00 & \textbf{0.41} & \textbf{0.41} & \textbf{0.21} & \textbf{0.08} & 0.00 & 0.00 & 0.00 & 0.03 & 0.03 & 0.00 & \textbf{1.00} & 0.01 & 0.01\\
    nl & 0.00 & 0.05 & \textbf{0.96} & 0.01 & \textbf{0.48} & \textbf{0.10} & \textbf{0.96} & 0.00 & \textbf{0.96} & \textbf{0.10} & 0.00 & 0.00 & \textbf{0.96} & \textbf{0.96} & 0.00 & 0.00 & 0.05 & \textbf{0.96} & \textbf{0.22} & 0.00 & 0.00\\
    pl & \textbf{0.10} & \textbf{0.07} & 0.00 & 0.01 & 0.00 & 0.00 & 0.00 & 0.00 & \textbf{0.12} & 0.00 & 0.00 & \textbf{0.10} & 0.00 & 0.00 & 0.00 & 0.00 & 0.00 & \textbf{0.10} & 0.05 & 0.00 & 0.00\\
    pt & 0.01 & \textbf{0.10} & 0.04 & \textbf{0.08} & 0.00 & 0.00 & 0.00 & 0.00 & \textbf{0.52} & \textbf{0.09} & 0.00 & \textbf{0.52} & 0.00 & 0.00 & 0.02 & 0.00 & 0.00 & \textbf{0.08} & 0.02 & 0.00 & 0.00\\
    ru & 0.01 & \textbf{0.48} & \textbf{0.06} & \textbf{0.07} & \textbf{0.07} & 0.00 & 0.00 & 0.00 & \textbf{0.83} & \textbf{0.83} & 0.00 & \textbf{0.83} & 0.00 & 0.00 & \textbf{0.06} & 0.00 & 0.00 & \textbf{0.48} & \textbf{1.00} & 0.00 & 0.00\\
    sv & \textbf{0.44} & \textbf{1.00} & \textbf{0.29} & \textbf{1.00} & \textbf{0.29} & 0.02 & \textbf{0.08} & 0.01 & \textbf{0.62} & \textbf{1.00} & 0.00 & \textbf{1.00} & \textbf{1.00} & 0.04 & \textbf{0.21} & \textbf{0.21} & 0.00 & \textbf{1.00} & \textbf{0.39} & 0.02 & 0.00\\
    th & 0.04 & \textbf{0.73} & 0.00 & 0.02 & \textbf{0.57} & 0.00 & \textbf{0.42} & \textbf{0.43} & \textbf{1.00} & \textbf{0.49} & 0.05 & 0.01 & \textbf{0.27} & \textbf{1.00} & 0.00 & \textbf{0.54} & 0.02 & \textbf{0.73} & \textbf{0.42} & \textbf{0.07} & 0.00\\
    uk & \textbf{1.00} & \textbf{0.24} & 0.03 & \textbf{1.00} & \textbf{1.00} & 0.00 & \textbf{0.13} & 0.02 & \textbf{1.00} & \textbf{1.00} & 0.00 & \textbf{1.00} & \textbf{0.18} & 0.04 & \textbf{0.31} & 0.02 & 0.01 & \textbf{1.00} & \textbf{1.00} & 0.00 & 0.00\\
    zh & 0.01 & 0.01 & \textbf{0.23} & 0.03 & 0.00 & 0.00 & 0.00 & 0.00 & \textbf{0.64} & \textbf{0.10} & 0.00 & \textbf{0.12} & 0.00 & 0.00 & 0.00 & 0.00 & 0.00 & 0.00 & 0.05 & 0.00 & \textbf{0.12}\\
    \midrule
    \multicolumn{22}{c}{\textbf{Test}}\\
    \midrule
    ar & \textbf{0.79} & \textbf{1.00} & \textbf{1.00} & \textbf{1.00} & \textbf{0.79} & \textbf{0.25} & \textbf{1.00} & 0.05 & \textbf{1.00} & \textbf{1.00} & 0.01 & \textbf{0.79} & \textbf{1.00} & \textbf{1.00} & \textbf{0.12} & \textbf{0.17} & 0.05 & \textbf{1.00} & \textbf{1.00} & 0.03 & 0.01\\
    de & 0.00 & \textbf{0.86} & \textbf{0.15} & \textbf{0.24} & 0.01 & 0.01 & 0.00 & 0.00 & \textbf{0.15} & \textbf{0.76} & 0.00 & \textbf{0.17} & \textbf{0.20} & 0.01 & 0.01 & 0.00 & 0.00 & \textbf{0.38} & \textbf{0.37} & 0.00 & 0.00\\
    en & 0.01 & 0.00 & \textbf{0.70} & \textbf{0.56} & 0.00 & \textbf{0.07} & 0.03 & 0.02 & \textbf{0.09} & \textbf{0.48} & 0.00 & 0.00 & 0.00 & \textbf{0.48} & 0.00 & \textbf{0.68} & 0.00 & 0.04 & 0.01 & 0.01 & 0.00\\
    es & 0.00 & \textbf{0.25} & \textbf{0.84} & 0.04 & 0.00 & \textbf{0.15} & \textbf{0.10} & 0.00 & \textbf{0.54} & \textbf{0.87} & 0.00 & \textbf{0.08} & \textbf{0.79} & \textbf{0.12} & 0.00 & 0.01 & 0.00 & \textbf{0.54} & \textbf{0.54} & 0.00 & 0.00\\
    et & \textbf{0.08} & \textbf{0.34} & \textbf{0.17} & \textbf{0.34} & \textbf{0.10} & 0.00 & 0.00 & 0.00 & \textbf{1.00} & \textbf{1.00} & 0.00 & \textbf{1.00} & 0.01 & 0.00 & \textbf{0.21} & 0.00 & 0.00 & \textbf{0.34} & \textbf{1.00} & 0.00 & 0.00\\
    fa & \textbf{0.34} & \textbf{0.48} & 0.03 & \textbf{0.34} & 0.01 & 0.00 & \textbf{0.38} & 0.00 & 0.01 & \textbf{0.34} & 0.00 & 0.01 & \textbf{0.14} & 0.03 & 0.00 & \textbf{0.38} & 0.00 & \textbf{0.37} & 0.00 & 0.04 & 0.00\\
    fr & \textbf{0.21} & \textbf{0.48} & \textbf{0.12} & \textbf{0.74} & 0.01 & 0.00 & 0.01 & 0.00 & \textbf{0.34} & \textbf{0.48} & 0.00 & \textbf{0.53} & \textbf{0.40} & 0.00 & \textbf{0.08} & \textbf{0.08} & 0.00 & \textbf{0.40} & 0.02 & 0.00 & 0.00\\
    id & 0.00 & 0.00 & \textbf{0.16} & \textbf{0.76} & 0.02 & 0.00 & \textbf{0.76} & 0.00 & 0.00 & \textbf{0.10} & 0.00 & 0.00 & 0.00 & 0.03 & 0.00 & 0.05 & 0.00 & \textbf{0.16} & \textbf{0.10} & 0.00 & 0.00\\
    it & 0.00 & \textbf{0.41} & \textbf{0.06} & 0.01 & 0.00 & 0.00 & 0.00 & 0.00 & \textbf{0.06} & \textbf{0.41} & 0.00 & 0.01 & \textbf{0.06} & 0.00 & 0.00 & 0.00 & 0.00 & \textbf{0.09} & 0.01 & 0.00 & 0.00\\
    ja & 0.00 & \textbf{0.51} & \textbf{0.51} & 0.03 & 0.03 & 0.00 & 0.00 & 0.03 & \textbf{0.51} & \textbf{0.51} & \textbf{0.51} & 0.04 & 0.00 & 0.00 & 0.00 & \textbf{0.51} & \textbf{0.51} & 0.00 & \textbf{0.87} & 0.03 & 0.02\\
    nl & 0.03 & \textbf{0.19} & \textbf{1.00} & \textbf{0.20} & \textbf{0.19} & \textbf{0.30} & \textbf{1.00} & 0.00 & \textbf{1.00} & \textbf{1.00} & 0.00 & 0.02 & \textbf{1.00} & \textbf{1.00} & 0.00 & 0.03 & 0.02 & \textbf{1.00} & \textbf{1.00} & 0.00 & 0.00\\
    pl & \textbf{0.56} & \textbf{0.07} & 0.00 & \textbf{0.70} & 0.01 & 0.00 & 0.01 & 0.00 & \textbf{0.68} & \textbf{0.09} & 0.00 & \textbf{0.48} & 0.04 & 0.00 & 0.03 & 0.00 & 0.00 & \textbf{0.48} & 0.02 & 0.00 & 0.00\\
    pt & 0.01 & \textbf{0.27} & \textbf{0.07} & \textbf{0.10} & 0.00 & 0.00 & 0.00 & 0.00 & \textbf{0.27} & \textbf{0.15} & 0.00 & \textbf{0.27} & 0.00 & 0.00 & 0.01 & 0.00 & 0.00 & \textbf{0.10} & 0.01 & 0.00 & 0.00\\
    ru & 0.00 & \textbf{0.23} & 0.02 & \textbf{0.21} & 0.00 & 0.00 & 0.00 & 0.00 & \textbf{0.21} & \textbf{0.83} & 0.00 & \textbf{0.30} & 0.00 & 0.00 & 0.02 & 0.00 & 0.00 & \textbf{0.23} & \textbf{0.23} & 0.00 & 0.00\\
    sv & \textbf{1.00} & \textbf{1.00} & \textbf{0.20} & \textbf{1.00} & \textbf{0.28} & 0.02 & \textbf{0.20} & 0.01 & \textbf{0.46} & \textbf{1.00} & 0.00 & \textbf{1.00} & \textbf{1.00} & 0.04 & \textbf{0.14} & \textbf{0.41} & 0.00 & \textbf{1.00} & \textbf{0.14} & 0.04 & 0.00\\
    th & 0.01 & \textbf{0.67} & 0.00 & 0.00 & \textbf{0.67} & 0.00 & \textbf{0.10} & \textbf{0.54} & \textbf{1.00} & \textbf{0.08} & 0.03 & 0.00 & \textbf{0.07} & \textbf{1.00} & 0.00 & \textbf{0.67} & 0.00 & \textbf{0.61} & \textbf{0.06} & \textbf{0.06} & 0.00\\
    uk & \textbf{1.00} & 0.04 & 0.01 & \textbf{1.00} & \textbf{1.00} & 0.00 & 0.04 & 0.01 & \textbf{1.00} & \textbf{1.00} & 0.00 & \textbf{1.00} & \textbf{0.50} & 0.02 & \textbf{0.33} & \textbf{0.09} & 0.00 & \textbf{1.00} & \textbf{0.74} & 0.01 & 0.00\\
    zh & 0.01 & 0.02 & \textbf{0.54} & 0.03 & 0.00 & 0.00 & 0.00 & 0.00 & \textbf{0.65} & \textbf{0.12} & 0.00 & \textbf{0.12} & 0.00 & 0.00 & 0.00 & 0.00 & 0.00 & 0.00 & \textbf{0.06} & 0.00 & \textbf{0.12}\\
    \bottomrule
    \end{tabular}
\end{table*}

\begin{table*}
    \caption{The frequency with which a \textbf{English monolingual} fine-tuned model pair showed no statistical difference (p-value above 0.05), no-pretraining in bold-face font.}
    \label{tab:en_monolingual_performance_posthoc_test_frequency}
    \footnotesize
    \centering
    \begin{tabular}{p{25mm}p{25mm}p{5mm}}
    \toprule
    \multicolumn{3}{c}{\textbf{Validation}}\\
    \midrule
    R-wav2vec2 & WavLM & 17\\
    R-wav2vec2 & wav2vec2 & 16\\
    HuBERT & UniSpeech & 14\\
    UniSpeech-SAT & \textbf{no-pretraining} & 14\\
    UniSpeech & UniSpeech-SAT & 13\\
    HuBERT & UniSpeech-SAT & 13\\
    wav2vec2 & WavLM & 11\\
    HuBERT & wav2vec2 & 10\\
    UniSpeech & wav2vec2 & 9\\
    HuBERT & R-wav2vec2 & 7\\
    HuBERT & WavLM & 7\\
    R-wav2vec2 & UniSpeech & 7\\
    UniSpeech-SAT & wav2vec2 & 7\\
    UniSpeech & WavLM & 5\\
    UniSpeech & \textbf{no-pretraining} & 5\\
    HuBERT & \textbf{no-pretraining} & 4\\
    R-wav2vec2 & UniSpeech-SAT & 2\\
    UniSpeech-SAT & WavLM & 2\\
    wav2vec2 & \textbf{no-pretraining} & 2\\
    R-wav2vec2 & \textbf{no-pretraining} & 1\\
    WavLM & \textbf{no-pretraining} & 1\\
    \midrule
    \multicolumn{3}{c}{\textbf{Test}}\\
    \midrule
    R-wav2vec2 & WavLM & 18\\
    R-wav2vec2 & wav2vec2 & 16\\
    UniSpeech-SAT & \textbf{no-pretraining} & 15\\
    HuBERT & UniSpeech & 14\\
    HuBERT & UniSpeech-SAT & 13\\
    HuBERT & wav2vec2 & 13\\
    wav2vec2 & WavLM & 12\\
    UniSpeech & UniSpeech-SAT & 11\\
    UniSpeech & wav2vec2 & 10\\
    UniSpeech-SAT & wav2vec2 & 8\\
    HuBERT & R-wav2vec2 & 7\\
    R-wav2vec2 & UniSpeech & 7\\
    HuBERT & WavLM & 6\\
    UniSpeech & WavLM & 5\\
    UniSpeech & \textbf{no-pretraining} & 5\\
    HuBERT & \textbf{no-pretraining} & 4\\
    R-wav2vec2 & UniSpeech-SAT & 1\\
    R-wav2vec2 & \textbf{no-pretraining} & 1\\
    UniSpeech-SAT & WavLM & 1\\
    wav2vec2 & \textbf{no-pretraining} & 1\\
    WavLM & \textbf{no-pretraining} & 1\\
    \bottomrule
    \end{tabular}
\end{table*}



\clearpage
\begin{table*}
    \caption{Average CER on \textbf{VoxPopuli monolingual} and no pre-trained fine-tuned models and the p-value of hypothesis test on models' performance. Values in bold-face correspond to $p > 0.05$.}
    \label{tab:vp_monolingual_overall_performance}
    \footnotesize
    \centering
    \begin{tabular}{p{5mm}p{20mm}p{20mm}p{20mm}p{20mm}p{20mm}p{20mm}p{12mm}}
    \toprule
    & \textbf{VP-es} & \textbf{VP-fr} & \textbf{VP-it} & \textbf{VP-nl} & \textbf{VP-sv} & \textbf{NP} & \textbf{p-value}\\
    \midrule
    \multicolumn{8}{c}{\textbf{Validation}}\\
    \midrule
    ar & 0.301 ± 0.004 & 0.291 ± 0.007 & 0.283 ± 0.006 & 0.283 ± 0.006 & 0.280 ± 0.002 & 0.782 ± 0.058 & 0.021\\
    de & 0.188 ± 0.009 & 0.175 ± 0.002 & 0.190 ± 0.007 & 0.138 ± 0.002 & 0.164 ± 0.006 & 0.749 ± 0.058 & 0.006\\
    en & 0.311 ± 0.005 & 0.293 ± 0.002 & 0.308 ± 0.009 & 0.283 ± 0.003 & 0.298 ± 0.007 & 0.762 ± 0.009 & 0.010\\
    es & 0.063 ± 0.002 & 0.134 ± 0.002 & 0.111 ± 0.002 & 0.142 ± 0.005 & 0.145 ± 0.008 & 0.762 ± 0.039 & 0.007\\
    et & 0.172 ± 0.005 & 0.177 ± 0.003 & 0.180 ± 0.008 & 0.192 ± 0.004 & 0.172 ± 0.005 & 0.725 ± 0.015 & 0.019\\
    fa & 0.197 ± 0.007 & 0.193 ± 0.003 & 0.302 ± 0.179 & 0.246 ± 0.005 & 0.188 ± 0.001 & 0.770 ± 0.046 & 0.027\\
    fr & 0.211 ± 0.002 & 0.172 ± 0.106 & 0.209 ± 0.004 & 0.229 ± 0.004 & 0.241 ± 0.006 & 0.716 ± 0.010 & 0.036\\
    id & 0.191 ± 0.009 & 0.204 ± 0.012 & 0.209 ± 0.003 & 0.208 ± 0.010 & 0.213 ± 0.007 & 0.807 ± 0.085 & 0.043\\
    it & 0.098 ± 0.000 & 0.114 ± 0.004 & 0.054 ± 0.001 & 0.133 ± 0.002 & 0.133 ± 0.004 & 0.728 ± 0.034 & 0.007\\
    ja & 0.564 ± 0.023 & 0.603 ± 0.011 & 0.565 ± 0.007 & 0.604 ± 0.012 & 0.595 ± 0.009 & 0.948 ± 0.006 & 0.014\\
    nl & 0.180 ± 0.001 & 0.173 ± 0.004 & 0.194 ± 0.002 & 0.043 ± 0.010 & 0.161 ± 0.003 & 0.723 ± 0.020 & 0.005\\
    pl & 0.153 ± 0.003 & 0.149 ± 0.006 & 0.151 ± 0.005 & 0.150 ± 0.003 & 0.151 ± 0.002 & 0.798 ± 0.012 & \textbf{0.144}\\
    pt & 0.122 ± 0.008 & 0.165 ± 0.003 & 0.144 ± 0.004 & 0.179 ± 0.003 & 0.208 ± 0.018 & 0.771 ± 0.009 & 0.005\\
    ru & 0.166 ± 0.003 & 0.161 ± 0.008 & 0.172 ± 0.009 & 0.176 ± 0.007 & 0.174 ± 0.003 & 0.754 ± 0.025 & 0.039\\
    sv & 0.252 ± 0.017 & 0.228 ± 0.002 & 0.252 ± 0.003 & 0.220 ± 0.010 & 0.092 ± 0.010 & 0.767 ± 0.015 & 0.008\\
    th & 0.353 ± 0.006 & 0.348 ± 0.002 & 0.349 ± 0.006 & 0.354 ± 0.008 & 0.346 ± 0.004 & 0.876 ± 0.098 & \textbf{0.104}\\
    uk & 0.153 ± 0.002 & 0.148 ± 0.006 & 0.148 ± 0.003 & 0.153 ± 0.007 & 0.159 ± 0.004 & 0.770 ± 0.003 & \textbf{0.054}\\
    zh & 0.707 ± 0.058 & 0.627 ± 0.002 & 0.632 ± 0.038 & 0.643 ± 0.027 & 0.614 ± 0.027 & 0.989 ± 0.007 & 0.048\\
    \midrule
    \multicolumn{8}{c}{\textbf{Test}}\\
    \midrule
    ar & 0.298 ± 0.008 & 0.294 ± 0.008 & 0.286 ± 0.003 & 0.286 ± 0.008 & 0.287 ± 0.003 & 0.792 ± 0.050 & 0.041\\
    de & 0.202 ± 0.010 & 0.189 ± 0.002 & 0.204 ± 0.007 & 0.151 ± 0.002 & 0.178 ± 0.005 & 0.751 ± 0.056 & 0.007\\
    en & 0.353 ± 0.006 & 0.336 ± 0.003 & 0.353 ± 0.010 & 0.328 ± 0.004 & 0.342 ± 0.007 & 0.773 ± 0.001 & 0.011\\
    es & 0.073 ± 0.002 & 0.146 ± 0.003 & 0.123 ± 0.001 & 0.155 ± 0.005 & 0.158 ± 0.008 & 0.767 ± 0.034 & 0.006\\
    et & 0.164 ± 0.006 & 0.167 ± 0.004 & 0.170 ± 0.010 & 0.179 ± 0.006 & 0.162 ± 0.005 & 0.741 ± 0.019 & 0.032\\
    fa & 0.228 ± 0.006 & 0.225 ± 0.002 & 0.337 ± 0.189 & 0.264 ± 0.005 & 0.220 ± 0.002 & 0.769 ± 0.041 & 0.028\\
    fr & 0.227 ± 0.003 & 0.184 ± 0.106 & 0.225 ± 0.004 & 0.246 ± 0.004 & 0.257 ± 0.007 & 0.718 ± 0.010 & 0.036\\
    id & 0.183 ± 0.011 & 0.201 ± 0.016 & 0.199 ± 0.003 & 0.197 ± 0.013 & 0.201 ± 0.004 & 0.814 ± 0.093 & \textbf{0.051}\\
    it & 0.105 ± 0.001 & 0.120 ± 0.004 & 0.058 ± 0.001 & 0.140 ± 0.002 & 0.139 ± 0.004 & 0.730 ± 0.034 & 0.007\\
    ja & 0.627 ± 0.015 & 0.658 ± 0.013 & 0.632 ± 0.016 & 0.657 ± 0.017 & 0.635 ± 0.009 & 0.956 ± 0.006 & 0.024\\
    nl & 0.213 ± 0.003 & 0.207 ± 0.006 & 0.230 ± 0.003 & 0.069 ± 0.011 & 0.196 ± 0.004 & 0.745 ± 0.025 & 0.006\\
    pl & 0.174 ± 0.004 & 0.169 ± 0.008 & 0.170 ± 0.005 & 0.169 ± 0.002 & 0.173 ± 0.002 & 0.806 ± 0.017 & \textbf{0.084}\\
    pt & 0.129 ± 0.008 & 0.171 ± 0.003 & 0.150 ± 0.004 & 0.187 ± 0.003 & 0.214 ± 0.017 & 0.767 ± 0.008 & 0.005\\
    ru & 0.185 ± 0.002 & 0.180 ± 0.008 & 0.190 ± 0.009 & 0.195 ± 0.007 & 0.192 ± 0.002 & 0.761 ± 0.026 & 0.037\\
    sv & 0.249 ± 0.017 & 0.226 ± 0.002 & 0.246 ± 0.002 & 0.219 ± 0.006 & 0.095 ± 0.010 & 0.754 ± 0.013 & 0.007\\
    th & 0.357 ± 0.005 & 0.350 ± 0.001 & 0.354 ± 0.008 & 0.357 ± 0.008 & 0.349 ± 0.003 & 0.878 ± 0.097 & \textbf{0.059}\\
    uk & 0.176 ± 0.002 & 0.171 ± 0.007 & 0.173 ± 0.003 & 0.177 ± 0.008 & 0.180 ± 0.003 & 0.776 ± 0.009 & \textbf{0.059}\\
    zh & 0.716 ± 0.056 & 0.635 ± 0.001 & 0.643 ± 0.038 & 0.653 ± 0.027 & 0.627 ± 0.027 & 0.989 ± 0.007 & 0.050\\
    \bottomrule
    \end{tabular}
\end{table*}

\begin{table*}
    \caption{\textbf{VoxPopuli monolingual} fine-tuned models ranking based on the average CER. For comparison purposes, the best multilingual and the fine-tuned model without pre-training are included in the ranking in bold-face font. Where, US-ML = UniSpeech-ML, and NP = no-pretraining. Multilingual and no-pretrained models are marked in bold-face.}
    \label{tab:vp_monolingual_performance_ranking}
    \footnotesize
    \centering
    \begin{tabular}{llllllll}
    \toprule
    & \textbf{1st} & \textbf{2nd} & \textbf{3rd} & \textbf{4th} & \textbf{5th} & \textbf{6th} & \textbf{7th}\\
    \midrule
    \multicolumn{8}{c}{\textbf{Validation}}\\
    \midrule
    ar & \textbf{XLS-R} & VP-sv & VP-it & VP-nl & VP-fr & VP-es & \textbf{NP}\\
    de & \textbf{XLSR-53} & VP-nl & VP-sv & VP-fr & VP-es & VP-it & \textbf{NP}\\
    en & \textbf{US-ML} & VP-nl & VP-fr & VP-sv & VP-it & VP-es & \textbf{NP}\\
    es & \textbf{US-ML} & VP-es & VP-it & VP-fr & VP-nl & VP-sv & \textbf{NP}\\
    et & \textbf{VP-100k} & VP-es & VP-sv & VP-fr & VP-it & VP-nl & \textbf{NP}\\
    fa & \textbf{XLSR-53} & VP-sv & VP-fr & VP-es & VP-nl & VP-it & \textbf{NP}\\
    fr & \textbf{US-ML} & VP-fr & VP-it & VP-es & VP-nl & VP-sv & \textbf{NP}\\
    id & \textbf{XLS-R} & VP-es & VP-fr & VP-nl & VP-it & VP-sv & \textbf{NP}\\
    it & \textbf{US-ML} & VP-it & VP-es & VP-fr & VP-nl & VP-sv & \textbf{NP}\\
    ja & \textbf{US-ML} & VP-es & VP-it & VP-sv & VP-fr & VP-nl & \textbf{NP}\\
    nl & \textbf{VP-100k} & VP-nl & VP-sv & VP-fr & VP-es & VP-it & \textbf{NP}\\
    pl & \textbf{XLS-R} & VP-fr & VP-nl & VP-it & VP-sv & VP-es & \textbf{NP}\\
    pt & \textbf{XLS-R} & VP-es & VP-it & VP-fr & VP-nl & VP-sv & \textbf{NP}\\
    ru & \textbf{XLS-R} & VP-fr & VP-es & VP-it & VP-sv & VP-nl & \textbf{NP}\\
    sv & \textbf{XLS-R} & VP-sv & VP-nl & VP-fr & VP-es & VP-it & \textbf{NP}\\
    th & \textbf{XLS-R} & VP-sv & VP-fr & VP-it & VP-es & VP-nl & \textbf{NP}\\
    uk & \textbf{XLS-R} & VP-fr & VP-it & VP-es & VP-nl & VP-sv & \textbf{NP}\\
    zh & \textbf{US-ML} & VP-sv & VP-fr & VP-it & VP-nl & VP-es & \textbf{NP}\\
    \midrule
    \multicolumn{8}{c}{\textbf{Test}}\\
    \midrule
    ar & \textbf{XLS-R} & VP-it & VP-nl & VP-sv & VP-fr & VP-es & \textbf{NP}\\
    de & \textbf{XLSR-53} & VP-nl & VP-sv & VP-fr & VP-es & VP-it & \textbf{NP}\\
    en & \textbf{US-ML} & VP-nl & VP-fr & VP-sv & VP-es & VP-it & \textbf{NP}\\
    es & \textbf{US-ML} & VP-es & VP-it & VP-fr & VP-nl & VP-sv & \textbf{NP}\\
    et & \textbf{VP-100k} & VP-sv & VP-es & VP-fr & VP-it & VP-nl & \textbf{NP}\\
    fa & \textbf{XLSR-53} & VP-sv & VP-fr & VP-es & VP-nl & VP-it & \textbf{NP}\\
    fr & \textbf{US-ML} & VP-fr & VP-it & VP-es & VP-nl & VP-sv & \textbf{NP}\\
    id & \textbf{XLS-R} & VP-es & VP-nl & VP-it & VP-fr & VP-sv & \textbf{NP}\\
    it & \textbf{US-ML} & VP-it & VP-es & VP-fr & VP-sv & VP-nl & \textbf{NP}\\
    ja & \textbf{US-ML} & VP-es & VP-it & VP-sv & VP-nl & VP-fr & \textbf{NP}\\
    nl & \textbf{XLSR-53} & VP-nl & VP-sv & VP-fr & VP-es & VP-it & \textbf{NP}\\
    pl & \textbf{XLS-R} & VP-fr & VP-nl & VP-it & VP-sv & VP-es & \textbf{NP}\\
    pt & \textbf{XLS-R} & VP-es & VP-it & VP-fr & VP-nl & VP-sv & \textbf{NP}\\
    ru & \textbf{XLS-R} & VP-fr & VP-es & VP-it & VP-sv & VP-nl & \textbf{NP}\\
    sv & \textbf{XLS-R} & VP-sv & VP-nl & VP-fr & VP-it & VP-es & \textbf{NP}\\
    th & \textbf{XLS-R} & VP-sv & VP-fr & VP-it & VP-es & VP-nl & \textbf{NP}\\
    uk & \textbf{XLS-R} & VP-fr & VP-it & VP-es & VP-nl & VP-sv & \textbf{NP}\\
    zh & \textbf{US-ML} & VP-sv & VP-fr & VP-it & VP-nl & VP-es & \textbf{NP}\\
    \bottomrule
    \end{tabular}
\end{table*}

\begin{table*}
    \caption{The p-value of post-hoc hypothesis test on \textbf{VoxPopuli monolingual} models' performance. Values where $p > 0.05$ are marked in bold-face.}
    \label{tab:vp_monolingual_performance_posthoc_test}
    \footnotesize
    \centering
    \begin{tabular}{lrrrrrrrrrrrrrrr}
    \toprule
    & \textbf{es} & \textbf{es} & \textbf{es} & \textbf{es} & \textbf{es} & \textbf{fr} & \textbf{fr} & \textbf{fr} & \textbf{fr} & \textbf{it} & \textbf{it} & \textbf{it} & \textbf{nl} & \textbf{nl} & \textbf{sv}\\
    & \textbf{fr} & \textbf{it} & \textbf{nl} & \textbf{sv} & \textbf{NP} & \textbf{it} & \textbf{nl} & \textbf{sv} & \textbf{NP} & \textbf{nl} & \textbf{sv} & \textbf{NP} & \textbf{sv} & \textbf{NP} & \textbf{NP}\\
    \midrule
    \multicolumn{16}{c}{\textbf{Validation}}\\
    \midrule
    ar & \textbf{0.93} & \textbf{0.09} & \textbf{0.10} & 0.02 & \textbf{0.93} & \textbf{0.87} & \textbf{0.93} & \textbf{0.23} & \textbf{0.12} & \textbf{1.00} & \textbf{1.00} & 0.01 & \textbf{1.00} & 0.01 & 0.00\\
    de & 0.02 & \textbf{0.40} & 0.00 & 0.00 & 0.01 & 0.01 & 0.00 & \textbf{0.07} & 0.00 & 0.00 & 0.00 & 0.02 & \textbf{0.07} & 0.00 & 0.00\\
    en & 0.03 & \textbf{0.73} & 0.00 & \textbf{0.09} & \textbf{0.15} & \textbf{0.07} & \textbf{0.17} & \textbf{0.73} & 0.00 & 0.00 & \textbf{0.17} & \textbf{0.07} & \textbf{0.07} & 0.00 & 0.00\\
    es & 0.00 & \textbf{0.08} & 0.00 & 0.00 & 0.00 & \textbf{0.08} & 0.02 & 0.01 & 0.00 & 0.00 & 0.00 & 0.00 & \textbf{0.78} & 0.01 & 0.02\\
    et & \textbf{0.68} & \textbf{0.62} & 0.02 & \textbf{1.00} & 0.00 & \textbf{1.00} & \textbf{0.29} & \textbf{0.68} & 0.03 & \textbf{0.33} & \textbf{0.68} & 0.04 & 0.02 & \textbf{0.68} & 0.00\\
    fa & \textbf{1.00} & \textbf{1.00} & \textbf{0.70} & \textbf{0.51} & \textbf{0.07} & \textbf{1.00} & \textbf{0.32} & \textbf{0.91} & 0.03 & \textbf{0.91} & \textbf{0.37} & \textbf{0.11} & 0.03 & \textbf{0.91} & 0.00\\
    fr & \textbf{1.00} & \textbf{1.00} & \textbf{1.00} & \textbf{0.26} & 0.03 & \textbf{1.00} & \textbf{1.00} & \textbf{0.30} & 0.03 & \textbf{1.00} & \textbf{0.23} & 0.02 & \textbf{1.00} & \textbf{0.30} & \textbf{1.00}\\
    id & \textbf{0.98} & \textbf{0.36} & \textbf{0.36} & \textbf{0.25} & 0.01 & \textbf{1.00} & \textbf{1.00} & \textbf{1.00} & \textbf{0.09} & \textbf{1.00} & \textbf{1.00} & \textbf{0.35} & \textbf{1.00} & \textbf{0.35} & \textbf{0.43}\\
    it & \textbf{0.08} & \textbf{0.08} & 0.00 & 0.00 & 0.00 & 0.00 & 0.01 & 0.02 & 0.00 & 0.00 & 0.00 & 0.00 & \textbf{0.78} & 0.02 & 0.01\\
    ja & 0.03 & \textbf{1.00} & 0.02 & \textbf{0.19} & 0.00 & 0.02 & \textbf{1.00} & \textbf{0.83} & \textbf{0.14} & 0.01 & \textbf{0.14} & 0.00 & \textbf{0.52} & \textbf{0.21} & 0.02\\
    nl & 0.02 & 0.02 & 0.00 & 0.00 & 0.00 & 0.00 & 0.00 & 0.02 & 0.00 & 0.00 & 0.00 & 0.02 & 0.02 & 0.00 & 0.00\\
    pl & \textbf{1.00} & \textbf{1.00} & \textbf{1.00} & \textbf{1.00} & \textbf{1.00} & \textbf{1.00} & \textbf{1.00} & \textbf{1.00} & \textbf{0.21} & \textbf{1.00} & \textbf{1.00} & \textbf{0.31} & \textbf{1.00} & \textbf{0.28} & \textbf{0.64}\\
    pt & 0.00 & 0.02 & 0.00 & 0.00 & 0.00 & 0.02 & 0.02 & 0.00 & 0.00 & 0.00 & 0.00 & 0.00 & 0.02 & 0.00 & 0.02\\
    ru & \textbf{1.00} & \textbf{0.96} & \textbf{0.59} & \textbf{0.59} & 0.02 & \textbf{0.59} & \textbf{0.31} & \textbf{0.36} & 0.01 & \textbf{1.00} & \textbf{1.00} & \textbf{0.28} & \textbf{1.00} & \textbf{0.54} & \textbf{0.49}\\
    sv & 0.03 & \textbf{0.99} & 0.01 & 0.00 & 0.03 & 0.01 & \textbf{0.99} & 0.03 & 0.00 & 0.00 & 0.00 & \textbf{0.06} & \textbf{0.06} & 0.00 & 0.00\\
    th & \textbf{1.00} & \textbf{1.00} & \textbf{1.00} & \textbf{1.00} & \textbf{0.91} & \textbf{1.00} & \textbf{1.00} & \textbf{1.00} & \textbf{0.25} & \textbf{1.00} & \textbf{1.00} & \textbf{0.25} & \textbf{1.00} & \textbf{0.85} & \textbf{0.08}\\
    uk & \textbf{1.00} & \textbf{1.00} & \textbf{1.00} & \textbf{1.00} & \textbf{0.29} & \textbf{1.00} & \textbf{1.00} & \textbf{0.40} & 0.03 & \textbf{1.00} & \textbf{0.44} & 0.04 & \textbf{1.00} & \textbf{0.21} & \textbf{1.00}\\
    zh & \textbf{0.39} & \textbf{0.77} & \textbf{1.00} & \textbf{0.24} & \textbf{1.00} & \textbf{1.00} & \textbf{1.00} & \textbf{1.00} & 0.05 & \textbf{1.00} & \textbf{1.00} & \textbf{0.10} & \textbf{1.00} & \textbf{0.21} & 0.03\\
    \midrule
    \multicolumn{16}{c}{\textbf{Test}}\\
    \midrule
    ar & \textbf{1.00} & \textbf{0.22} & \textbf{0.52} & \textbf{0.52} & \textbf{1.00} & \textbf{0.75} & \textbf{1.00} & \textbf{1.00} & \textbf{0.38} & \textbf{1.00} & \textbf{1.00} & 0.02 & \textbf{1.00} & 0.05 & 0.05\\
    de & 0.02 & \textbf{0.78} & 0.00 & 0.00 & 0.01 & 0.01 & 0.00 & \textbf{0.08} & 0.00 & 0.00 & 0.00 & 0.02 & \textbf{0.08} & 0.00 & 0.00\\
    en & 0.05 & \textbf{1.00} & 0.00 & \textbf{0.35} & \textbf{0.13} & 0.05 & \textbf{0.35} & \textbf{0.35} & 0.00 & 0.00 & \textbf{0.35} & \textbf{0.13} & 0.05 & 0.00 & 0.01\\
    es & 0.00 & 0.05 & 0.00 & 0.00 & 0.00 & 0.05 & 0.03 & 0.00 & 0.00 & 0.00 & 0.00 & 0.00 & \textbf{0.15} & 0.00 & 0.03\\
    et & \textbf{1.00} & \textbf{1.00} & \textbf{0.20} & \textbf{1.00} & 0.02 & \textbf{1.00} & \textbf{0.36} & \textbf{1.00} & 0.04 & \textbf{1.00} & \textbf{0.75} & \textbf{0.18} & \textbf{0.08} & \textbf{1.00} & 0.01\\
    fa & \textbf{1.00} & \textbf{1.00} & \textbf{0.54} & \textbf{0.73} & 0.05 & \textbf{1.00} & \textbf{0.49} & \textbf{0.78} & 0.04 & \textbf{0.79} & \textbf{0.44} & \textbf{0.12} & 0.04 & \textbf{0.79} & 0.00\\
    fr & \textbf{1.00} & \textbf{1.00} & \textbf{1.00} & \textbf{0.26} & 0.03 & \textbf{1.00} & \textbf{1.00} & \textbf{0.30} & 0.03 & \textbf{1.00} & \textbf{0.23} & 0.02 & \textbf{1.00} & \textbf{0.30} & \textbf{1.00}\\
    id & \textbf{0.87} & \textbf{0.62} & \textbf{1.00} & \textbf{0.28} & 0.01 & \textbf{1.00} & \textbf{1.00} & \textbf{1.00} & \textbf{0.25} & \textbf{1.00} & \textbf{1.00} & \textbf{0.38} & \textbf{1.00} & \textbf{0.10} & \textbf{0.82}\\
    it & \textbf{0.08} & \textbf{0.08} & 0.00 & 0.00 & 0.00 & 0.00 & 0.01 & 0.02 & 0.00 & 0.00 & 0.00 & 0.00 & \textbf{0.78} & 0.02 & 0.01\\
    ja & \textbf{0.13} & \textbf{1.00} & \textbf{0.13} & \textbf{1.00} & 0.00 & \textbf{0.29} & \textbf{1.00} & \textbf{0.29} & \textbf{0.44} & \textbf{0.29} & \textbf{1.00} & 0.01 & \textbf{0.29} & \textbf{0.44} & 0.01\\
    nl & \textbf{0.40} & 0.02 & 0.00 & 0.01 & 0.00 & 0.01 & 0.00 & 0.02 & 0.00 & 0.00 & 0.00 & \textbf{0.07} & \textbf{0.07} & 0.00 & 0.00\\
    pl & \textbf{1.00} & \textbf{1.00} & \textbf{1.00} & \textbf{1.00} & \textbf{1.00} & \textbf{1.00} & \textbf{1.00} & \textbf{1.00} & \textbf{0.12} & \textbf{1.00} & \textbf{1.00} & \textbf{0.12} & \textbf{1.00} & \textbf{0.11} & \textbf{0.74}\\
    pt & 0.00 & 0.02 & 0.00 & 0.00 & 0.00 & 0.02 & 0.02 & 0.00 & 0.00 & 0.00 & 0.00 & 0.00 & 0.02 & 0.00 & 0.02\\
    ru & \textbf{1.00} & \textbf{1.00} & \textbf{0.52} & \textbf{0.60} & 0.02 & \textbf{0.60} & \textbf{0.24} & \textbf{0.34} & 0.01 & \textbf{1.00} & \textbf{1.00} & \textbf{0.21} & \textbf{1.00} & \textbf{0.56} & \textbf{0.47}\\
    sv & 0.03 & \textbf{0.45} & 0.00 & 0.00 & 0.01 & 0.01 & \textbf{0.19} & 0.01 & 0.00 & 0.00 & 0.00 & 0.03 & \textbf{0.07} & 0.00 & 0.00\\
    th & \textbf{0.69} & \textbf{1.00} & \textbf{1.00} & \textbf{0.63} & \textbf{1.00} & \textbf{1.00} & \textbf{1.00} & \textbf{1.00} & 0.04 & \textbf{1.00} & \textbf{1.00} & \textbf{0.20} & \textbf{1.00} & \textbf{0.57} & 0.04\\
    uk & \textbf{1.00} & \textbf{1.00} & \textbf{1.00} & \textbf{1.00} & \textbf{0.26} & \textbf{1.00} & \textbf{1.00} & \textbf{0.43} & 0.04 & \textbf{1.00} & \textbf{0.57} & 0.05 & \textbf{1.00} & \textbf{0.24} & \textbf{1.00}\\
    zh & \textbf{0.42} & \textbf{0.67} & \textbf{1.00} & \textbf{0.31} & \textbf{1.00} & \textbf{1.00} & \textbf{1.00} & \textbf{1.00} & 0.05 & \textbf{1.00} & \textbf{1.00} & \textbf{0.09} & \textbf{1.00} & \textbf{0.22} & 0.04\\
    \bottomrule
    \end{tabular}
\end{table*}

\begin{table*}
    \caption{The frequency with which a \textbf{VoxPopuli monolingual} fine-tuned model pair showed no statistical difference (p-value above 0.05). No-pretraining is presented in bold-face font.}
    \label{tab:vp_monolingual_performance_posthoc_test_frequency}
    \small
    \centering
    \begin{tabular}{p{25mm}p{25mm}p{5mm}}
    \toprule
    \multicolumn{3}{c}{\textbf{Validation}}\\
    \midrule
    VP-es & VP-it & 16\\
    VP-nl & VP-sv & 14\\
    VP-fr & VP-nl & 13\\
    VP-fr & VP-sv & 13\\
    VP-fr & VP-it & 12\\
    VP-it & VP-sv & 12\\
    VP-es & VP-fr & 11\\
    VP-es & VP-sv & 11\\
    VP-nl & \textbf{no-pretraining} & 10\\
    VP-it & VP-nl & 10\\
    VP-es & VP-nl & 9\\
    VP-it & \textbf{no-pretraining} & 8\\
    VP-es & \textbf{no-pretraining} & 7\\
    VP-sv & \textbf{no-pretraining} & 6\\
    VP-fr & \textbf{no-pretraining} & 5\\
    \midrule
    \multicolumn{3}{c}{\textbf{Test}}\\
    \midrule
    VP-es & VP-it & 15\\
    VP-nl & VP-sv & 15\\
    VP-es & VP-fr & 13\\
    VP-fr & VP-nl & 13\\
    VP-fr & VP-sv & 13\\
    VP-es & VP-sv & 12\\
    VP-it & VP-sv & 12\\
    VP-es & VP-nl & 11\\
    VP-fr & VP-it & 11\\
    VP-it & VP-nl & 11\\
    VP-nl & \textbf{no-pretraining} & 10\\
    VP-it & \textbf{no-pretraining} & 9\\
    VP-es & \textbf{no-pretraining} & 6\\
    VP-sv & \textbf{no-pretraining} & 5\\
    VP-fr & \textbf{no-pretraining} & 4\\
    \bottomrule
    \end{tabular}
\end{table*}

\clearpage
\begin{figure*}
  \centering
  \includegraphics[width=\textwidth]{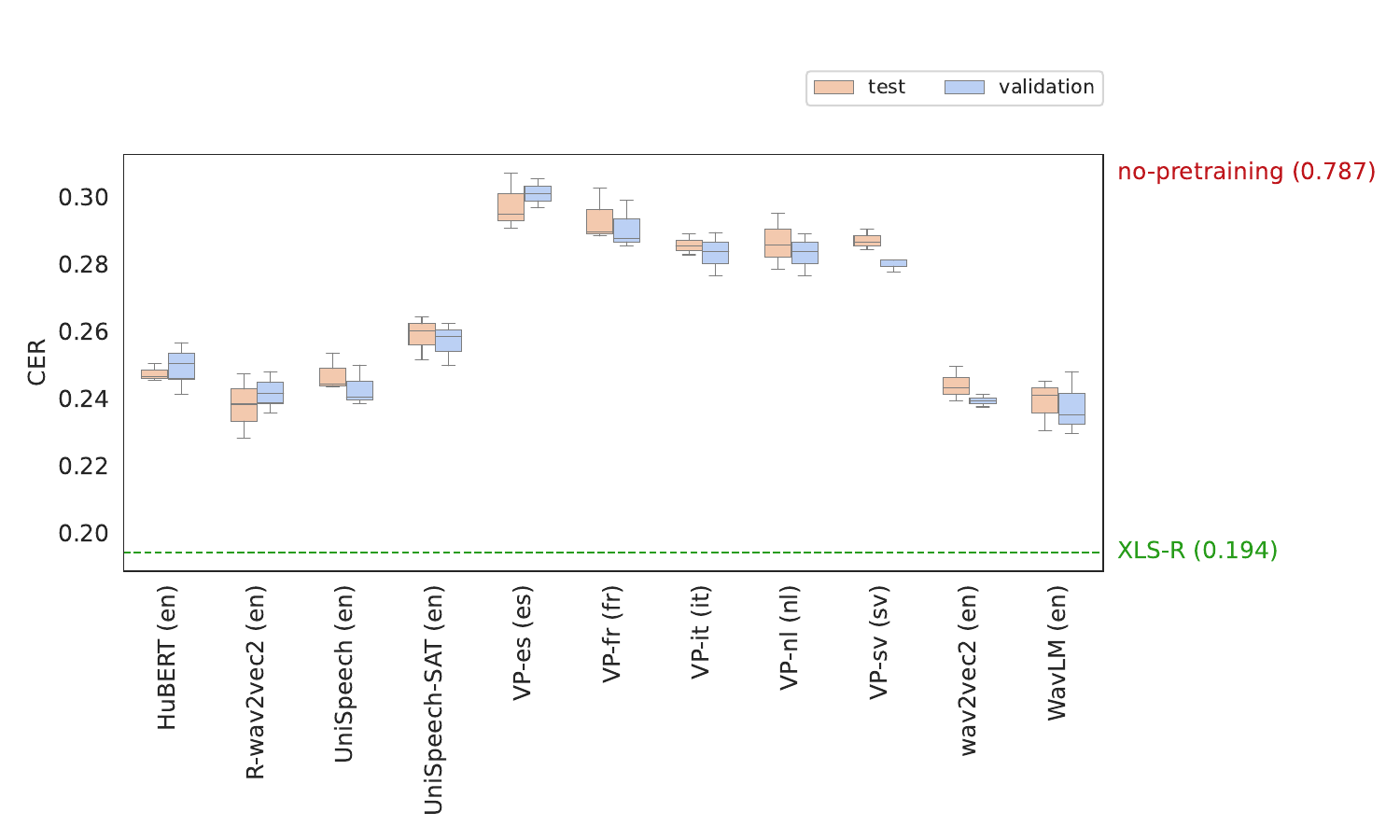}
  \caption{Overall performance over the \textbf{monolingual} pre-trained models for the \textbf{Arabic} language. The dashed green line is the performance of the best multilingual of the language and the red one is the performance of the fine-tuned model without any pre-training.}
  \label{fig:f_monolingual_ar}
\end{figure*}

\begin{figure*}
  \centering
  \includegraphics[width=\textwidth]{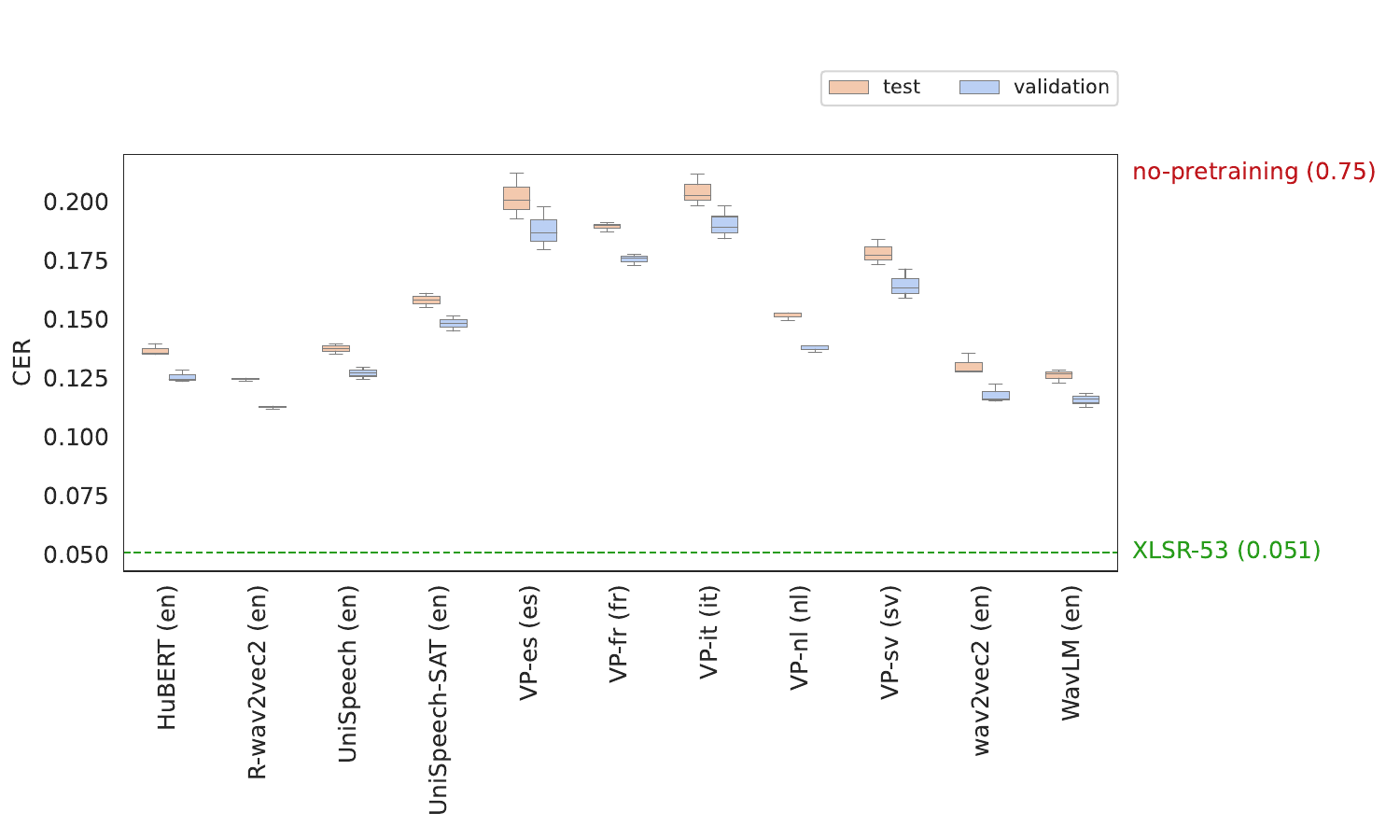}
  \caption{Overall performance over the \textbf{monolingual} pre-trained models for the \textbf{German} language. The dashed green line is the performance of the best multilingual of the language and the red one is the performance of the fine-tuned model without any pre-training.}
  \label{fig:f_monolingual_de}
\end{figure*}

\begin{figure*}
  \centering
  \includegraphics[width=\textwidth]{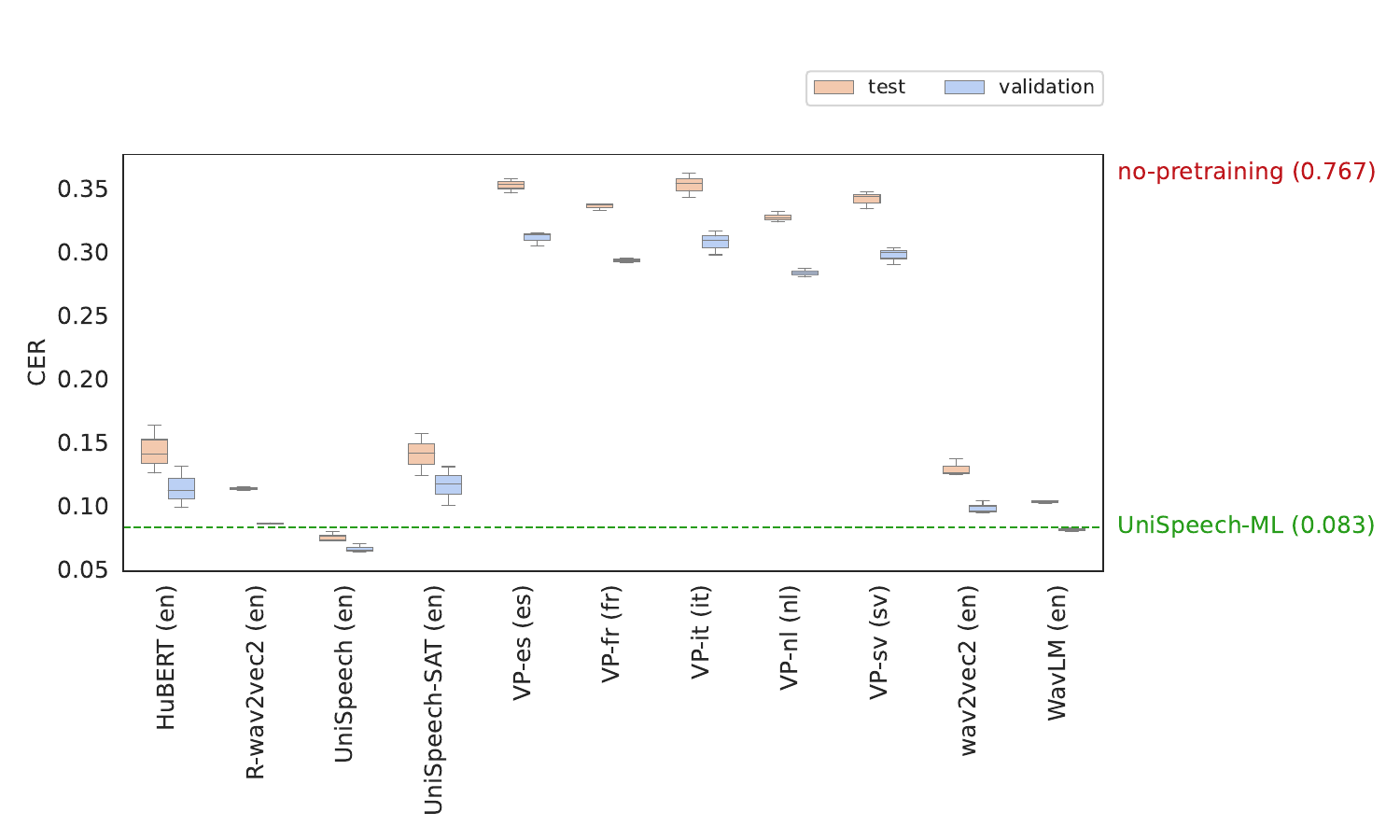}
  \caption{Overall performance over the \textbf{monolingual} pre-trained models for the \textbf{English} language. The dashed green line is the performance of the best multilingual of the language and the red one is the performance of the fine-tuned model without any pre-training.}
  \label{fig:f_monolingual_en}
\end{figure*}

\begin{figure*}
  \centering
  \includegraphics[width=\textwidth]{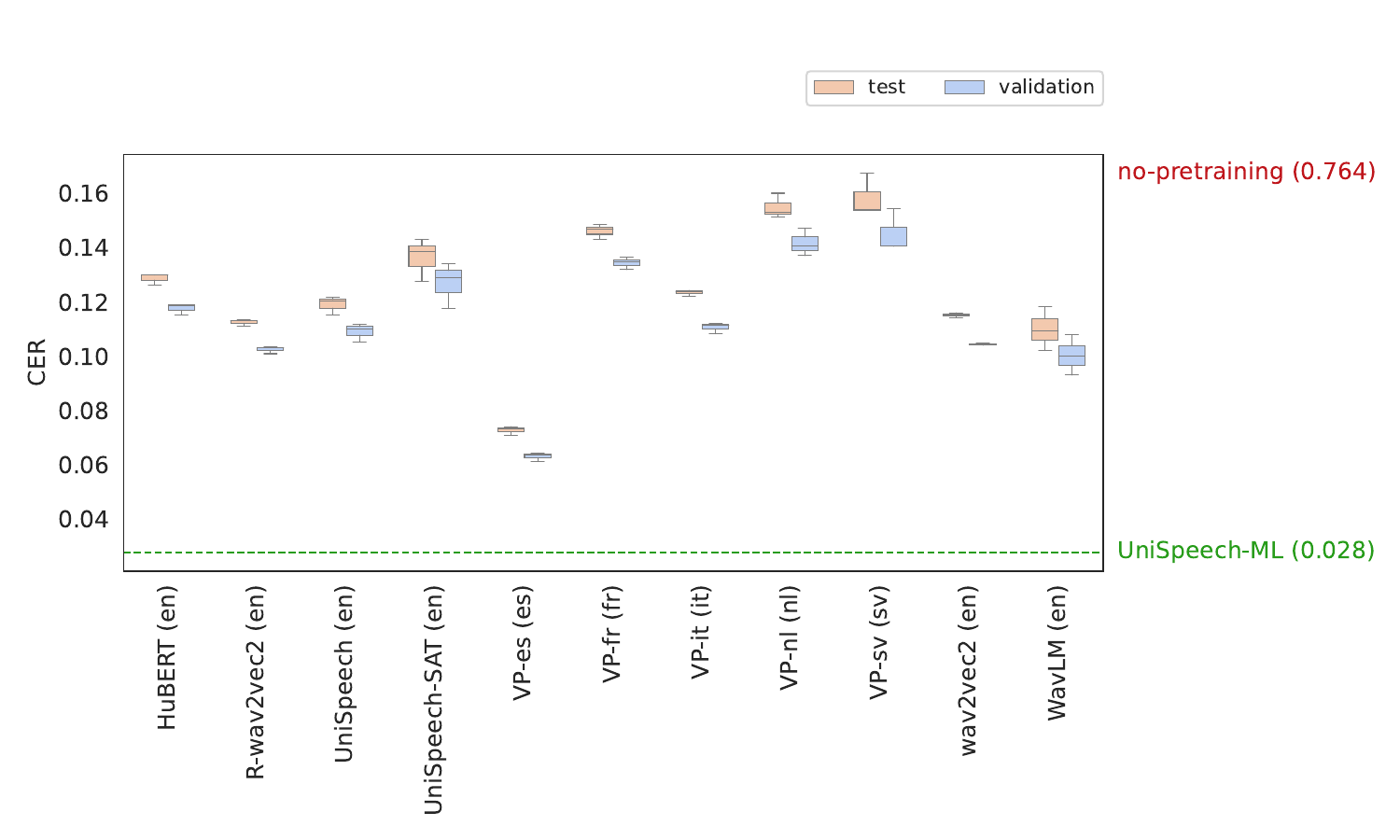}
  \caption{Overall performance over the \textbf{monolingual} pre-trained models for the \textbf{Spanish} language. The dashed green line is the performance of the best multilingual of the language and the red one is the performance of the fine-tuned model without any pre-training.}
  \label{fig:f_monolingual_es}
\end{figure*}

\begin{figure*}
  \centering
  \includegraphics[width=\textwidth]{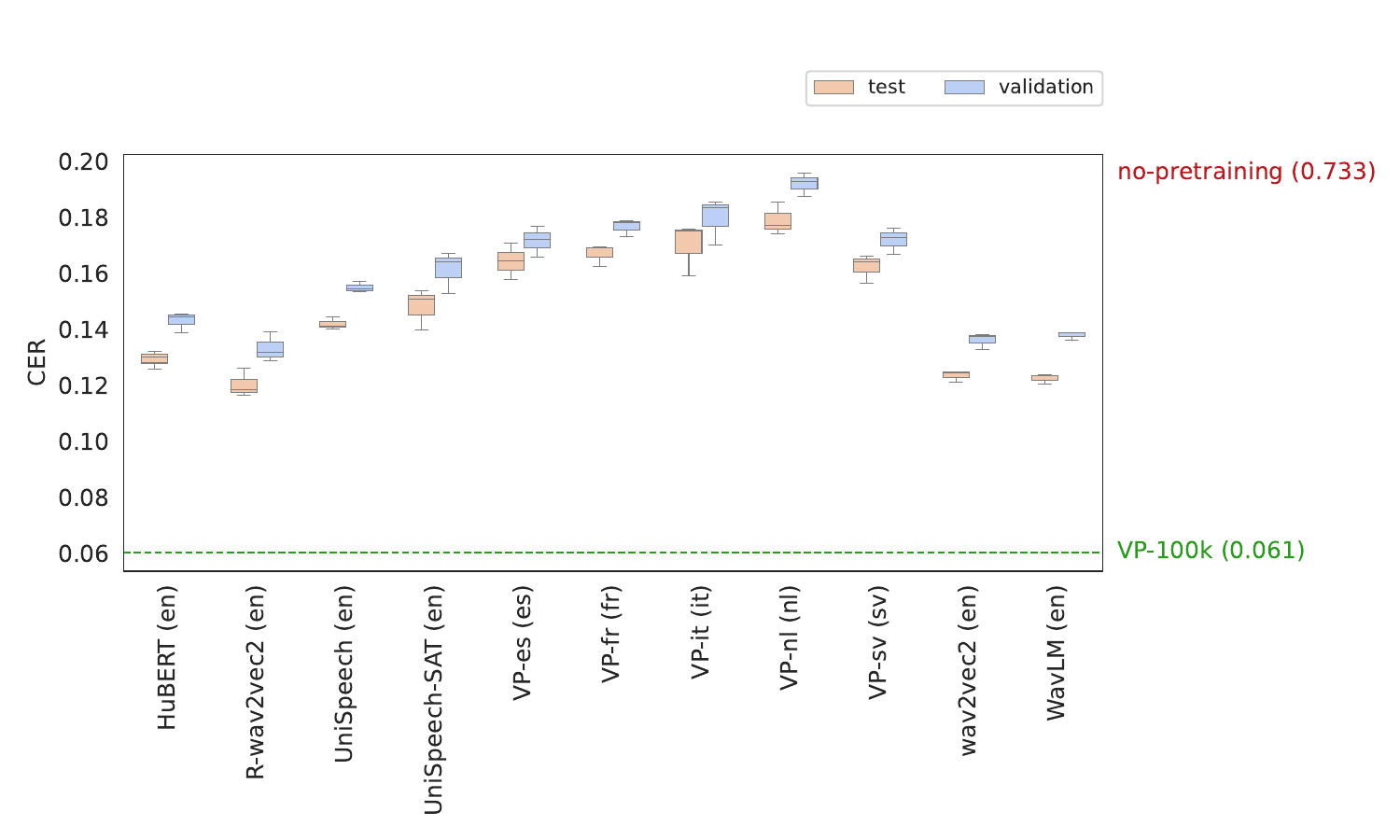}
  \caption{Overall performance over the \textbf{monolingual} pre-trained models for the \textbf{Estonian} language. The dashed green line is the performance of the best multilingual of the language and the red one is the performance of the fine-tuned model without any pre-training.}
  \label{fig:f_monolingual_et}
\end{figure*}

\begin{figure*}
  \centering
  \includegraphics[width=\textwidth]{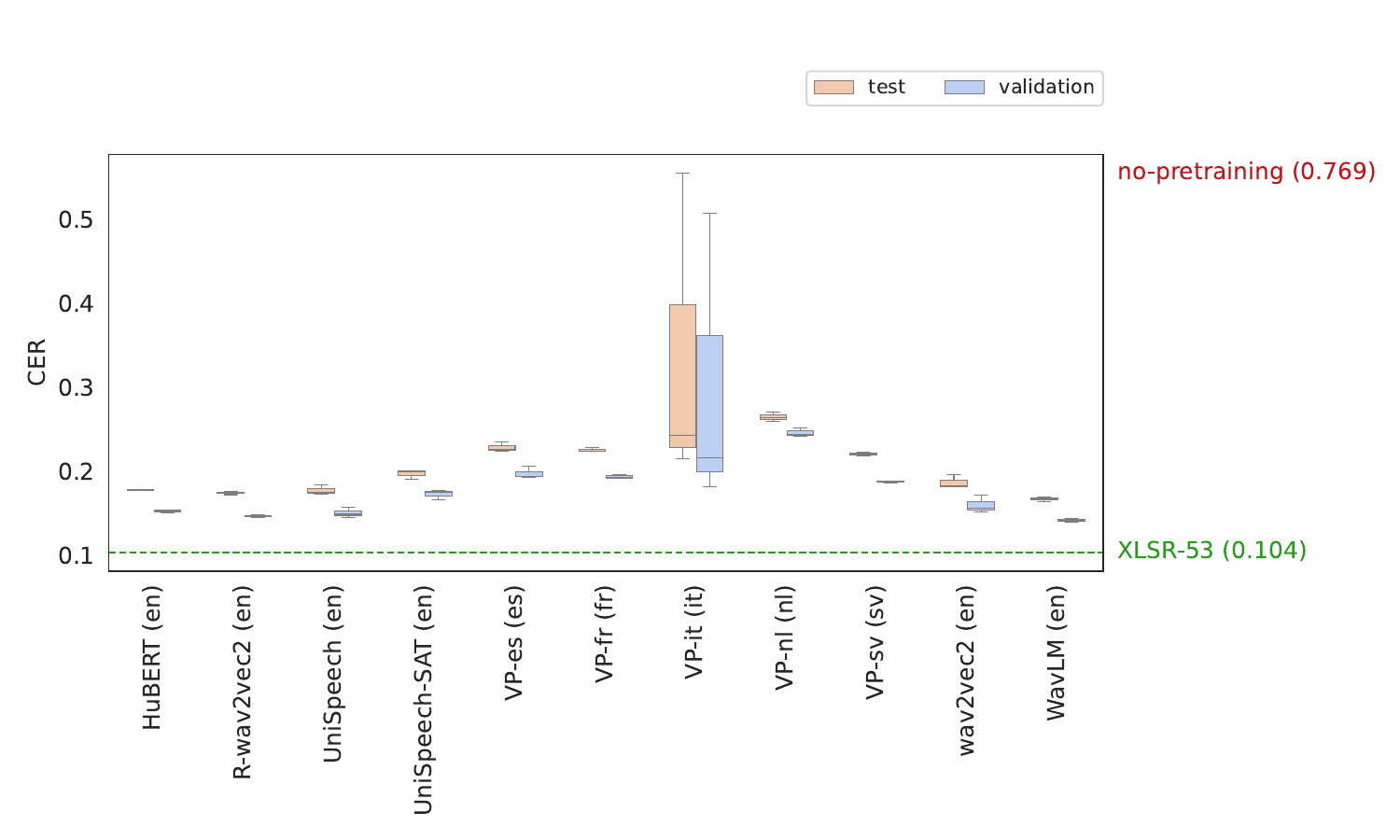}
  \caption{Overall performance over the \textbf{monolingual} pre-trained models for the \textbf{Persian} language. The dashed green line is the performance of the best multilingual of the language and the red one is the performance of the fine-tuned model without any pre-training.}
  \label{fig:f_monolingual_fa}
\end{figure*}

\begin{figure*}
  \centering
  \includegraphics[width=\textwidth]{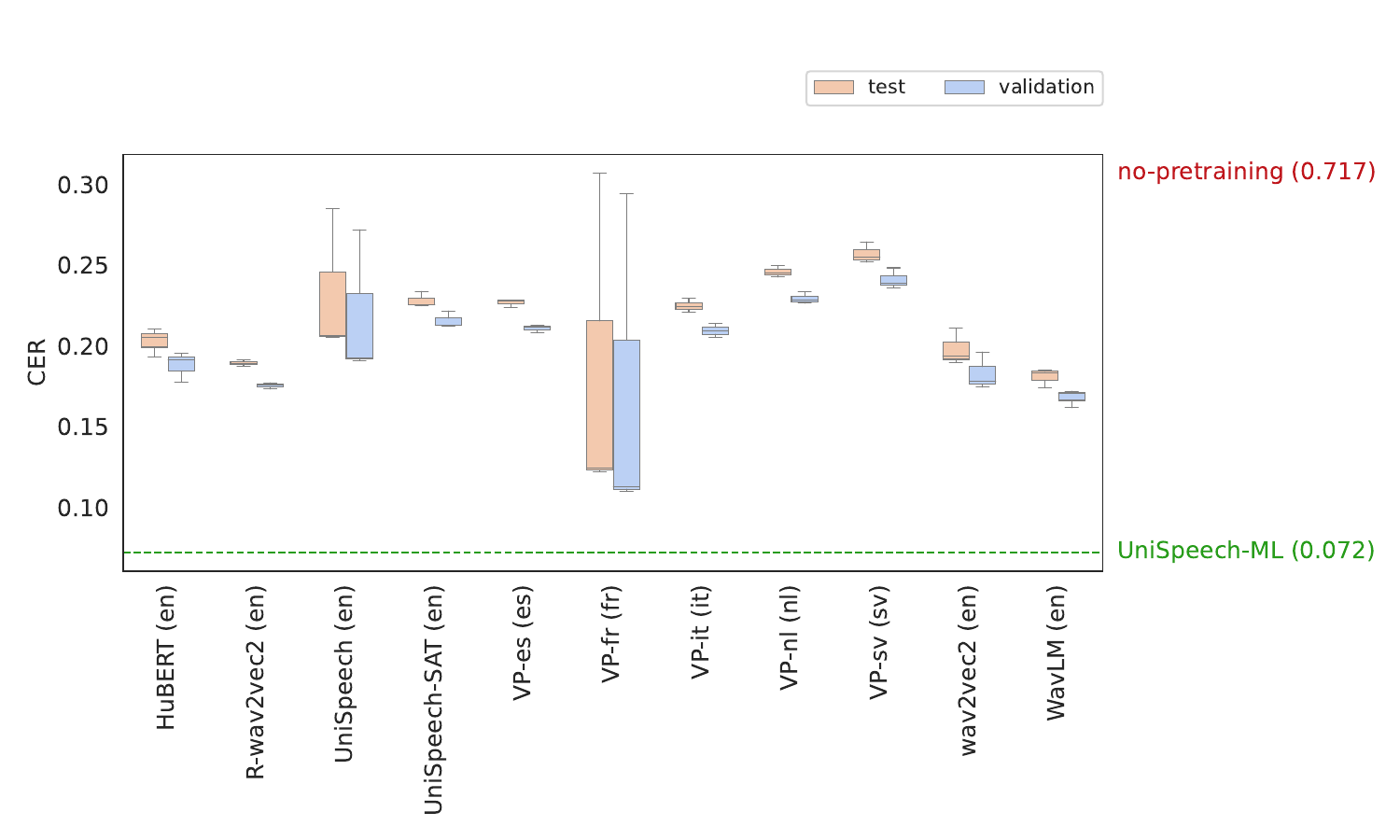}
  \caption{Overall performance over the \textbf{monolingual} pre-trained models for the \textbf{French} language. The dashed green line is the performance of the best multilingual of the language and the red one is the performance of the fine-tuned model without any pre-training.}
  \label{fig:f_monolingual_fr}
\end{figure*}

\begin{figure*}
  \centering
  \includegraphics[width=\textwidth]{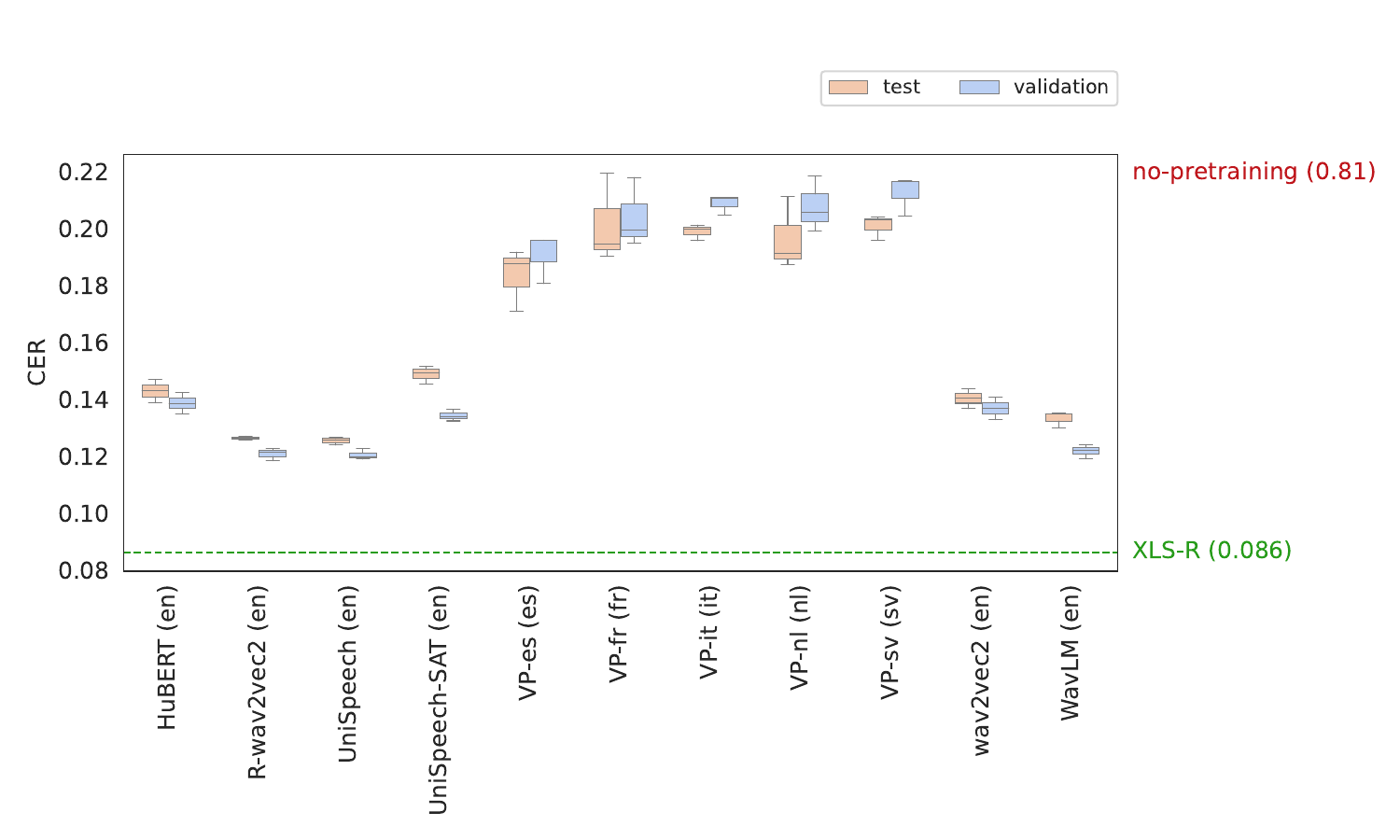}
  \caption{Overall performance over the \textbf{monolingual} pre-trained models for the \textbf{Indonesian} language. The dashed green line is the performance of the best multilingual of the language and the red one is the performance of the fine-tuned model without any pre-training.}
  \label{fig:f_monolingual_id}
\end{figure*}

\begin{figure*}
  \centering
  \includegraphics[width=\textwidth]{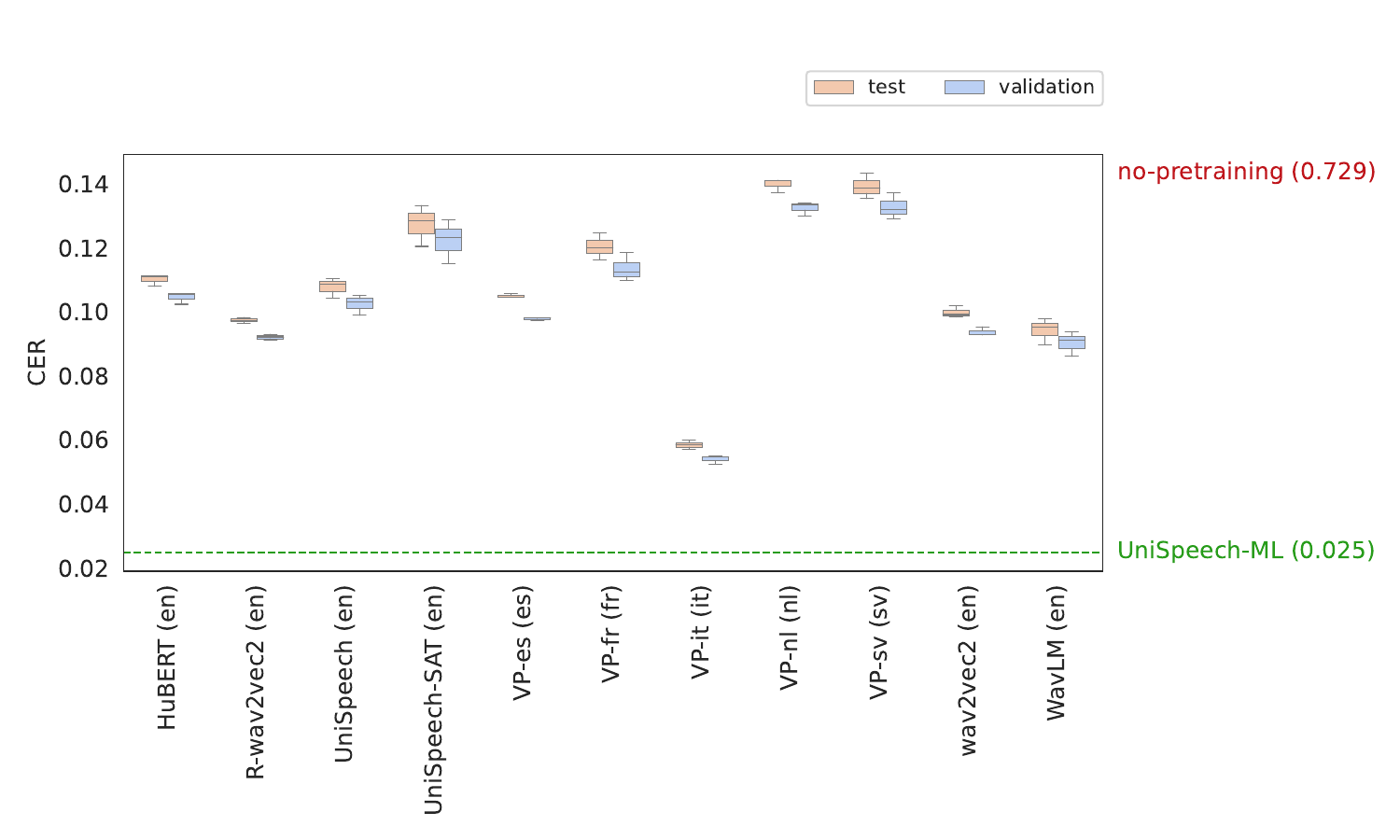}
  \caption{Overall performance over the \textbf{monolingual} pre-trained models for the \textbf{Italian} language. The dashed green line is the performance of the best multilingual of the language and the red one is the performance of the fine-tuned model without any pre-training.}
  \label{fig:f_monolingual_it}
\end{figure*}

\begin{figure*}
  \centering
  \includegraphics[width=\textwidth]{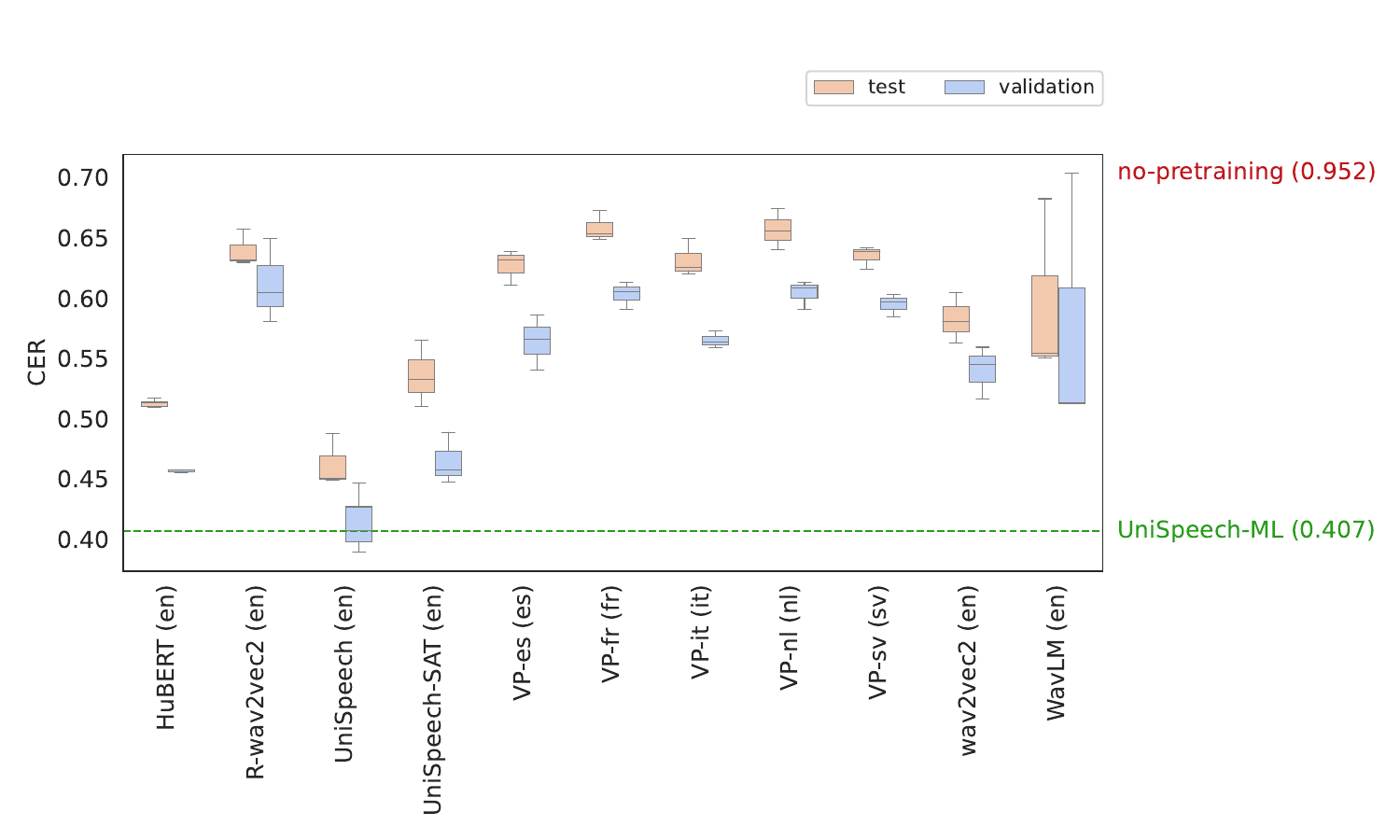}
  \caption{Overall performance over the \textbf{monolingual} pre-trained models for the \textbf{Japanese} language. The dashed green line is the performance of the best multilingual of the language and the red one is the performance of the fine-tuned model without any pre-training.}
  \label{fig:f_monolingual_ja}
\end{figure*}

\begin{figure*}
  \centering
  \includegraphics[width=\textwidth]{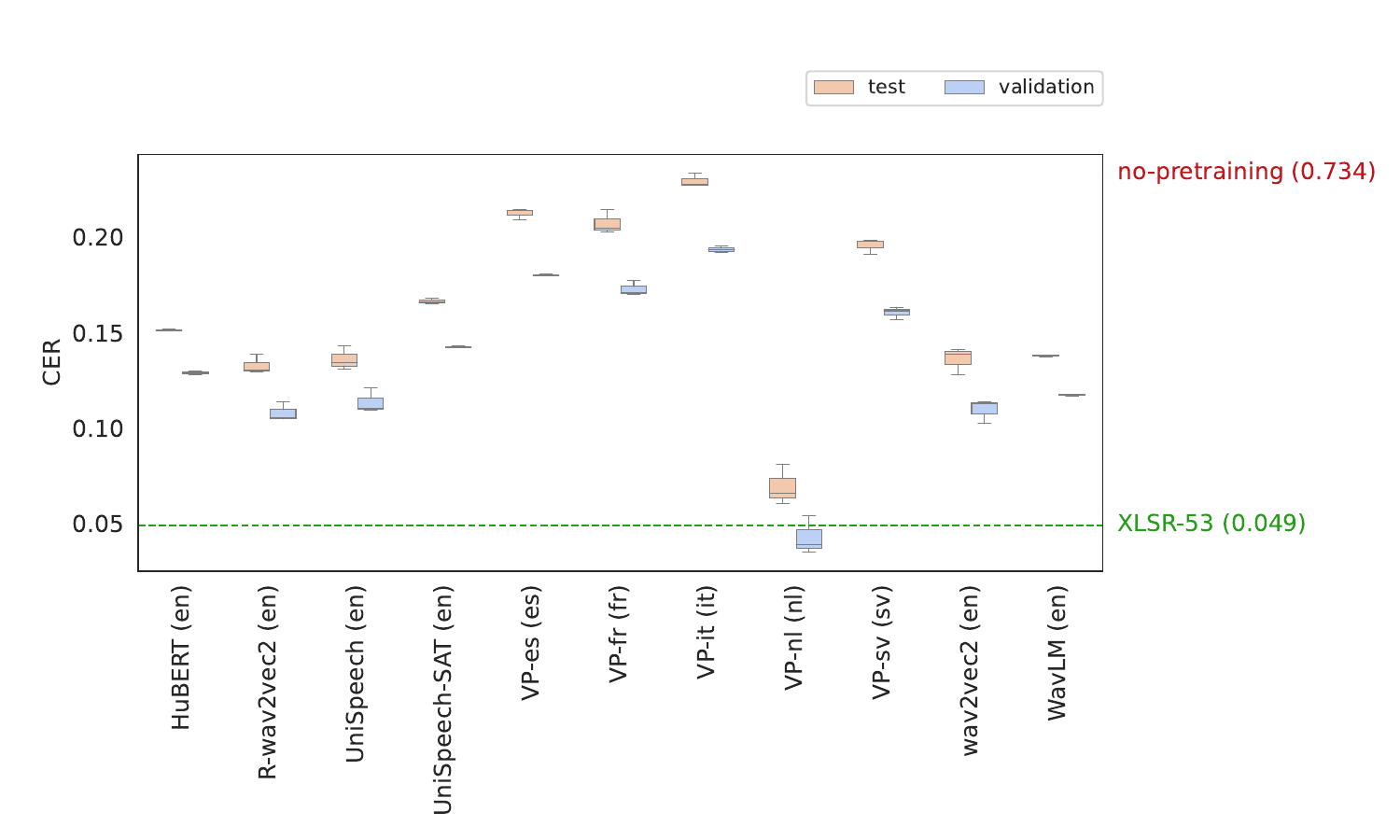}
  \caption{Overall performance over the \textbf{monolingual} pre-trained models for the \textbf{Dutch} language. The dashed green line is the performance of the best multilingual of the language and the red one is the performance of the fine-tuned model without any pre-training.}
  \label{fig:f_monolingual_nl}
\end{figure*}

\begin{figure*}
  \centering
  \includegraphics[width=\textwidth]{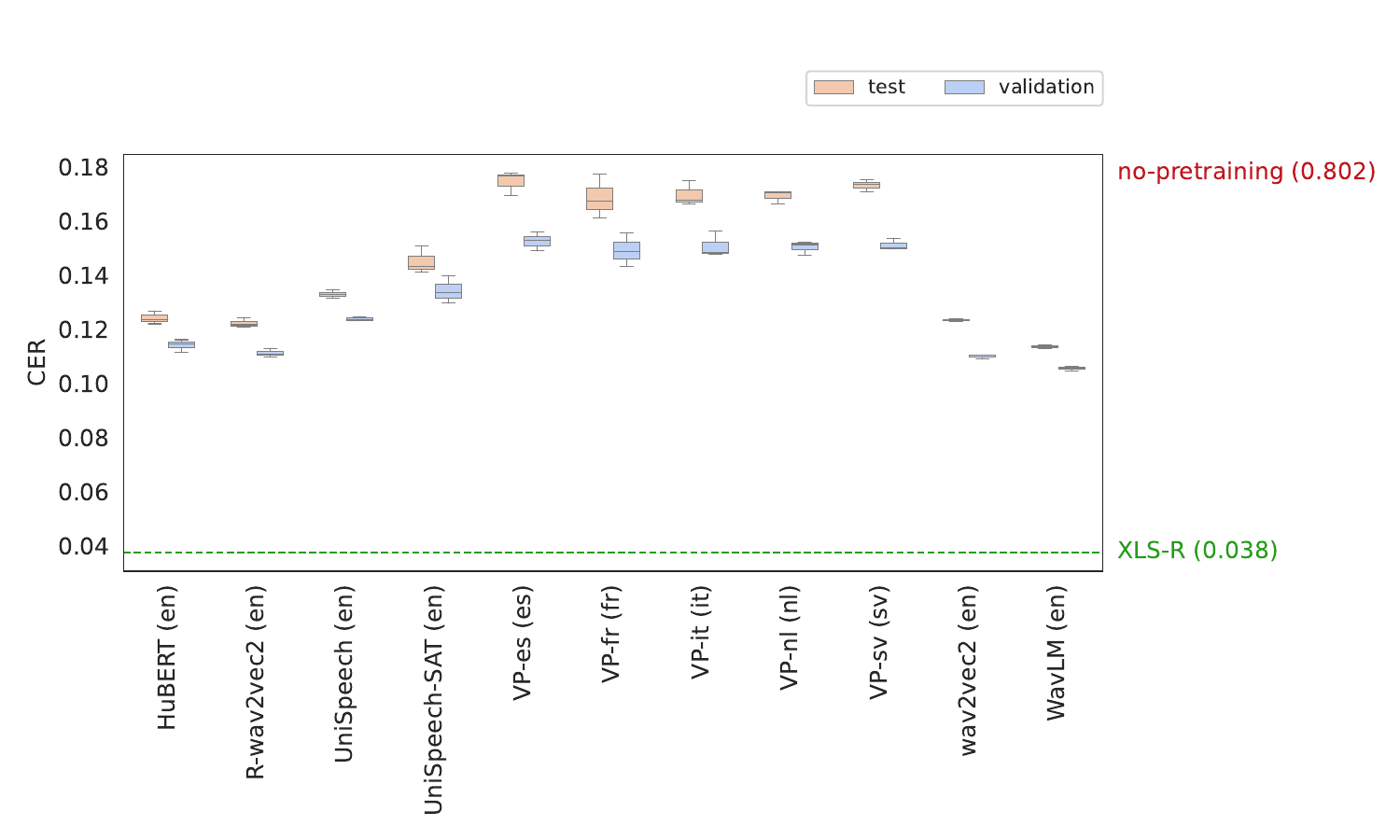}
  \caption{Overall performance over the \textbf{monolingual} pre-trained models for the \textbf{Polish} language. The dashed green line is the performance of the best multilingual of the language and the red one is the performance of the fine-tuned model without any pre-training.}
  \label{fig:f_monolingual_pl}
\end{figure*}

\begin{figure*}
  \centering
  \includegraphics[width=\textwidth]{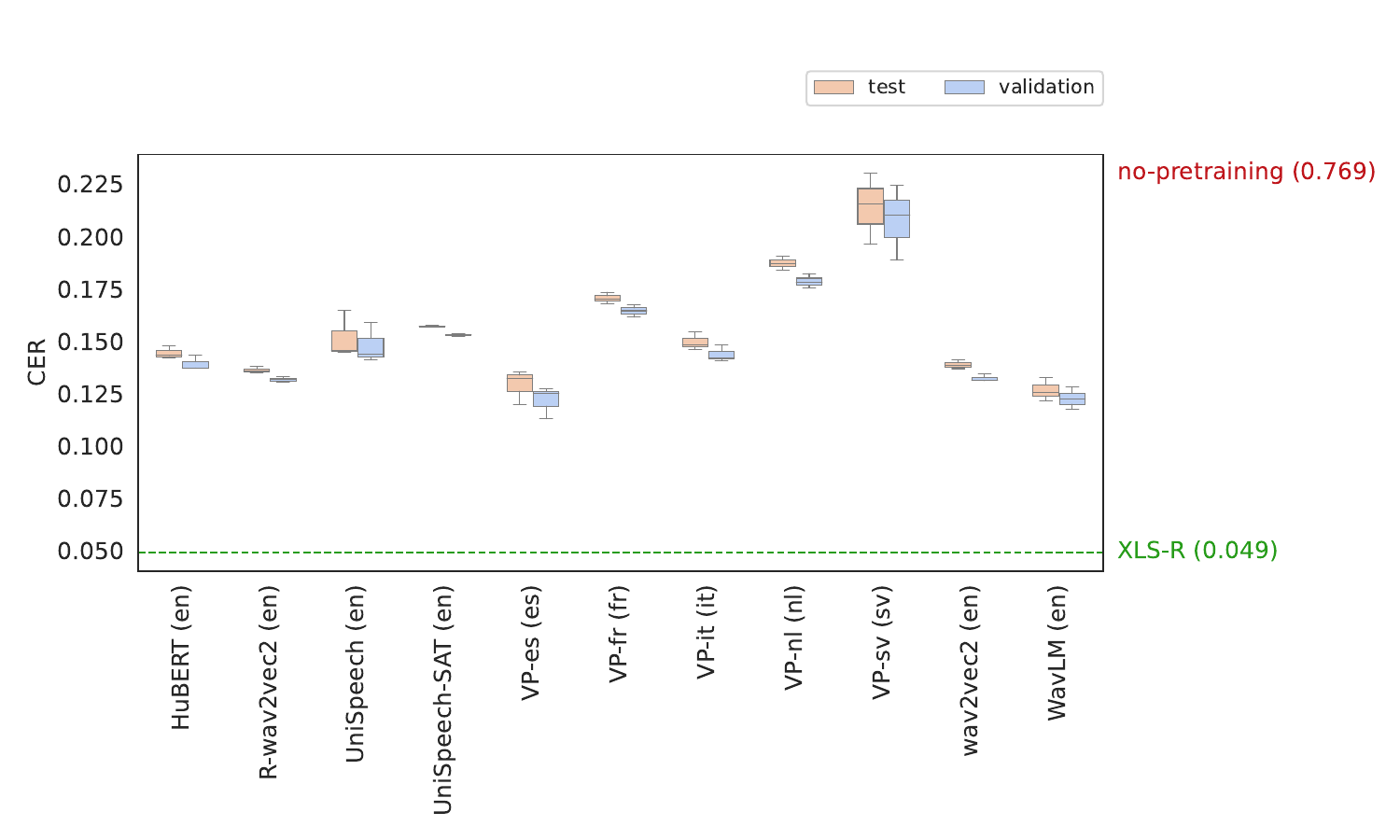}
  \caption{Overall performance over the \textbf{monolingual} pre-trained models for the \textbf{Portuguese} language. The dashed green line is the performance of the best multilingual of the language and the red one is the performance of the fine-tuned model without any pre-training.}
  \label{fig:f_monolingual_pt}
\end{figure*}

\begin{figure*}
  \centering
  \includegraphics[width=\textwidth]{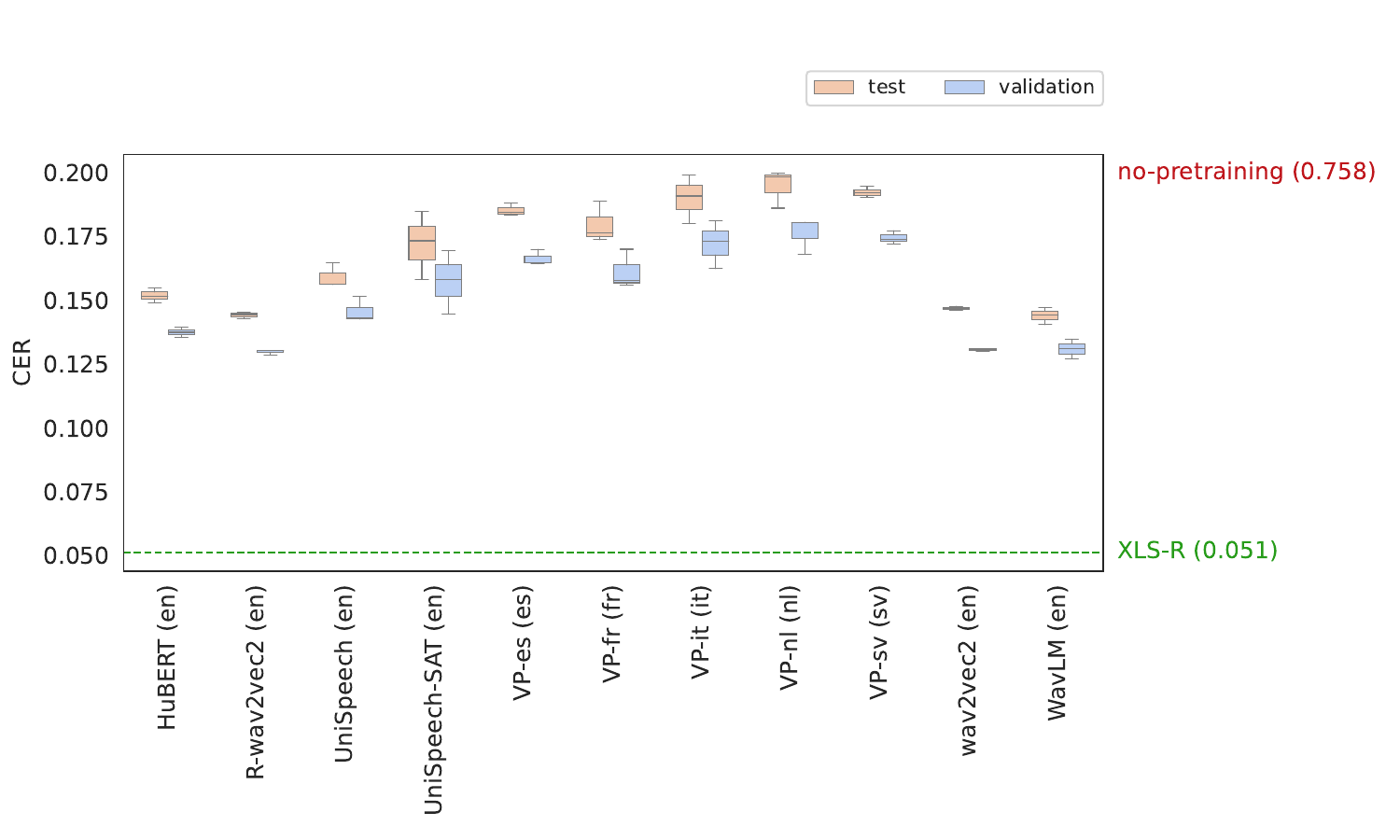}
  \caption{Overall performance over the \textbf{monolingual} pre-trained models for the \textbf{Russian} language. The dashed green line is the performance of the best multilingual of the language and the red one is the performance of the fine-tuned model without any pre-training.}
  \label{fig:f_monolingual_ru}
\end{figure*}

\begin{figure*}
  \centering
  \includegraphics[width=\textwidth]{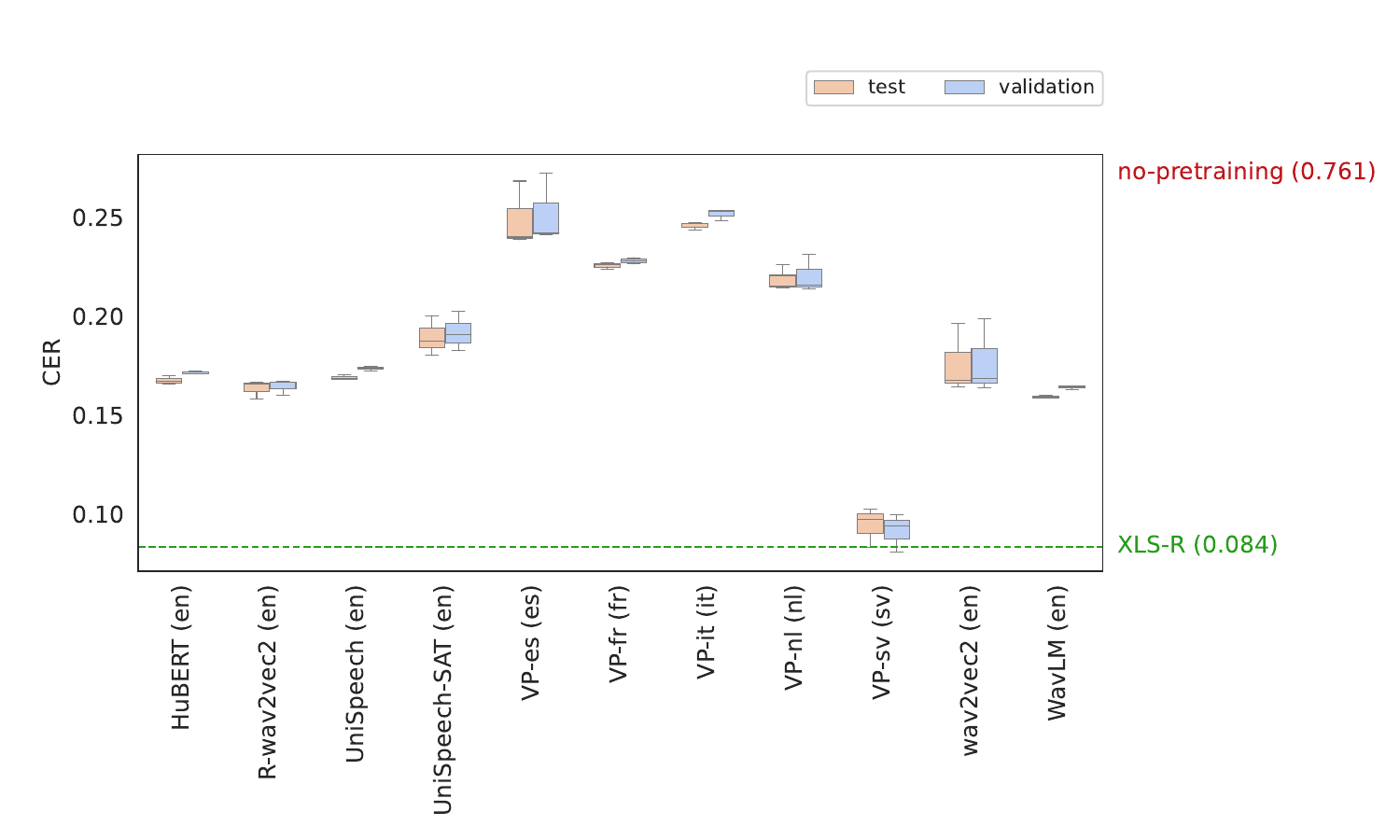}
  \caption{Overall performance over the \textbf{monolingual} pre-trained models for the \textbf{Swedish} language. The dashed green line is the performance of the best multilingual of the language and the red one is the performance of the fine-tuned model without any pre-training.}
  \label{fig:f_monolingual_sv}
\end{figure*}

\begin{figure*}
  \centering
  \includegraphics[width=\textwidth]{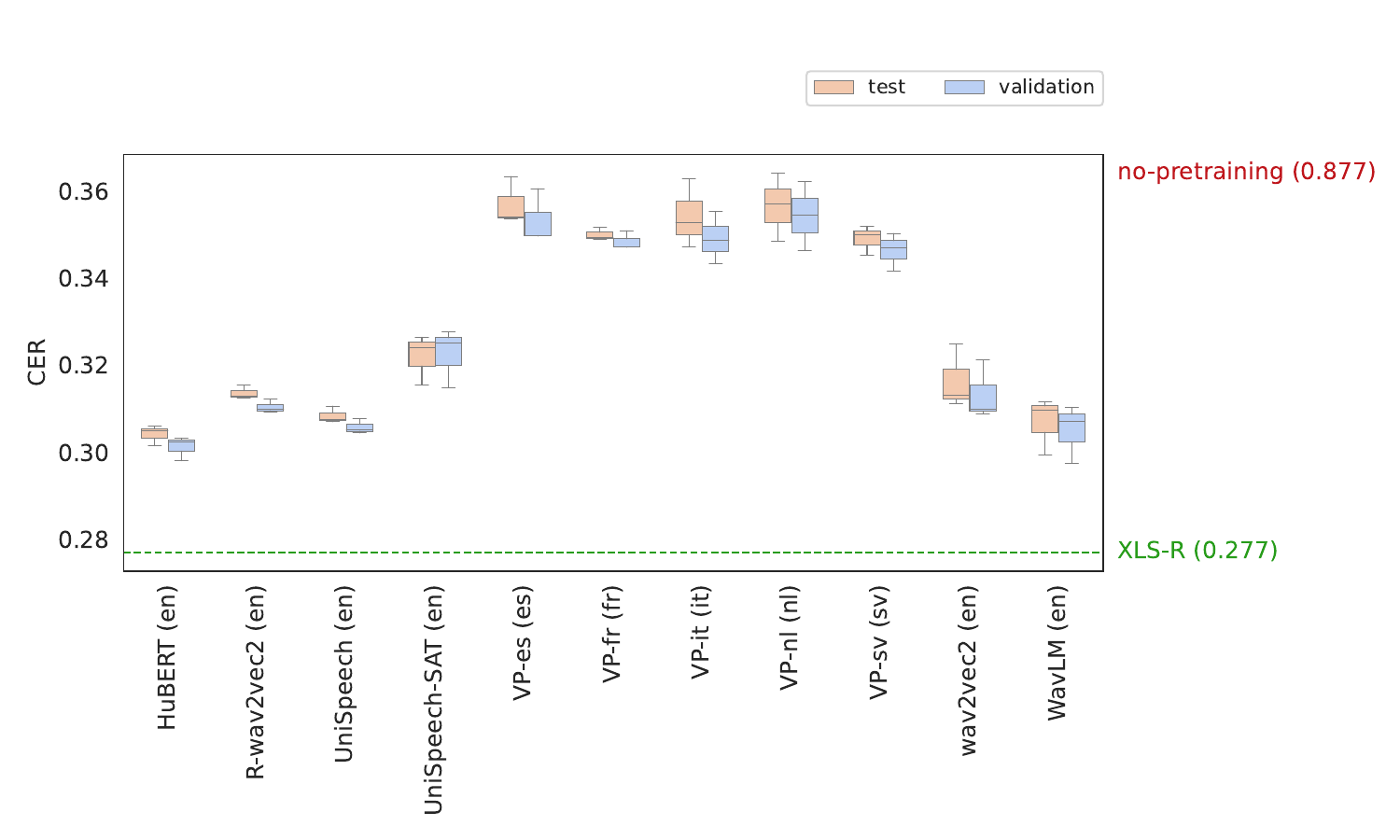}
  \caption{Overall performance over the \textbf{monolingual} pre-trained models for the \textbf{Thai} language. The dashed green line is the performance of the best multilingual of the language and the red one is the performance of the fine-tuned model without any pre-training.}
  \label{fig:f_monolingual_th}
\end{figure*}

\begin{figure*}
  \centering
  \includegraphics[width=\textwidth]{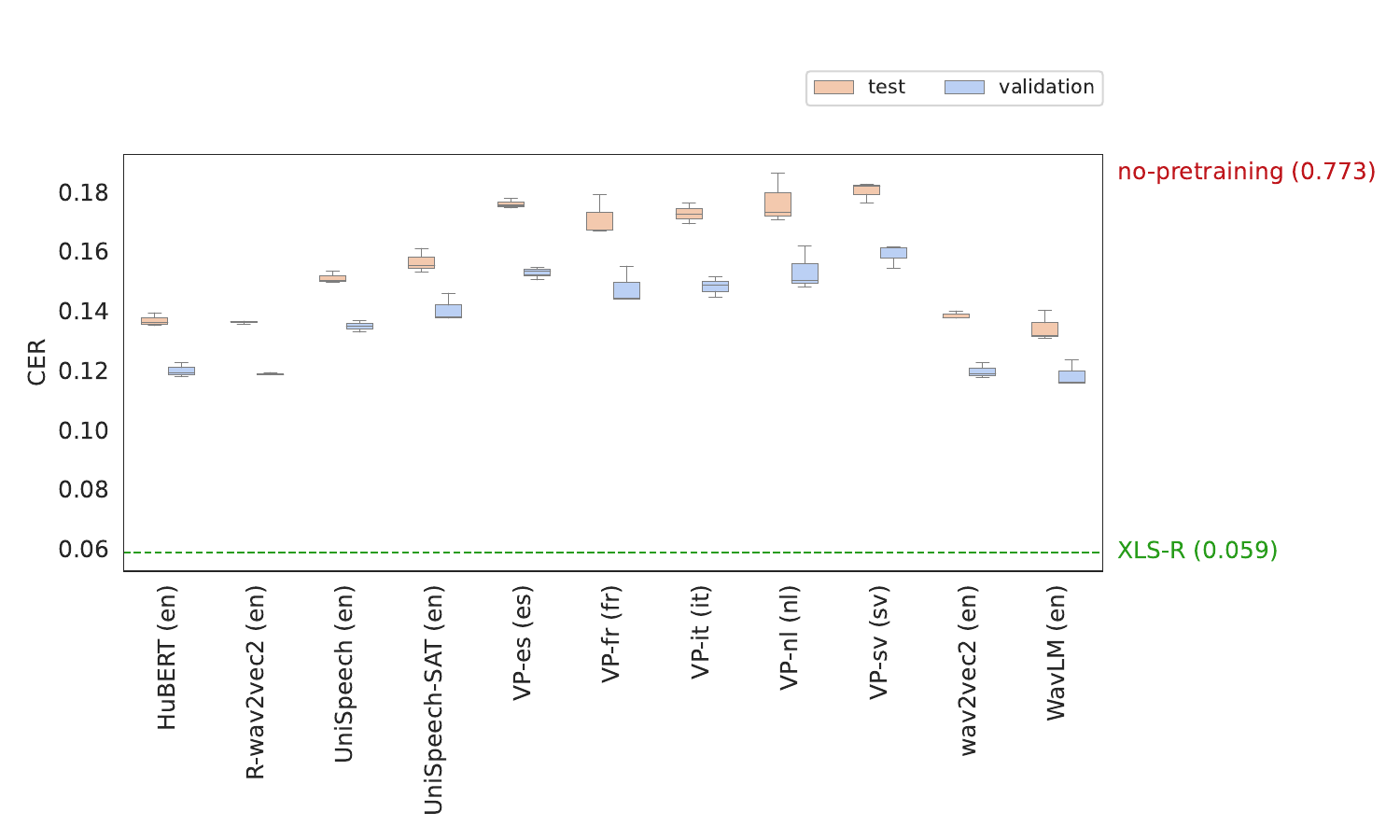}
  \caption{Overall performance over the \textbf{monolingual} pre-trained models for the \textbf{Ukrainian} language. The dashed green line is the performance of the best multilingual of the language and the red one is the performance of the fine-tuned model without any pre-training.}
  \label{fig:f_monolingual_uk}
\end{figure*}

\begin{figure*}
  \centering
  \includegraphics[width=\textwidth]{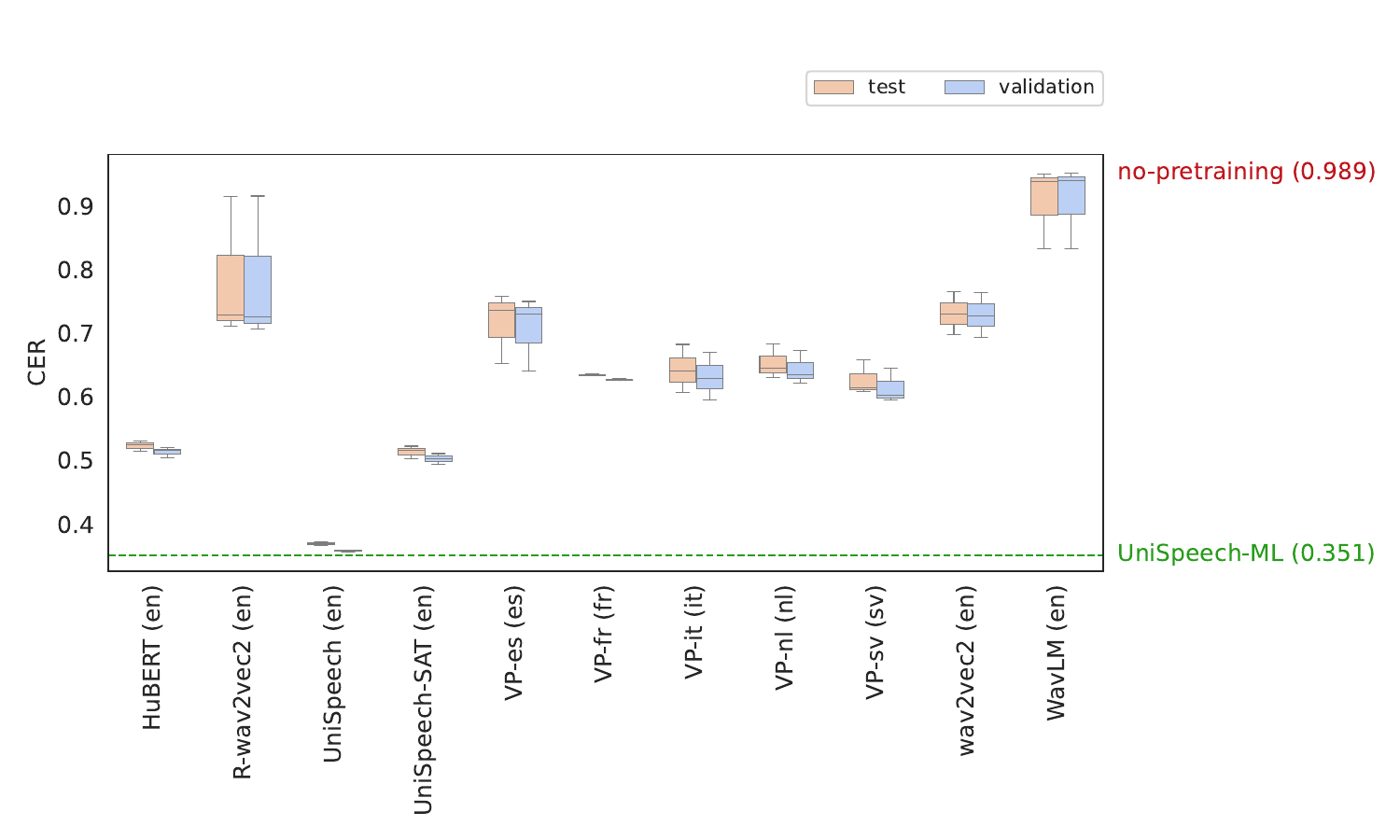}
  \caption{Overall performance over the \textbf{monolingual} pre-trained models for the \textbf{Chinese} language. The dashed green line is the performance of the best multilingual of the language and the red one is the performance of the fine-tuned model without any pre-training.}
  \label{fig:f_monolingual_zh}
\end{figure*}

\end{document}